\documentclass[jounal]{acmsiggraph}

\usepackage{booktabs} 

\usepackage[ruled]{algorithm2e} 
\usepackage{amsmath}
\usepackage{amssymb}
\usepackage{mathrsfs}
\usepackage[british,UKenglish,USenglish,english,american]{babel}
\usepackage{booktabs}
\usepackage{color}
\usepackage{subfigure}
\usepackage{array}

\newcommand{\Eref}[1]{Equation~(\ref{#1})}
\newcommand{\Fref}[1]{Figure~\ref{#1}}
\newcommand{\Sref}[1]{Section~\ref{#1}}

\newcommand{\fref}[1]{Figure~\ref{#1}}

\newcommand\ignore[1]{}

\title{Visual Attribute Transfer through Deep Image Analogy \thanks{Supplemental material: \url{https://liaojing.github.io/html/data/analogy_supplemental.pdf}; source code: \url{https://github.com/msracver/Deep-Image-Analogy}. }}

\author{Jing Liao$^1$, Yuan Yao$^2$ \thanks{This work was done when Yuan Yao was an intern at MSR Asia.}, Lu Yuan$^1$,
       Gang Hua$^1$, and Sing Bing Kang$^1$
       \\
       $^1$Microsoft Research,  
       $^2$Shanghai Jiao Tong University
       }

\pdfauthor{Liao et al.}


\begin{document}

\teaser{
   \includegraphics[width=0.95\linewidth]{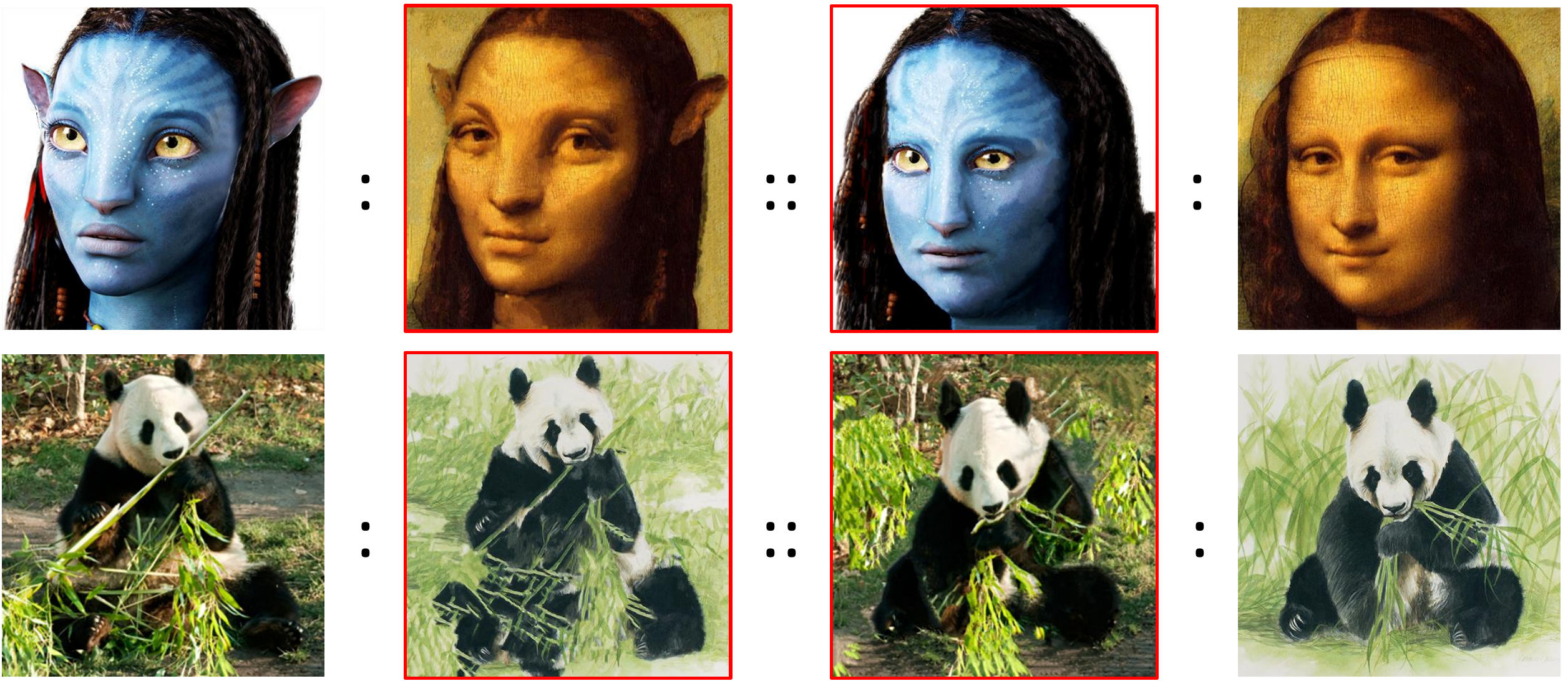}
   \centering
{
\begin{tabular}{p{.22\linewidth}p{.22\linewidth}p{.22\linewidth}p{.22\linewidth}}
\centering{$A \quad \!$(input)$\!$} & \centering{$A' \quad  \!$(output)$\!$} & \centering{$B \quad  \!$(output)$\!$} & \centering{$B'\quad  \!$(input)$\!$}
\end{tabular}

}

   \caption{Our technique allows us to establish semantically-meaningful dense correspondences between two input images $A$ and $B'$. $A'$ and $B$ are the reconstructed results subsequent to transfer of visual attributes.}
   \label{fig:teaser}
 }
\maketitle

\begin{abstract}

We propose a new technique for visual attribute transfer across images that may have very different appearance but have perceptually similar semantic structure. 
By visual attribute transfer, we mean transfer of visual information (such as color, tone, texture, and style) from one image to another. 
For example, one image could be that of a painting or a sketch while the other is a photo of a real scene, and both depict the same type of scene.

Our technique finds semantically-meaningful dense correspondences between two input images. To accomplish this, it adapts the notion of ``image analogy"~\cite{hertzmann2001analogy} with features extracted from a Deep Convolutional Neutral Network for matching; we call our technique \emph{deep image analogy}. A coarse-to-fine strategy is used to compute the nearest-neighbor field for generating the results. We validate the effectiveness of our proposed method in a variety of cases, including style/texture transfer, color/style swap, sketch/painting to photo, and time lapse.

\end{abstract}
\section{Introduction}

Many types of compelling image stylization effects have been demonstrated over the years, including color transfer, texture transfer, and style transfer. Their appeal is especially strong in the context of social media, where photo sharing and entertainment are important elements. A number of popular apps such as Prisma and Facetune have successfully capitalized on this appeal.

The applications of color transfer, texture transfer, and style transfer share a common theme, which we characterize as {\em visual attribute transfer}. In other words, visual attribute transfer refers to copying visual attributes of one image such as color, texture, and style, to another image.

In this paper, we describe a new technique for visual attribute transfer for a pair of images that may be very different in appearance but semantically similar. In the context of images, by "semantic", we refer to high-level visual content involving identifiable objects. We deem two different images to be semantically similar if both are of the same type of scene containing objects with the same names or classes, e.g., nature scenes featuring pandas and foliage, headshots, or indoor scenes featuring dining rooms with tables, chairs, walls, and ceiling. Figure~\ref{fig:teaser} shows two semantically similar pairs. The images in each pair of inputs ($A$ and $B'$) look dramatically different, but have objects with similar identities.

Our technique establishes semantically-meaningful dense correspondences between the input images, which allow effective visual attribute transfer. Low-level matching methods, such as optical flow~\cite{Brox2004flow} and PatchMatch~\cite{barnes2009patchmatch}, are designed to match local intensities. Hence, they fail to match under large visual variations.
While other methods such as SIFT flow~\cite{Liu2011siftflow} or deep match~\cite{weinzaepfel2013deepmatch} are more reliable in matching sparse features, they are also not able to handle extreme visual variations, such as matching across an artistic painting and a real photograph. This is because these methods are still fundamentally based on low-level features.

Methods have been proposed to match different particular domains, such as drawings/paintings to photographs~\cite{Russell2011painting}, sketches to photos~\cite{Chen2009sketch}, and photos under different illuminants~\cite{Chong2008illumination}. However, these methods typically are very domain-specific and do not easily generalize. Schechtman and Irani~\shortcite{Shechtman2007selfsimilarity} propose a more general solution using local self-similarity descriptors that are invariant across visual domains, and Shrivastava et. al.~\shortcite{shrivastava2011crossdomain} consider relative weighting between the descriptors for cross-domain image matching. Unfortunately, these methods do not produce dense correspondences between images of different domains.

We handle the dense correspondence problem using ideas related to \emph{image analogies}~\cite{hertzmann2001analogy}, which involve dense mappings between images from different domains. An image analogy is codified as $A : A' :: B : B'$, where $B'$ relates to $B$ in the same way as $A'$ relates to $A$, and additionally, $A$ and $A'$ (also $B$ and $B'$) are in pixel-wise correspondences. In forming an image analogy, typically $A$, $A'$, and either $B$ or $B'$ are given, and the goal is to solve for the sole remaining image. In contrast, for our scenario only a source image $A$ and an example image $B'$ are given, and both $A'$ and $B$ represent latent images to be estimated, imposing a bidirectional constraint to better match $A$ to $B'$.

Solving the visual attribute transfer problem is equivalent to finding both unknown images $A'$ and $B$. Instead of applying image analogy to image pixels directly, we use a Deep Convolutional Neutral Network (CNN)~\cite{krizhevsky2012imagenet} to construct a feature space in which to form image analogies. It has been shown that such deep features are better representations for semantic structures~\cite{zeiler2014cnn}. We call our new technique \emph{deep image analogy}.

Our approach leverages pre-trained CNNs for object recognition (e.g., VGG-19~\cite{simonyan2014very}) to construct such a feature space.
A nice property of CNN representations is that they gradually encode image information from low-level details to high-level semantic content. This provides a good decomposition of semantic structure and visual attributes for transfer. Besides, the spatial correspondence between intermediate feature layers in CNN architectures is approximately maintained. Both properties facilitate a coarse-to-fine strategy for computing the nearest-neighbor field (NNF), a core component used in reconstructing images. To speed up the required nearest-neighbor search, we adapt PatchMatch~\cite{barnes2009patchmatch} to the CNN feature domain. 
Our method uses the bidirectional constraint in the patch similarity metric, which have previously been shown to work well on re-targeting problems~\cite{Simakov2008bidirection}. In the present context, the use of the bidirectional constraint introduces a useful symmetry and helps to mitigate the risk of mismatches.

Our major technical contributions are:
\begin{enumerate}
  \item We present a new method ``deep image analogy", which we show to be effective for visual attribute transfer across images in very different domains.
  \item We extend PatchMatch and reconstruction from the image domain to the feature domain, which serves to guide semantically-meaningful visual attribute transfer.
\end{enumerate}

We show how our deep image analogy technique can be effectively applied to a variety of visual attribute transfer cases, namely style/texture transfer, color/style swap, sketch/painting to photo, and time lapse. Our technique also has the effect of generating {\em two} results instead of the typical one generated from a one-way transfer. Such results can be seen in~\Fref{fig:teaser}. Our deep image analogy is designed to work on images with similar content composition. It is not effective for images which are semantically unrelated (e.g., a headshot and a countryside photo), and is not designed to handle large geometric variations (including scales and rotations).

\section{Related Work}

In this section, we review techniques that are related to visual attribute transfer (color, texture, and style transfer, as well as image analogy). We also briefly discuss two other topics very relevant to our work: dense image correspondence and neural style transfer.

\subsection{Visual Attribute Transfer}

Much work has been done on the transfer of various visual attributes (e.g., color, tone, texture, style) from one image to another, and we discuss only representative papers for brevity. Most of these approaches are, however, not general, in that they are designed to transfer a specific type of visual attribute. As we show later, our technique is designed to handle more types of visual attributes.

\textbf{\emph{Color Transfer.}}  Early color transfer techniques tend to be global, i.e., a global transformation is applied to a source image to match the color statistics of a reference image~\cite{Reinhard2001color,Pitie2005NPD}. They work well when the source and reference images are of similar scenes, even though the spatial layouts can be dissimilar. Other methods incorporate user input~\cite{Xiaobo2010color,Welsh2002color} or a large database~\cite{dale09restore,laffont2014transient} to guide the color transfer. Local color transfer methods infer local color statistics in different color regions by establishing region correspondences~\cite{Yuwing2005color}. More recently, Claudio et al.~\shortcite{Claudio2012color} transfer local color between regions with the same annotated class; in a similar vein, Wu et al.~\shortcite{Wu2013color} transfer the color style across the same semantic regions.

\textbf{\emph{Texture Transfer.}} Most early texture transfer algorithms rely on non-parametric sampling for texture synthesis while using different ways to preserve the structure of the target image. For instance, Efros and Freeman~\shortcite{efros2001image} introduce a correspondence map that includes features of the target image such as image intensity to constrain the texture synthesis procedure. Ashikhmin~\shortcite{Ashikhmin2003texture} focuses on transferring high-frequency texture information while preserving the coarse scale of the target image. Lee et al.~\shortcite{Lee2010texture} improve this algorithm by additionally augmenting texture transfer with edge orientation information.

\textbf{\emph{Style Transfer.}} Style transfer is used as a means to migrate an artistic style from an example image to a source image. The decomposition of content and style in artistic images is bound to the coupling between the source content and the example style. Zhang \emph{et. al.}~\shortcite{Zhang2013style} decompose the input images into three components: draft, paint, and edge; the style is transferred from the template image to the source image through the paint and edge components. Frigo \emph{et. al.}~\shortcite{frigo2016split} view style transfer as the composition of local texture transfer and global color transfer, and suggest a simple yet efficient adaptive image partition for the decomposition of style and content. Shih \emph{et. al.}~\shortcite{shih2014style} robustly transfer the local statistics of an example portrait onto a new one using a multi-scale technique that separates local detail from structure.

\textbf{\emph{Image Analogy.}} Texture or style transfer can also be done in a more supervised manner, where a pair of images $A$ and $A'$ are manually registered, and the analog of image $B$ (similar in style with $A$) is to be found (resulting in $B'$). This approach, called image analogy, is first reported in \cite{hertzmann2001analogy} and extended in various ways~\cite{Cheng2008analogy,Benard2013animation,scott2015analogy,Barnes2015PatchTable}. In our work, we apply the concept of image analogy as a weak supervision, in a scenario where only an example image $B'$ and an original image $A$ are given.

\subsection{Dense Correspondence}

Finding dense correspondences between two images is a fundamental problem in computer vision and graphics. Initial correspondence methods were designed for stereo matching, optical flow, and image alignment~\cite{lucas1981flow}. These methods compute a dense correspondence field, but they assume brightness consistency and local motion, and may be hard to handle occlusion well.

The development of various local invariant features (e.g., SIFT~\cite{Lowe2004sift}) has brought up significant progress. These features are robust to typical appearance variations (illumination, blur, compression), and a wide range of 2D transformations. Some methods combine sparse features with dense matching to cope with large-displacement optical flow~\cite{Brox2009largeflow}, while others perform matching of visually different scenes (e.g., SIFT-flow~\cite{Liu2011siftflow}, Deep flow~\cite{weinzaepfel2013deepmatch}, Daisy flow~\cite{Yang2014daisyflow}, and Region foremost~\cite{Shen2016RFM}). More recently, CNN-based features such as outputs of a certain convolution layer~\cite{long2014convnets}), object proposals~\cite{Bumsub2016porposal}) or end-to-end trained network \cite{zhou2016learning} have been employed with flow algorithms, and shown potential to better align intra-class objects than handcrafted features (e.g., SIFT~\cite{Lowe2004sift}). However, these methods assume locality and smoothness of the flow and thus may fail to align objects under large displacements or non-rigid deformations. 

PatchMatch~\cite{barnes2009patchmatch} relaxes the rigidity assumption, and is a fast randomized algorithm for finding a dense NNF for patches. There are two extensions to handle patch variations in geometry and appearance. The Generalized PatchMatch~\cite{Barnes2010generalpatch} algorithm allows these patches to undergo translations, rotations, and scale changes. NRDC~\cite{haCohen2011nrdc} handles consistent tonal and color changes through iterative matching and refinement of appearance transformations. Recently, a multi-scale patch generazation called ``Needle"~\cite{Lotan2016Needle} has been shown to facilitate reliable matching of degraded images. These approaches are still fundamentally based on low-level features, and as a result, fail to match images that are visually very different but semantically similar. Our proposed technique seeks to address this problem.
\begin{figure}[t]   
\centering
 \footnotesize
\includegraphics[width=0.95\linewidth]{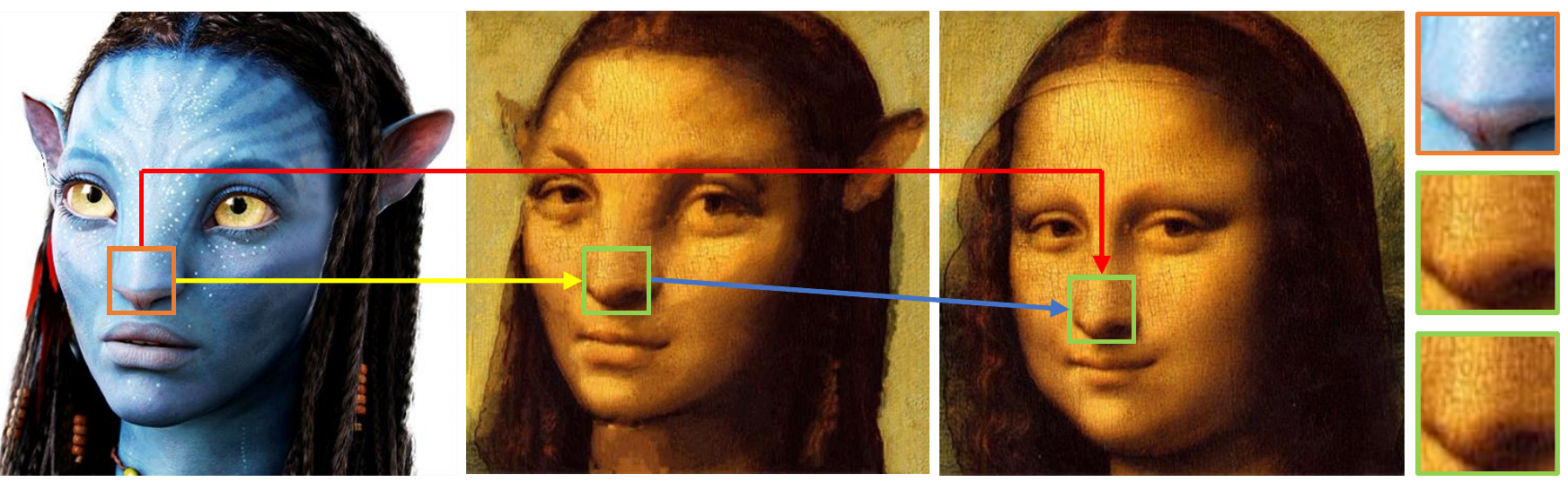}
   \centering

{

\begin{tabular}{p{.25\linewidth}p{.25\linewidth}p{.25\linewidth}p{.06\linewidth}}
\footnotesize
\hspace*{-.2\linewidth}\centering{$A$} & \centering{$A'$}  & \centering{$B'$}&
\end{tabular}
}
	
   \caption{Our method separates the difficult mapping $A \rightarrow B'$(red) into two tractable mappings: in-place mapping $A \rightarrow A'$ (yellow) and similar-appearance mapping $A' \rightarrow B'$ (blue).} 
 \label{fig:map}
\end{figure}

\subsection{Neural Style Transfer}

Our matching approach uses deep features generated by Deep Convolutional Neural Networks (CNN)~\cite{krizhevsky2012imagenet}. It has been shown in high-level image recognition tasks that such deep features are better representations for images~\cite{zeiler2014cnn}. DeepDream~\cite{alexander2015deepdream} is a recent attempt to generate artistic work using a CNN. This inspired work on neutral style transfer~\cite{gatys2016neural}, which successfully applied CNN (pre-trained VGG-16 networks~\cite{simonyan2014very}) to the problem of style transfer, or texture transfer~\cite{gatys2015texture}. 

This method is able to produce more impressive stylization results than traditional texture transfer, since a CNN is effective in decomposing content and style from images. This idea is further extended to portrait painting style transfer~\cite{selim2016} by adding face constraints. The most related work to ours is patch-based style transfer by combining a Markov Random Field (MRF) and a CNN~\cite{chuanli2016mrf}. Yang et al. \shortcite{yang2016high} futher extend this idea to image inpainting. These two works also use patch similarity metric based on CNN features, but it only serves as a data term to optimize pixel values of the output image. They do not explicitly establish the correspondence between two images as our method does. 

Another approach of neural style transfer is to directly learn a feed-forward generator network for a specific style. For example, Johnson et al.~\shortcite{johnson2016perceptual} define a perceptual loss function to help learn a transfer network designed to produce results that match those of Gatys et al.~\cite{gatys2016neural}. Ulyanov et al.~\shortcite{ulyanov2016texture} propose a texture network for both texture synthesis and style transfer. Li and Wand~\shortcite{li2016precomputed} introduce a Markovian Generative Adversarial Network to speed up their previous approach~\cite{chuanli2016mrf}. Chen et al.~\shortcite{dong2017stylebank} and Dumoulin et al.~\shortcite{Dumoulin2016style} further extend the style transfer network from a single style to multiple styles.

Despite major progress, these methods are unable to guarantee that the transferred results are structure-preserving. In addition, they often generate stylization results with texture that is randomly distributed. In contrast, our technique transfers style in a more structure-preserving manner, and this is due to semantic-based dense correspondences.

\begin{figure}[t]
\centering
\footnotesize
    \setlength{\tabcolsep}{0.003\linewidth}
 \begin{tabular}{lcccccc}
    & $relu5\_1$ & $relu4\_1$ & $relu3\_1$ & $relu2\_1$ & $relu1\_1$\\
 $A$ &
   \includegraphics[width=0.182\linewidth]{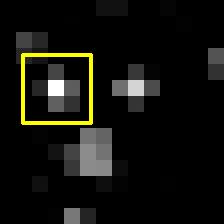}&
   \includegraphics[width=0.182\linewidth]{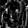}&
  \includegraphics[width=0.182\linewidth]{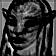}&
   \includegraphics[width=0.182\linewidth]{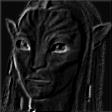}&
   \includegraphics[width=0.182\linewidth]{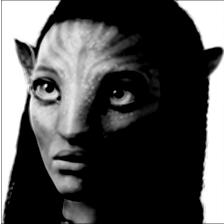}\\

 $B'$ &
   \includegraphics[width=0.182\linewidth]{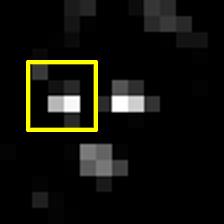}&
   \includegraphics[width=0.182\linewidth]{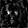}&
  \includegraphics[width=0.182\linewidth]{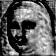}&
   \includegraphics[width=0.182\linewidth]{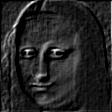}&
   \includegraphics[width=0.182\linewidth]{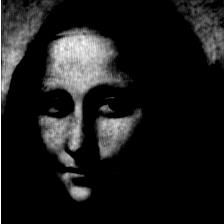}\\

 \end{tabular}

 \caption{The input image $A$ (or $B'$) is encoded by the CNN (e.g. VGG-19) as a set of filtered images at each layer. Here we visualize one representative filtered image for each layer. }

  \label{fig:feature}
\end{figure}

\section{Motivation}
\label{sec:motivation}
Given an image pair $A$ and $B'$, which may differ in appearance but have similar semantic structure, the goal is to find the mapping from $A$ to $B'$ (or from $B'$ to $A$) for visual attribute transfer.
It is assumed that the image pair has different visual attributes.

\emph{\textbf{Analogy with Bidirectional Constraint.}} It is non-trivial to directly match $A$ and $B'$ due to their appearance differences. Our solution is to formulate the mapping as a problem of \emph{image analogies}~\cite{hertzmann2001analogy}: $A : A' :: B : B'$, where $A'$ and $B$ are unknown latent variables. This analogy implies two constraints: 1) $A$ and $A'$ (also $B$ and $B'$) correspond at the same spatial position; 2) $A$ and $B$ (also $A'$ and $B'$) are similar in appearance (color, lighting, texture and etc.). As illustrated in Figure~\ref{fig:map}, the difficult mapping problem from $A$ and $B'$ (red arrow) can be separated into one in-place mapping from $A$ to $A'$ (yellow arrow) and one similar-appearance mapping from $A'$ to $B'$ (blue arrow). The mapping from $B'$ to $A$ is achieved in the same way with the help of $B$.   

These forward and reverse mapping functions between images $A$ and $B'$ are denoted as $\Phi_{a\rightarrow b}$ and $\Phi_{b\rightarrow a}$, respectively. More specifically, $\Phi_{a\rightarrow b}$ maps a pixel at location $p$ from $A$ to $B'$. Because of in-place mappings $A \rightarrow A'$ and $B \rightarrow B'$, $\Phi_{a\rightarrow b}$ is also the mapping from $A$ to $B$, $A'$ to $B'$, and $A'$ to $B$. It is found by imposing the constraints
\begin{equation}\label{ejavascript:void(0);qn:forward_map}
A(p)=B(\Phi_{a\rightarrow b}(p)) \text{ and } A'(p)=B'(\Phi_{a\rightarrow b}(p)).
\end{equation}
$\Phi_{b\rightarrow a}$ is the reverse mapping. To further achieve symmetric and consistent mappings, we consider the bidirectional constraint, namely $\Phi_{b\rightarrow a}(\Phi_{a\rightarrow b}(p)) = p$ and $\Phi_{a\rightarrow b}(\Phi_{b\rightarrow a}(p)) = p$.
The latent images $A'$, $B$ and mapping functions $\Phi_{a\rightarrow b}$, $\Phi_{b\rightarrow a}$ are alternatively optimized in our objective function described in Section~\ref{sec:algo}.

\emph{\textbf{Reconstruction using CNN.}} The recovery of latent images $A'$ and $B$ is crucial in our method. Here we discuss the reconstruction of $A'$; $B$ is recovered in the same way. The ideal $A'$ should comprise the content structure from $A$ and corresponding visual details from $B'$. We solve this problem using an image decomposition and reconstruction hierarchy. We adopt recent CNNs trained on an object recognition dataset, which compresses image information progressively from precise appearance (fine) to actual content (coarse)~\cite{gatys2016neural} (Figure~\ref{fig:feature}). At the coarsest layer, $A$ and $B'$ may have very similar content information for better matching as indicated by yellow rectangles. As a result, we can assume $A'$ to be $A$ at this layer of the CNN. In contrast, this assumption fails in other image decompositions such as the Laplacian pyramid, where $A$ and $B'$ may still differ remarkably in colors at the coarsest layer.

At other layers of the CNN, we selectively extract content structures from features of $A$ and detail information from features of $B'$ by a weighted mask, which helps construct a linearly weighted combination to recover features of $A'$.  We will describe the implementation details in Section~\ref{subsec:reconstruction}. The updated latent images will carry the fusion information to the next layer for further mapping refinement.

\emph{\textbf{Deep PatchMatch.}} Given latent images $A'$ and $B$, inferring $\Phi_{a \rightarrow b}$ and $\Phi_{b \rightarrow a}$ is equivalent to computing a forward NNF and a reverse NNF between $A$ and $B$ as well as between $A'$ and $B'$. More specifically, the forward NNF maps $A$ to $B$; this NNF also maps $A'$ to $B'$. The reverse NNF is similarly defined. PatchMatch~\cite{barnes2009patchmatch} is known as a randomized fast algorithm for computing approximate NNF between two images. Good patch matches can be found through random sampling, and spatial coherence in the imagery allows us to propagate such matches quickly to surrounding areas. Instead, we consider PatchMatch in a deep feature domain, which can provide better representations for semantic-level correspondences and be able to be incorporated into our CNN-based latent image reconstruction.

\begin{figure}[t] \centering
 \footnotesize
   \includegraphics[width=0.95\linewidth]{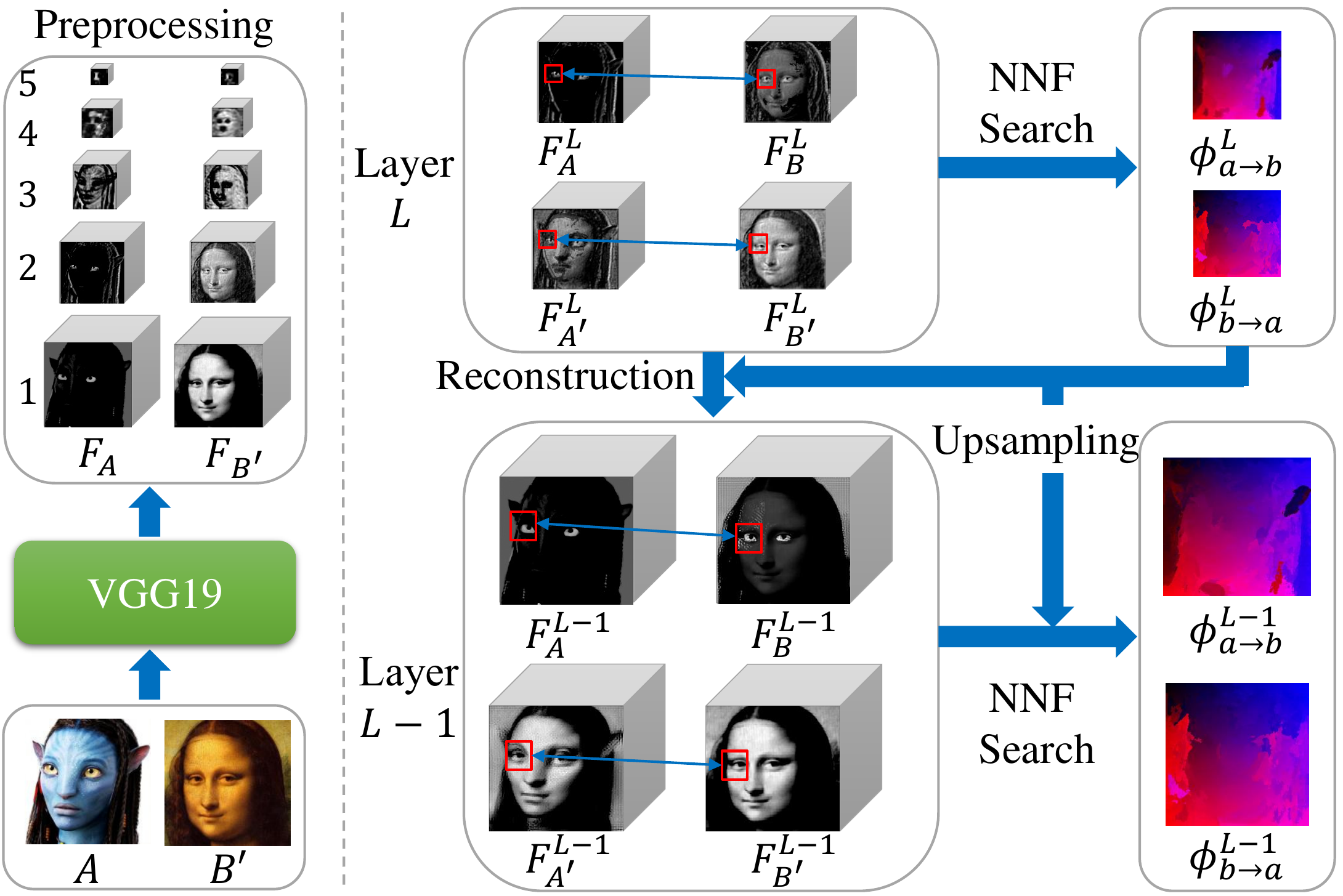}
	
   \caption{System pipeline. }

   \label{fig:pipeline}
   \end{figure}
   
\section{Deep Image Analogy}
\label{sec:algo}

We achieve visual attribute transfer through image analogy and deep CNN features; an overall process we refer to as \emph{deep image analogy}. \Fref{fig:pipeline} shows the system pipeline. We first compute deep features for the input image pair $A/B'$ through pre-trained CNN, and initialize feature maps of two latent images $A'/B$ at the coarsest CNN layer in the preprocessing step (in~\Sref{subsec:preprocessing}). Then, at each layer, we compute a forward NNF and a reverse NNF that establish correspondences between feature maps of $A$ and $B$ as well as between feature maps of $A'$ and $B'$. (\Sref{subsec:nnsearch}). 
The extracted NNFs together with feature maps are used to reconstruct features of latent images $A'/B$ at the next CNN layer (in~\Sref{subsec:reconstruction}). The NNFs obtained at the current layer are further upsampled to the next layer as their initialization (in~\Sref{subsec:nnupsample}). These three steps (NNF search, latent image reconstruction and NNF upsampling) are repeated at each layer, updating correspondences from coarse to fine.

\subsection{Preprocessing}\label{subsec:preprocessing}

Our algorithm starts with precomputing feature maps by a VGG-19 network~\cite{simonyan2014very} that is trained on the ImageNet database~\cite{russakovsky2015imagenet} for object recognition. We obtain the pyramid of feature maps $\{F_{A}^L\}$ and $\{F_{B'}^L\}$ ($L = 1...5$) for the input images $A$ and $B'$. The feature map of each layer is extracted from the {\textit{reluL\_}1} layer. It is a 3D tensor with $width \times heights \times channel$, and its spatial resolution increases from $L=5$ to $1$, as shown on \Fref{fig:pipeline}(\emph{left}). 

The features of $A'$ and $B$ are unknown. We estimate them in a coarse-to-fine manner, which needs a good initialization at the coarsest layer ($L = 5$). Here, we let $F_{A'}^5=F_{A}^5$ and $F_{B}^5= F_{B'}^5$ initially, that is, we view $A$ and $A'$ (also $B$ and $B'$) to be very similar at the top layer where the images have been transformed by the CNN into representations with the actual content, but being invariant to precise appearance (as shown in \fref{fig:feature}).

\subsection{Nearest-neighbor Field Search}\label{subsec:nnsearch}

At layer $L$, we estimate a forward NNF and a reverse NNF; they are represented by mapping functions $\phi^L_{a\rightarrow b}$ and $\phi^L_{b\rightarrow a}$, respectively. $\phi^L_{a\rightarrow b}$ maps a point in feature map $F_{A}^L$ to another in feature map $F_{B}^L$. Note that $\phi^L_{a\rightarrow b}$ also maps $F_{A'}^L$ to $F_{B'}^L$. $\phi^L_{b\rightarrow a}$ is similarly defined in the reverse direction.
$\phi^L_{a\rightarrow b}$ is computed by minimizing the following energy function:
\begin{equation}\label{eqn:nnsearch}
\begin{split}
\phi _{a \to b}^L(p) = \mathop {\arg \min }\limits_q \sum\limits_{x \in N(p)
, y \in N(q)}(|| {\overline F_{A}^L(x)}- {\overline F_{B}^L(y)}||^2 \\
+ ||{\overline F_{A'}^L(x)} - {\overline F_{B'}^L(y)}||^2),
\end{split}
\end{equation}
where $N(p)$ is the patch around $p$. We set the patch size to be $3 \times 3$ when $L=5,4,3$, and $5 \times 5$ when $L=2,1$. For each patch around pixel $p$ in the source $A$ (or $A'$), we find its nearest neighbor position $q = \phi^L_{a\rightarrow b}(p)$ in the target $B$ (or $B'$). The mapping function $\phi^L_{b\rightarrow a}(p)$ is similarly computed.

$F(x)$ in \Eref{eqn:nnsearch} is a vector that represents all channels of the $L$-th feature layer at position $x$. We use normalized features $\overline F^L(x) = \frac{F^L(x)}{|F^L(x)|}$ in our patch similarity metric, because relative values are more meaningful than absolute values in networks.

\begin{figure}[t]
\footnotesize
    \setlength{\tabcolsep}{0.003\linewidth}
 \begin{tabular}{cccc}
    \includegraphics[width=0.24\linewidth]{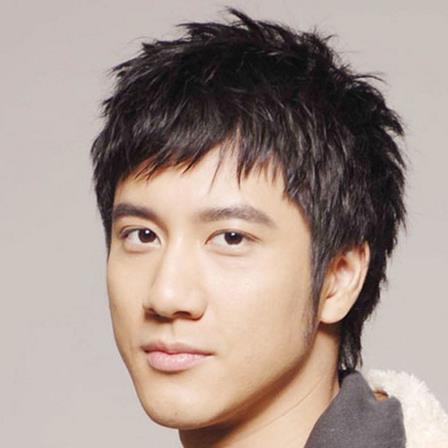}&
   \includegraphics[width=0.24\linewidth]{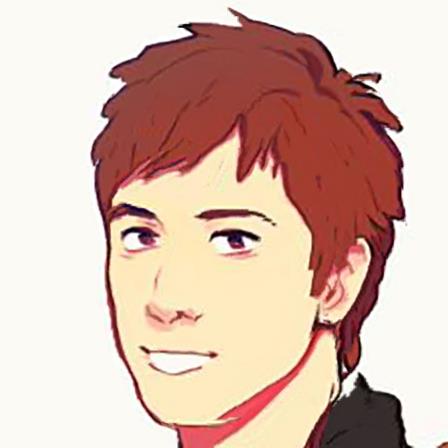}&
  \includegraphics[width=0.24\linewidth]{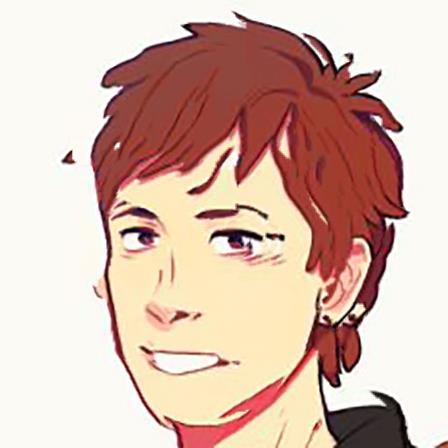}&
   \includegraphics[width=0.24\linewidth]{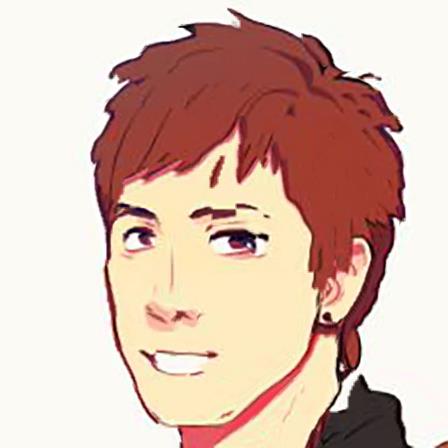}\\
   \includegraphics[width=0.24\linewidth]{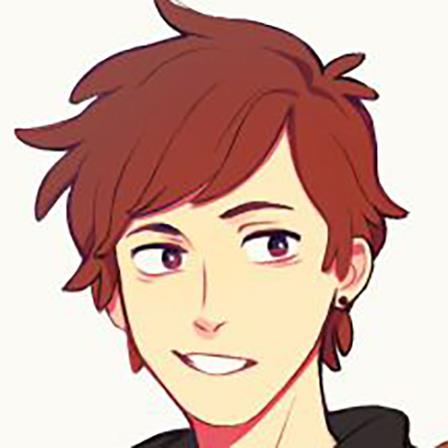}&
   \includegraphics[width=0.24\linewidth]{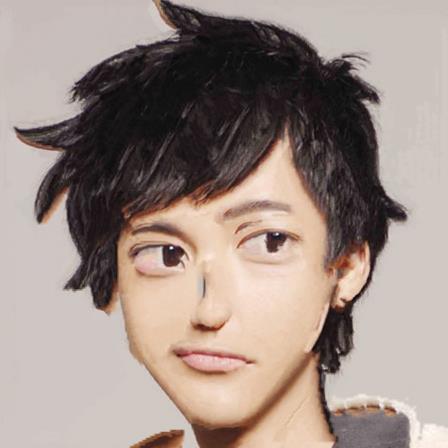}&
  \includegraphics[width=0.24\linewidth]{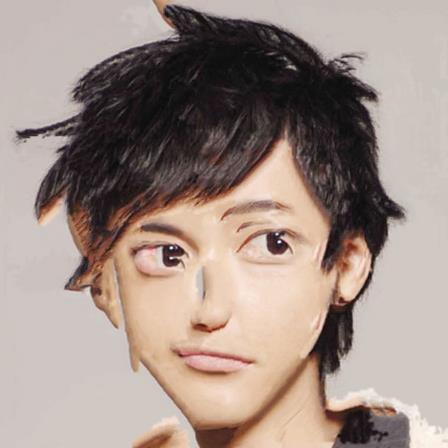}&
   \includegraphics[width=0.24\linewidth]{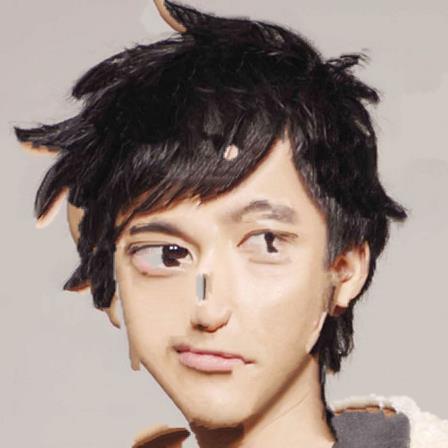}\\
(a) & (b) & (c) & (d)
 \end{tabular}

 \caption{Benefits of bidirectional constraint and deconvolution. (a)input images (b) bidirectional + deconvolution (c) single-direction + deconvolution (d) bidirectional + resampling. }

  \label{fig:Single}
\end{figure}

\Eref{eqn:nnsearch} does not require a regularization term because local smoothness is implicitly achieved through aggregation of overlapping patches. Such a unary-only energy formulation can be efficiently optimized with the PatchMatch method~\cite{barnes2009patchmatch}. We adapt this approach of random search and propagation to support two pairs of multi-channel feature maps.

Our NNF search considers the constraints imposed by forward and reverse NNFs in the following manner. As indicated in \Eref{eqn:nnsearch}, the estimation of $\phi^L_{a\rightarrow b}$ relies on four feature maps. Among them, the reconstruction of $F_{B}^{L}$ depends on the reverse mapping at previous layer $\phi^{L+1}_{b\rightarrow a}$ which will be described in Section~\ref{subsec:reconstruction}. In other words, $\phi^L_{a\rightarrow b}$ is constrained by $\phi^{L+1}_{b\rightarrow a}$ through $F_{B}^{L}$; symmetrically $\phi^L_{b\rightarrow a}$ is constrained by $\phi^{L+1}_{a\rightarrow b}$ through $F_{A'}^{L}$. With these constraints, $\phi^L_{a\rightarrow b}$ and $\phi^L_{b\rightarrow a}$  usually will agree, and thus discourages ambiguous \textit{1-to-n} mapping. In contrast, if we consider only single-direction mapping, \Eref{eqn:nnsearch} becomes:
\begin{equation}\label{eqn:nnsearch2}
\phi _{a \to b}^L(p) = \mathop {\arg \min }\limits_q \sum\limits_{x \in N(p), y \in N(q)}|| {\overline F_{A'}^L(x)}- {\overline F_{B'}^L(y)}||^2
\end{equation}
for optimizing the mapping $\phi^L_{a\rightarrow b}$. Unfortunately, it is prone to misalignment. We show a comparison in Figure~\ref{fig:Single}(b)(c).

At every layer, we have to recover features of $A'$ and $B$ before NNF search. In this section, we will mainly discuss the reconstruction of $A'$; $B$ can be estimated likewise. As mentioned in Section~\ref{sec:motivation}, the reconstruction of $F^{L-1}_{A'}$ at layer $L-1$ is the fusion of content from $F^{L-1}_{A}$ and details from $F^{L-1}_{B'}$. Let us define the feature map $R^{L-1}_{B'}$ to be the modified version of $F^{L-1}_{B'}$ to fit the structure of $A$. $F^{L-1}_{A'}$ is computed using a weighted sum:
\begin{equation}\label{eqn:reconstruction}
F_{A'}^{L-1}=F_{A}^{L-1}\circ W^{L-1}_A+ R_{B'}^{L-1} \circ (1-W^{L-1}_A),
\end{equation}
where $\circ$ is element-wise multiplication on each channel of feature map, and $W^{L-1}_A$ is a 2D weight map (with the elements ranging from $0$ to $1$) used to separate content structures from details. As mentioned in Section~\ref{subsec:preprocessing}, $F^{L-1}_{A}$ is pre-computed. Next, we will introduce how to compute $R_{B'}^{L-1}$ and $W^{L-1}_A$ respectively.

Ideally, the $R^{L-1}_{B'}$ should be equal to the warped $F_{B'}^{L-1}$ with $\phi^{L-1}_{a \rightarrow b}$: $R_{B'}^{L-1}=F_{B'}^{L-1}(\phi^{L-1}_{a \rightarrow b})$. However, $\phi^{L-1}_{a \rightarrow b}$ is initially unknown at layer $L-1$. A na\"{\i}ve method is to directly upscale $\phi^L_{a \rightarrow b}$ to the dimension at layer $L-1$, and then warp $F_{B'}^{L-1}$ with it. Unfortunately, this method cannot preserve mapped structures from previous layers (in Figure~\ref{fig:Single}(d)). This is because spatial correspondence between two adjacent layers cannot be exactly maintained due to non-linear modules in CNNs (i.e., ReLU, Max-Pooling).

To address this problem, we present a new approach, which first warps the feature maps at the previous layer before deconvolving the warped features for the current layer. Specifically, we warp $F^L_{B'}$ by using $\phi^L_{a \rightarrow b}$, getting $F^L_{B'}(\phi^L_{a \rightarrow b})$. Let the sub-net of CNN including all computing units between layer $L-1$ and $L$ be denoted as $\textsc{CNN}_{L - 1}^L(\cdot)$. Our objective is to make the output of $\textsc{CNN}_{L - 1}^L(R_{B'}^{L-1})$ be as close as possible to the target features $F^L_{B'}(\phi^L_{a \rightarrow b})$. Hence, $R_{B'}^{L-1}$, which is randomized initially, can be solved by minimizing the following loss function:
\begin{equation}
\mathcal{L}_{R_{B'}^{L-1}} = || {\textsc{CNN}_{L - 1}^L(R_{B'}^{L-1}) - F^L_{B'}(\phi^L_{a \rightarrow b})} ||^2.
\label{eqn:deconv}
\end{equation}
The gradient with respect to the feature values $\partial \mathcal{L}_{R_{B'}^{L-1}}/\partial R_{B'}^{L-1}$ can be used as input for some numerical optimization strategy. Here we use gradient descent (L-BFGS)~\cite{Zhu1994BFGS}. The optimization for feature deconvolution is similar to update the stylization result by pre-trained CNN, described in~\cite{gatys2016neural}.

\subsection{Latent Image Reconstruction}\label{subsec:reconstruction}

\begin{figure}[t]
\footnotesize

 \includegraphics[width=0.95\linewidth]{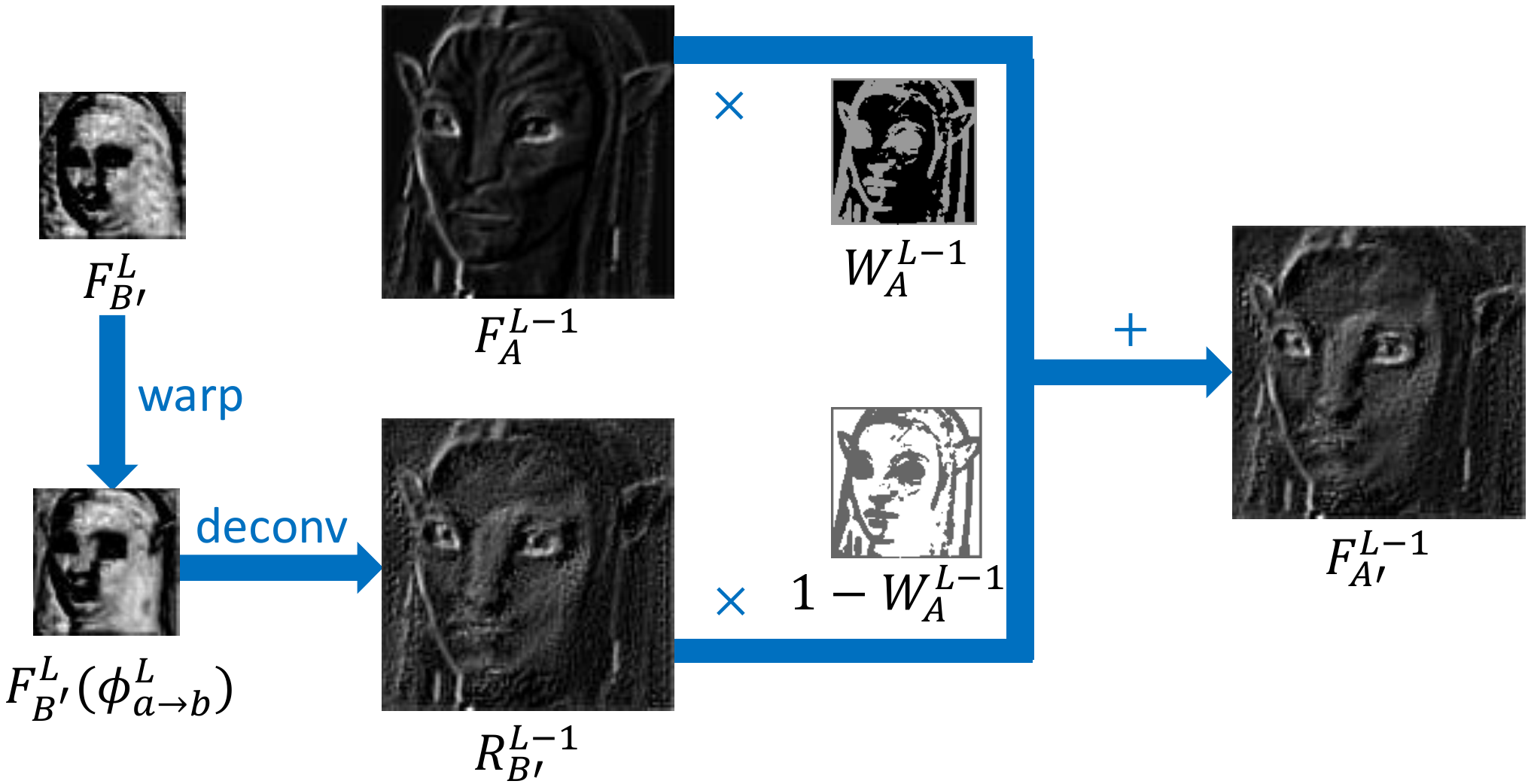}

 \caption{Recovering features of latent image $A'$ by weighted combination of the structures from $A$ and the visual details sampled from $B'$. }
  \label{fig:combine}
    
\end{figure}

\begin{figure}[t]
\footnotesize
    \setlength{\tabcolsep}{0.003\linewidth}
 \begin{tabular}{ccccc}
    \includegraphics[width=0.105\linewidth]{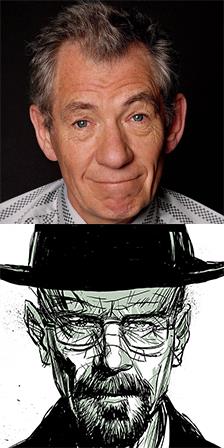}&
   \includegraphics[width=0.21\linewidth]{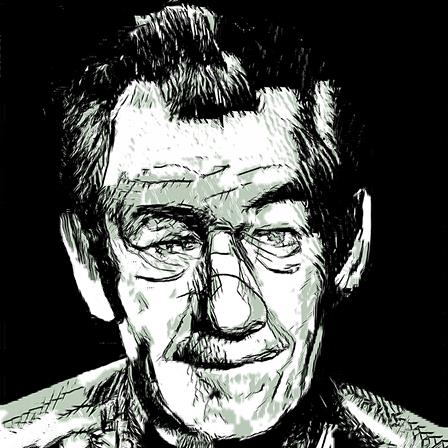}&
  \includegraphics[width=0.21\linewidth]{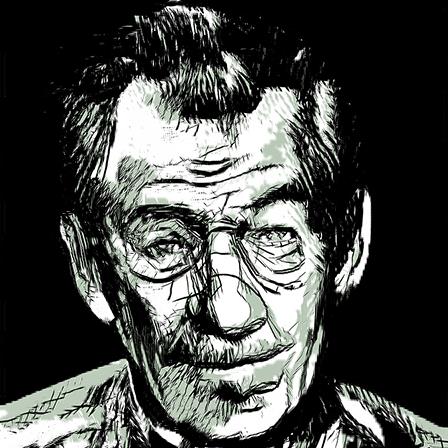}&
   \includegraphics[width=0.21\linewidth]{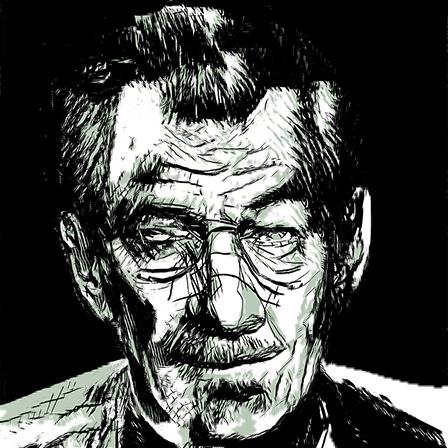} &
   \includegraphics[width=0.21\linewidth]{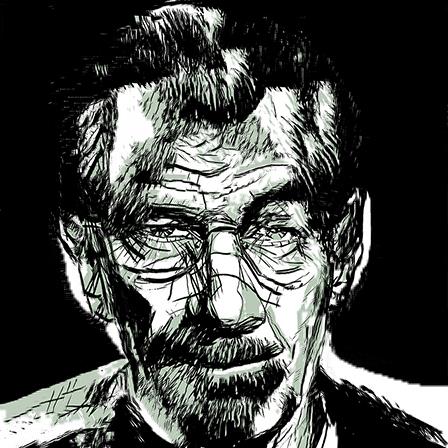}
  \\
  Inputs&$+0.2$&$+0.1$&$+0.0$&$-0.1$\\
 \end{tabular}

 \caption{Effect of different global offsets for weights $\{\alpha^L\}_{L=1}^{4}$.}
  \label{fig:weights}

\end{figure}

The role of the 2D weight map $W^{L-1}_A$ is to separate content structures from details. Multiplying $F^{L-1}_{A}$ (or $R_{B'}^{L-1}$) with $W^{L-1}_A$ (or $1 - W^{L-1}_A$) helps extract content information from $F^{L-1}_{A}$ (or details from $R_{B'}^{L-1}$). All feature channels share the same 2D weight map, which can be further be represented by
\begin{equation}
W^{L-1}_A = \alpha_{L-1} M_{A}^{L-1},
\label{eqn:featurefusion}
\end{equation}
where $M_{A}^{L-1}$ is a function that specifies the magnitudes of neuron responses at layer L-1, in order to preserve content structures from $A$ when they are present. We apply a sigmoid function to get $M_{A}^{L-1}(x)=\frac{1}{1+exp({-\kappa\times(|F_{A}^{L-1}(x)|^2-\tau)})}$, with $\kappa = 300, \tau = 0.05$,
and $|F_{A}^{L-1}(x)|^2$ being normalized to $[0, 1]$. In the reverse direction, $M_{B'}^{L-1}$ is computed using $F_{B'}^{L-1}$ instead of $F_{A}^{L-1}$.

The scalar $\alpha^{L-1}$ (in \Eref{eqn:featurefusion}) controls the trade-off between content and attribute such as style, which is similar to~\cite{gatys2016neural}. When $\alpha^{L-1}\rightarrow 1$, it means that $A'$ should be exactly consistent with $A$ in structure. Reducing $\alpha^{L-1}$ would lead to structure variations in $A'$ compared to $A$.  The coarse layer represents more structure information than the fine layer due to higher-level abstraction, so we gradually decrease $\alpha_{L-1}$ from coarse to fine layer. For all of our visual attribute transfer results, the default setting is $\{\alpha^L\}_{L=4, 3, 2, 1}=\{0.8,0.7,0.6,0.1\}$. We only vary the default settings using two global offsets, namely $\{\alpha^L\}_{L=1}^{4} \pm 0.1$. Figure~\ref{fig:weights} shows the effect of changing the global offset. Larger weights tend to produce results with more similar structure to inputs ($A$ or $B'$).

In summary, latent image reconstruction involves feature map warp at the current layer, deconvolution for the next layer, and fusion, as illustrated in Fig \ref{fig:combine}.

\begin{figure}[t]
\centering
\footnotesize
    \setlength{\tabcolsep}{0.003\linewidth}
 \begin{tabular}{m{0.155\linewidth}cccccc}
    &  $L=5$ & $L=4$ & $L=3$ & $L=2$ & $L=1$\\
  $\phi^L_{a\rightarrow b}$ &
   \includegraphics[width=0.155\linewidth]{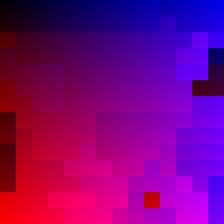}&
   \includegraphics[width=0.155\linewidth]{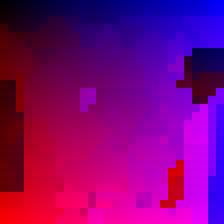}&
  \includegraphics[width=0.155\linewidth]{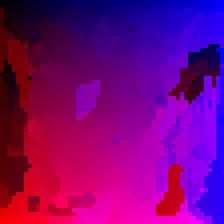}&
   \includegraphics[width=0.155\linewidth]{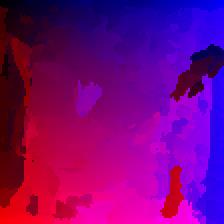}&
   \includegraphics[width=0.155\linewidth]{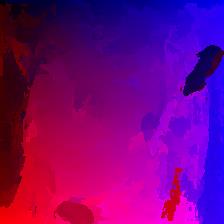}\\
    $B'(\phi^L_{a\rightarrow b})$  &
   \includegraphics[width=0.155\linewidth]{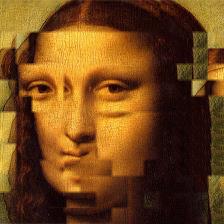}&
   \includegraphics[width=0.155\linewidth]{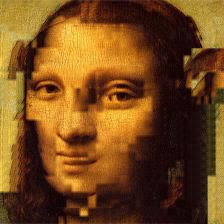}&
  \includegraphics[width=0.155\linewidth]{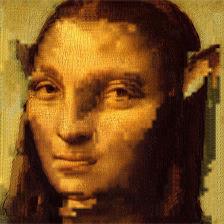}&
   \includegraphics[width=0.155\linewidth]{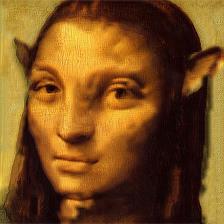}&
   \includegraphics[width=0.155\linewidth]{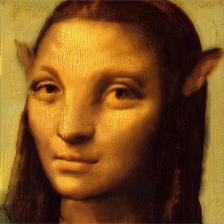}\\
    \vspace{-0.2in}$B'(\phi^L_{a\rightarrow b})$ (layer independent)&
   \includegraphics[width=0.155\linewidth]{images/pipeline/resultSC_0.jpg}&
   \includegraphics[width=0.155\linewidth]{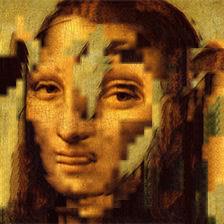}&
  \includegraphics[width=0.155\linewidth]{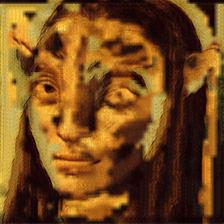}&
   \includegraphics[width=0.155\linewidth]{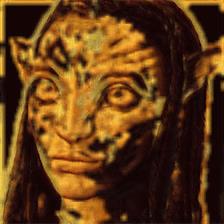}&
   \includegraphics[width=0.155\linewidth]{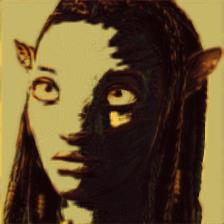}\\\\   
$\phi^L_{b\rightarrow a}$ &
   \includegraphics[width=0.155\linewidth]{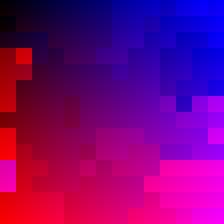}&
   \includegraphics[width=0.155\linewidth]{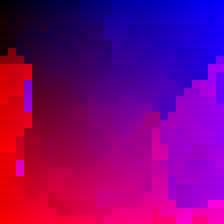}&
  \includegraphics[width=0.155\linewidth]{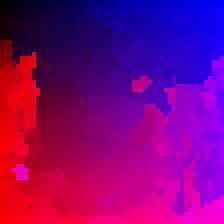}&
   \includegraphics[width=0.155\linewidth]{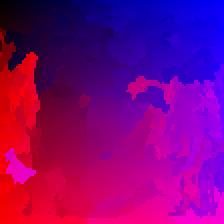}&
   \includegraphics[width=0.155\linewidth]{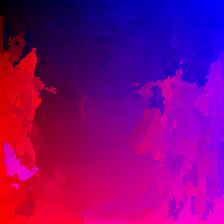}\\
$A(\phi^L_{b\rightarrow a})$ &
   \includegraphics[width=0.155\linewidth]{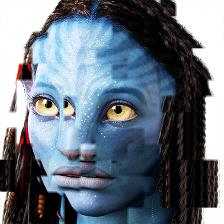}&
   \includegraphics[width=0.155\linewidth]{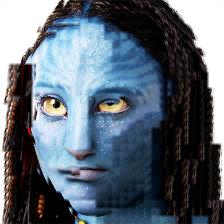}&
  \includegraphics[width=0.155\linewidth]{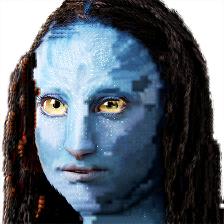}&
   \includegraphics[width=0.155\linewidth]{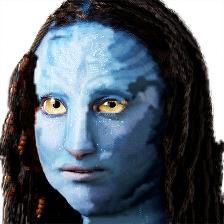}&
   \includegraphics[width=0.155\linewidth]{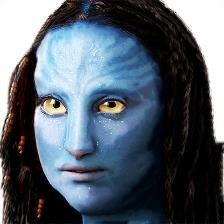}\\
   \vspace{-0.2in}$A(\phi^L_{b\rightarrow a})$ (layer independent)&
   \includegraphics[width=0.155\linewidth]{images/pipeline/resultCS_0.jpg}&
   \includegraphics[width=0.155\linewidth]{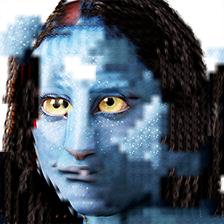}&
  \includegraphics[width=0.155\linewidth]{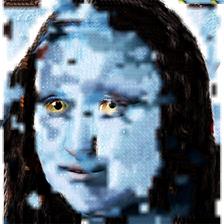}&
   \includegraphics[width=0.155\linewidth]{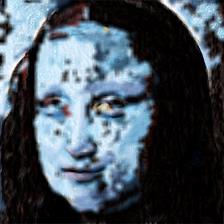}&
   \includegraphics[width=0.155\linewidth]{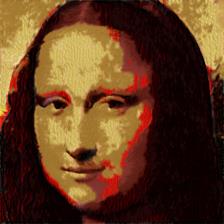}\\

 \end{tabular}

 \caption{Visualization of NNFs (top row), reconstructed results with NNFs (middle row) and reconstructed results based on layer-independent NNFs (bottom row) of each layer.}
  
  \label{fig:NNF}
\end{figure}

\subsection{Nearest-neighbor Field Upsampling}\label{subsec:nnupsample}

Our NNFs are computed in a coarse-to-fine manner. At the coarsest layer, the mappings $\phi^5_{a\rightarrow b}$ and $\phi^5_{b\rightarrow a}$ are randomly initialized ($L=5$). For other layers, as an initial guess, we upsample $\phi^L_{a \rightarrow b}$, $\phi^L_{b \rightarrow a}$, to layer $L-1$. Since the NNF obtained by PatchMatch is only piece-wise smooth, we use nearest-neighbor interpolation instead of linear interpolation.

The upscaled NNFs only serve as initial guesses, and they are further refined by NNF search (described in~\Sref{subsec:nnsearch}). Additionally, the initial guess serves as guidance to limit the random search space at every layer. A similar scheme is adopted in the extension of PatchMatch algorithm~\cite{Hu2016patchmatch}. In our work, the search space is limited to be the receptive field of the network at each layer. For VGG-19, the random search radii of layers $\{4,3,2,1\}$ are $\{6,6,4,4\}$, respectively. Figure~\ref{fig:NNF}(top rows) shows how our mappings are gradually optimized from coarse to fine. Compared with the layer-independent matching results (bottom rows of Figure~\ref{fig:NNF}), our hierarchical matching scheme successfully propagates the correspondences from coarse levels to fine levels where matches are ambiguous, as shown in Figure~\ref{fig:NNF}(middle rows).

\subsection{Output}
After we obtain the final NNFs $\phi^1_{a\rightarrow b}$ (also $\phi^1_{b\rightarrow a}$) at the lowest feature layer, we let the pixel-location mapping functions $\Phi_{a\rightarrow b}$ and $\Phi_{b\rightarrow a}$ equal to $\phi^1_{a\rightarrow b}$ and $\phi^1_{b\rightarrow a}$ respectively, since the features at the lowest layer have the same spatial dimension as the input images. We then reconstruct $A'$ by patch aggregation in the pixel layer of image: $A'(p)=\frac{1}{n}\sum \limits_{x \in N(p)}(B'(\Phi_{a\rightarrow b}(x))$, where $n= 5 \times 5$ is the size of patch $N(p)$. $B$ is reconstructed in a similar way.

\begin{algorithm}[ht]
\DontPrintSemicolon
\SetAlgoLined
\SetKwInOut{Input}{Input}\SetKwInOut{Output}{Output}
\Input{Two RGB images $A$ and $B'$.}
\Output{Two pixel-location mapping functions: $\Phi_{a\rightarrow b}$, $\Phi_{b\rightarrow a}$; and two RGB images $A'$, $B$. }
\BlankLine
 \textbf{Preprocessing} (\Sref{subsec:preprocessing}):\\
 \quad$\{F_{A}^{L}\}_{L=1}^5, \{F_{B'}^{L}\}_{L=1}^5\leftarrow$ feed $A, B'$ to VGG-19 and get features.\\
 \quad$F_{A'}^5=F_{A}^5$, $F_{B}^5= F_{B'}^5$, and randomize mapping function $\phi^5_{a\rightarrow b}$, $\phi^5_{b\rightarrow a}$.\\
\For{$L =5$ to 1}{
   \textbf{NNF search} (\Sref{subsec:nnsearch}):\\
   \quad$\phi^L_{a\rightarrow b}\leftarrow$ map $F_{A}^L$ to $F_{B}^L$, $F_{A'}^L$ to $F_{B'}^L$.\\
    \quad$\phi^L_{b\rightarrow a}\leftarrow$ map $F_{B}^L$ to $F_{A}^L$, $F_{B'}^L$ to $F_{A'}^L$.\\
    \If{$L>1$}
    { \textbf{Reconstruction} (\Sref{subsec:reconstruction}):\\
     \quad Warp $F_{B'}^{L}$ with $\phi^L_{a\rightarrow b}$ to $F_{B'}^{L}(\phi^L_{a\rightarrow b})$.\\
    \quad Deconvolve $R_{B'}^{L-1}$ with $F_{B'}^{L}(\phi^L_{a\rightarrow b})$ and $\textsc{CNN}_{L - 1}^L(\cdot)$.\\
    \quad $F_{A'}^{L-1}\leftarrow$ weighted blend $F_{A}^{L-1}$ and $R_{B'}^{L-1}$.\\
    \quad Warp $F_{A}^{L}$ with $\phi^L_{b\rightarrow a}$ to $F_{A}^{L}(\phi^L_{b\rightarrow a})$\\
    \quad Deconvolve $R_{A}^{L-1}$ with $F_{A}^{L}(\phi^L_{b\rightarrow a})$ and $\textsc{CNN}_{L - 1}^L(\cdot)$.\\
    \quad $F_{B}^{L-1}\leftarrow$ weighted blend $F_{B'}^{L-1}$ and $R_{A}^{L-1}$.\\
    \textbf{NNF upsampling} (\Sref{subsec:nnupsample}):\\
    \quad Upsample $\phi^L_{a\rightarrow b}$ to $\phi^{L-1}_{a\rightarrow b}$ \\
    \quad Upsample $\phi^L_{b\rightarrow a}$ to $\phi^{L-1}_{b\rightarrow a}$ \\
    }
}
$\Phi_{a\rightarrow b}$= $\phi^1_{a\rightarrow b}$,  $\Phi_{b\rightarrow a}$= $\phi^1_{b\rightarrow a}$ \\
$A'(p)=\frac{1}{n}\sum \limits_{x \in N(p)}(B'(\Phi_{a\rightarrow b}(x))$,  $B(p)=\frac{1}{n}\sum \limits_{x \in N(p)}(A(\Phi_{b\rightarrow a}(x))$\\

\caption{Deep Analogy algorithm}
\label{ag:bda}
\end{algorithm}

\subsection{Algorithm and Performance}

The pseudo code of our implementation is listed in Algorithm~\ref{ag:bda}. Our core algorithm is developed in CUDA. All our experiments are conducted on a PC with an Intel E5 2.6GHz CPU and an NVIDIA Tesla K40m GPU. The runtime of each module is based on the input image with resolution $448 \times 448$. There are two bottlenecks. One is deep PatchMatch ($\sim 40$ seconds), which needs to compute patch similarities on hundreds of feature channels. Another is feature deconvolution ($\sim 120$ seconds), which may require hundreds of iterations to converge for~\Eref{eqn:deconv}.

\begin{figure*}[ht] \centering
  { \footnotesize
    \setlength{\tabcolsep}{0.003\linewidth}
    \begin{tabular}{cccccc}

      \includegraphics[width=0.16\linewidth]{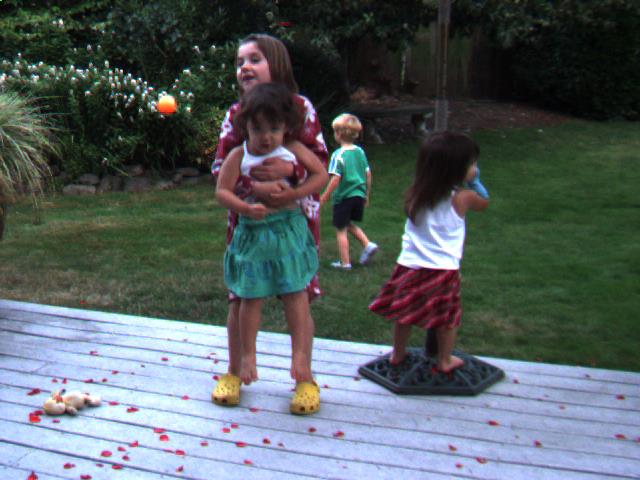} &
      \includegraphics[width=0.16\linewidth]{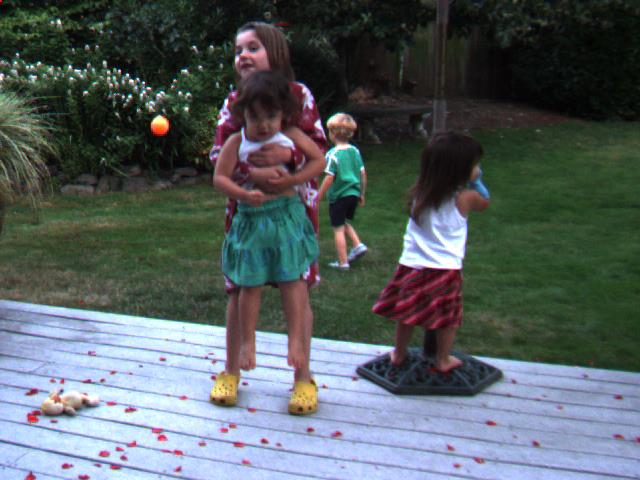} &
      \includegraphics[width=0.16\linewidth]{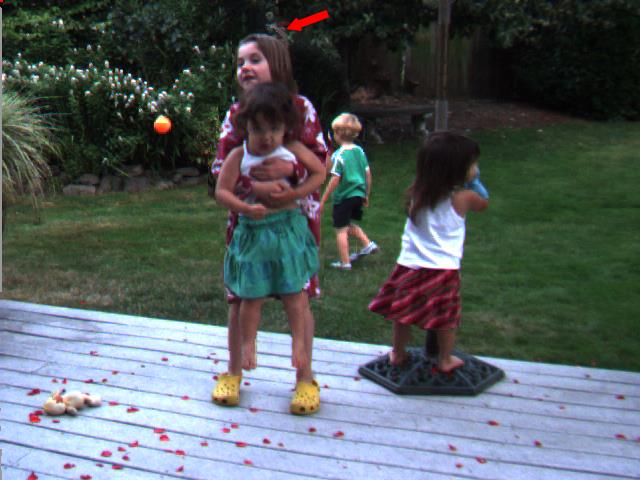}&
       \includegraphics[width=0.16\linewidth]{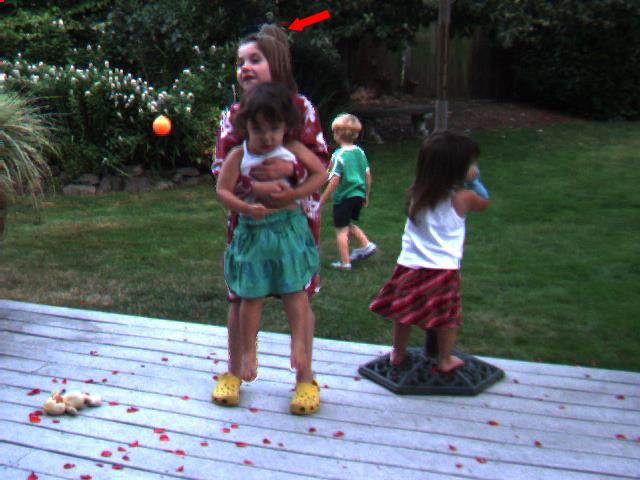} &
      \includegraphics[width=0.16\linewidth]{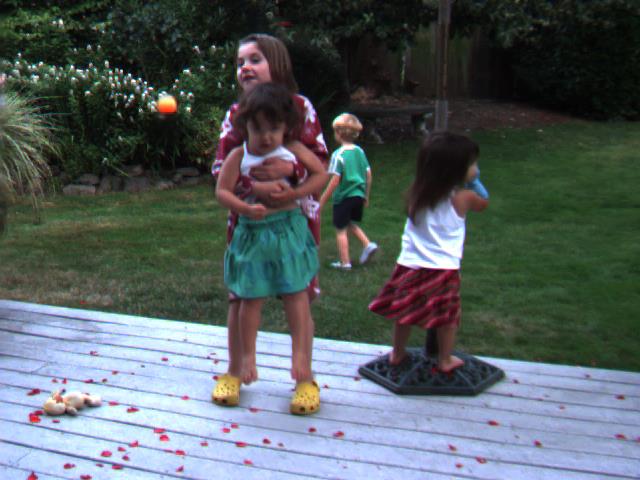} &
      \includegraphics[width=0.16\linewidth]{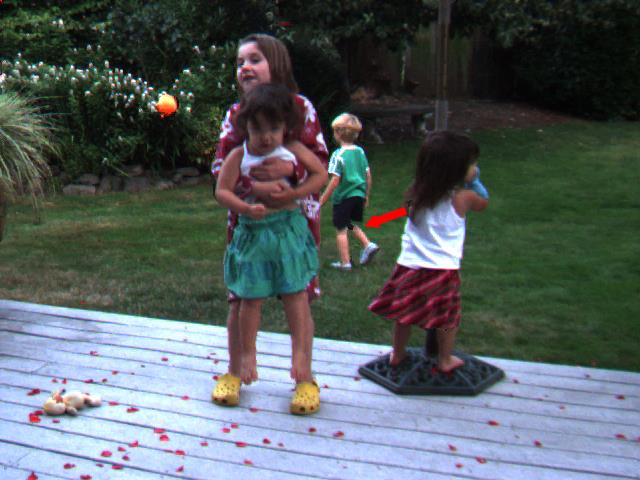}
     \\
     (a) Input (src) & (b) Input (ref) & (c) SIFT flow & (d) DeepFlow2 & (e) PatchMatch & (f) Ours
      \end{tabular}
  }%

  \caption{Comparison of different dense correspondences methods on input pairs with the same scene but slightly different views or motions. Some reconstruction errors are indicated by the red arrows. }
  \label{fig:S1}
\end{figure*}

{\begin{figure*}[ht] \centering
  { \footnotesize
    \setlength{\tabcolsep}{0.003\linewidth}
    \begin{tabular}{cccccccc}
      \includegraphics[width=0.118\linewidth]{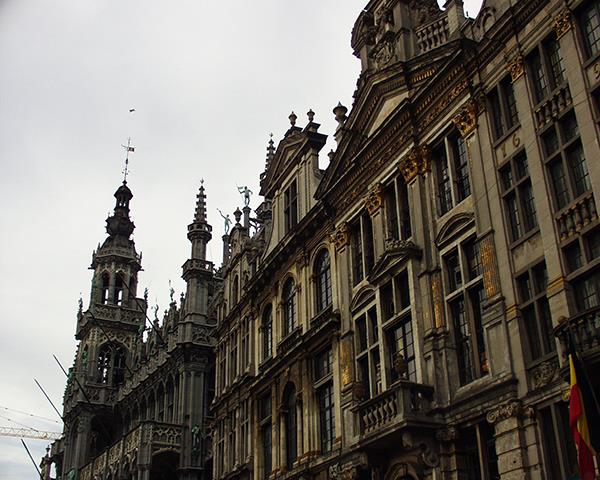} &
            \includegraphics[width=0.118\linewidth]{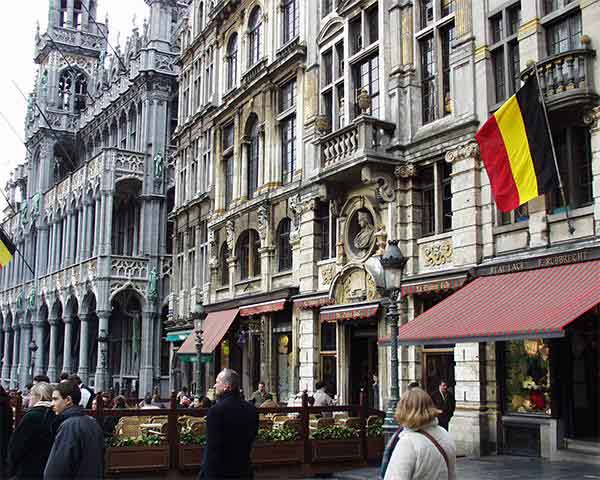} &
   \includegraphics[width=0.118\linewidth]{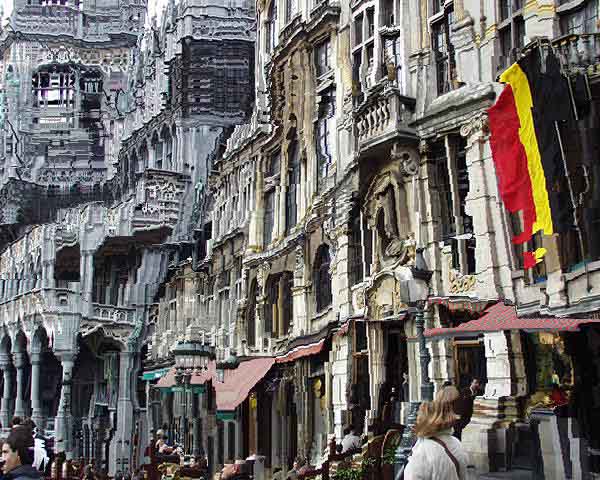} &
\includegraphics[width=0.118\linewidth]{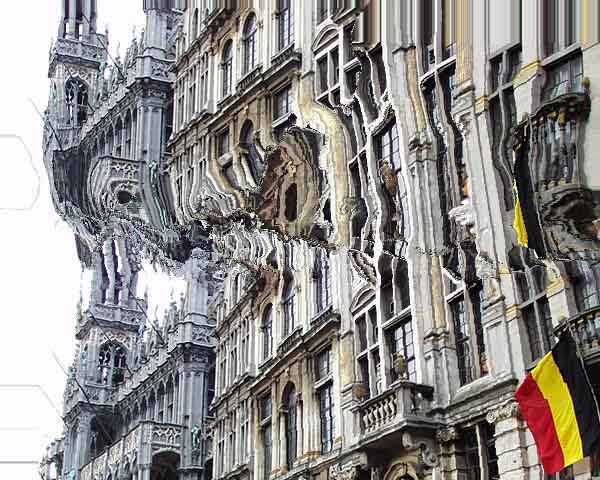} &
 \includegraphics[width=0.118\linewidth]{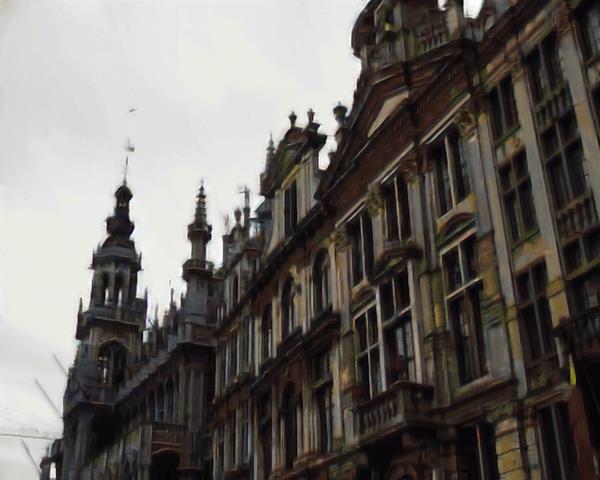} &
  \includegraphics[width=0.118\linewidth]{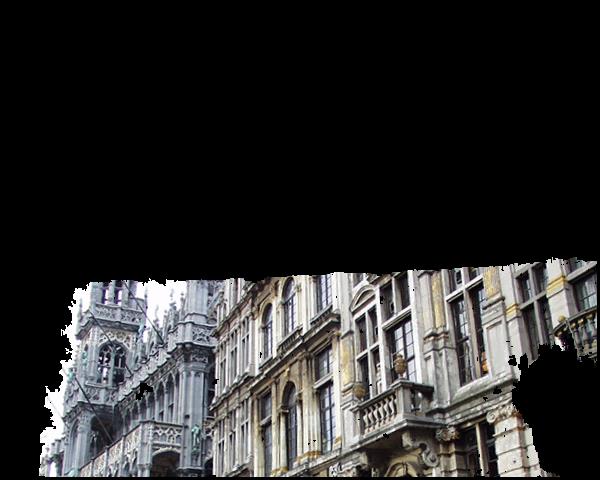} &
  \includegraphics[width=0.118\linewidth]{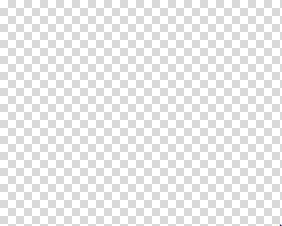}
       &
              \includegraphics[width=0.118\linewidth]{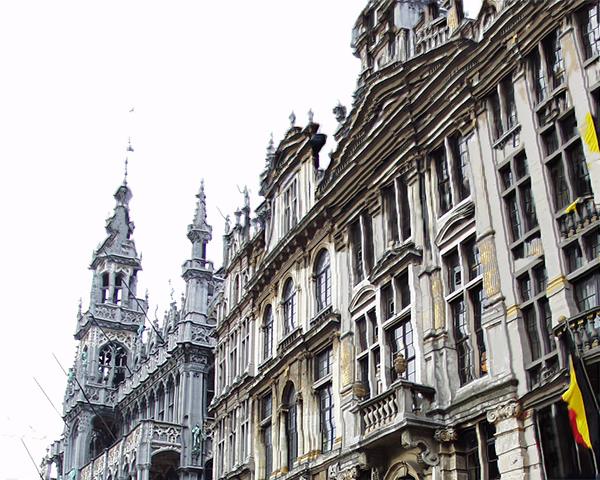} \\
                    \includegraphics[width=0.118\linewidth]{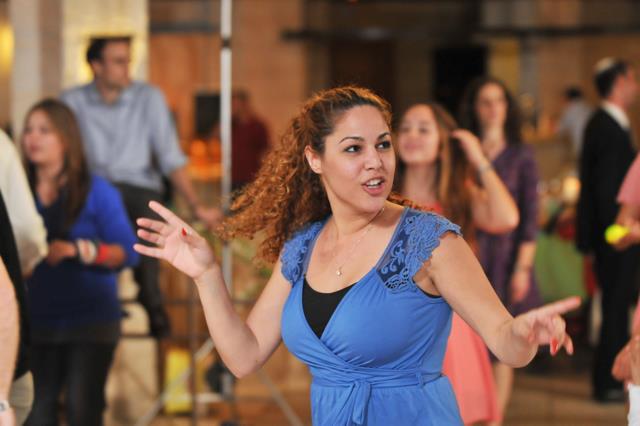} &
            \includegraphics[width=0.118\linewidth]{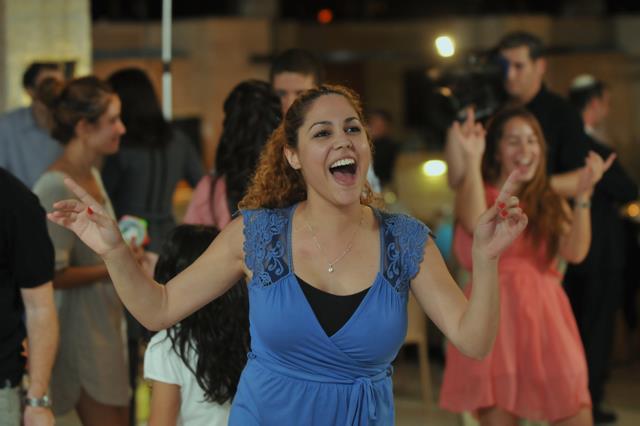} &
   \includegraphics[width=0.118\linewidth]{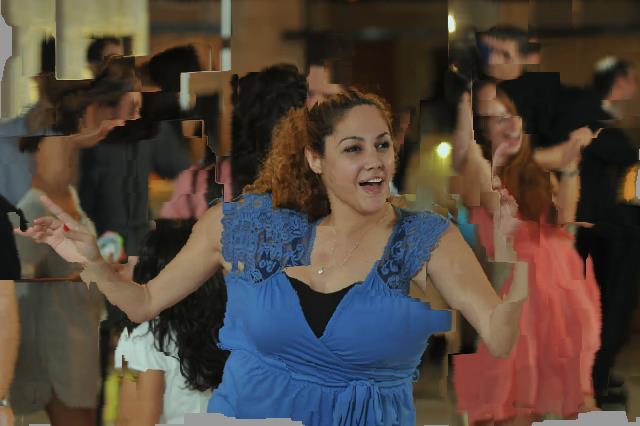} &
\includegraphics[width=0.118\linewidth]{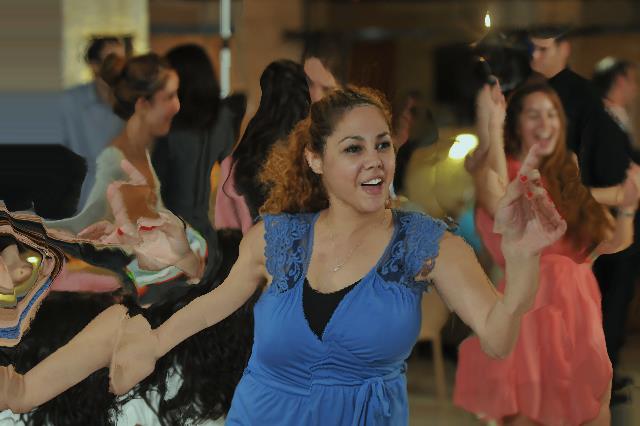} &
 \includegraphics[width=0.118\linewidth]{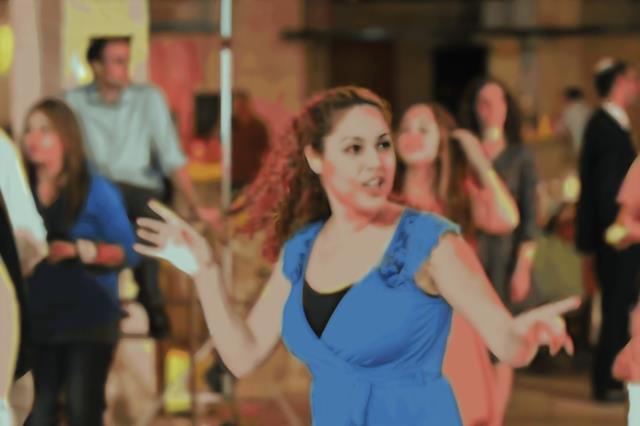} &
  \includegraphics[width=0.118\linewidth]{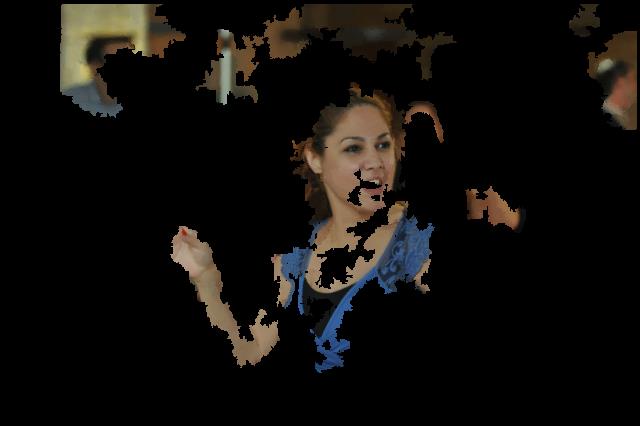} &
       \includegraphics[width=0.118\linewidth]{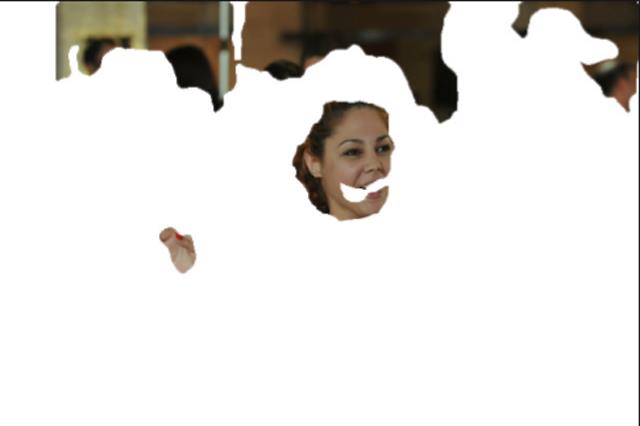} &
              \includegraphics[width=0.118\linewidth]{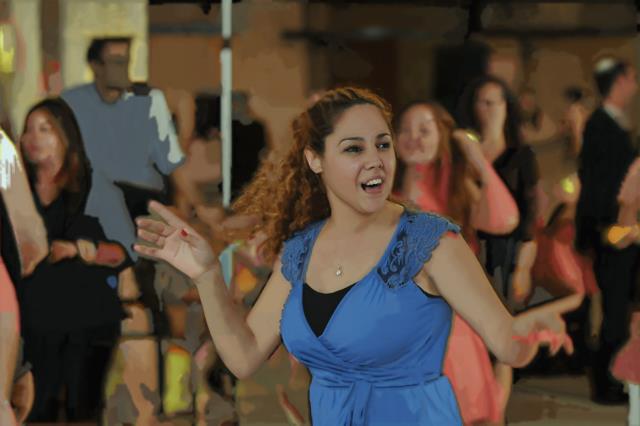} \\
               \includegraphics[width=0.118\linewidth]{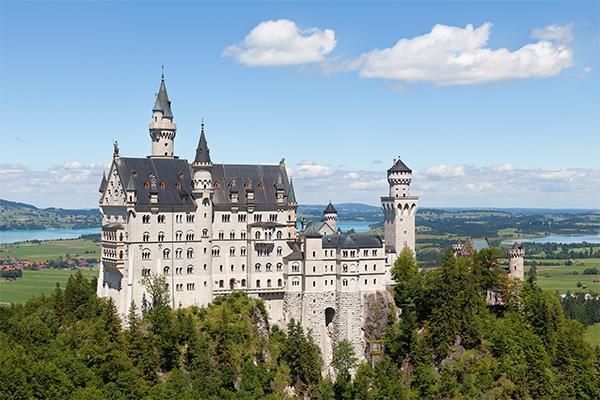} &
            \includegraphics[width=0.118\linewidth]{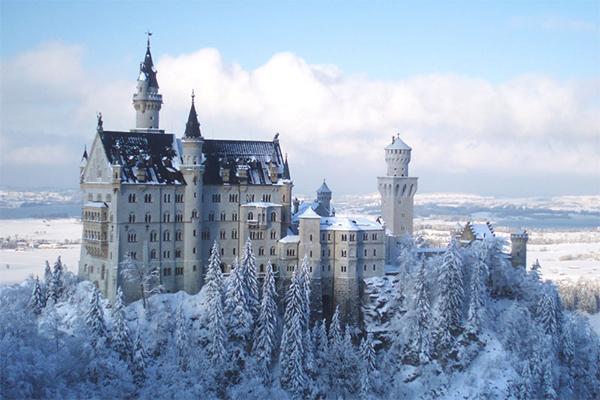} &
   \includegraphics[width=0.118\linewidth]{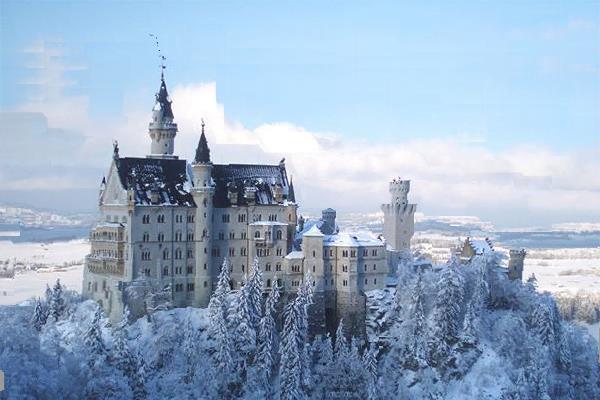} &
\includegraphics[width=0.118\linewidth]{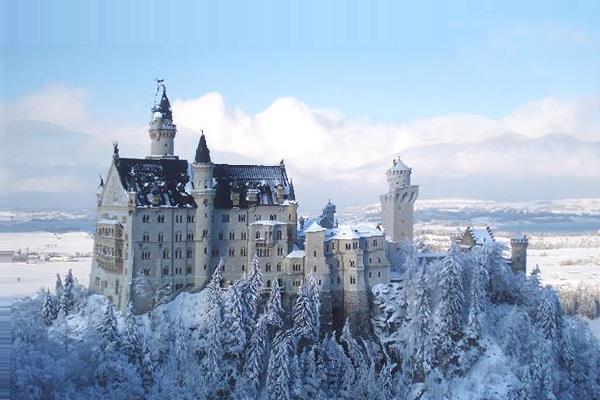} &
 \includegraphics[width=0.118\linewidth]{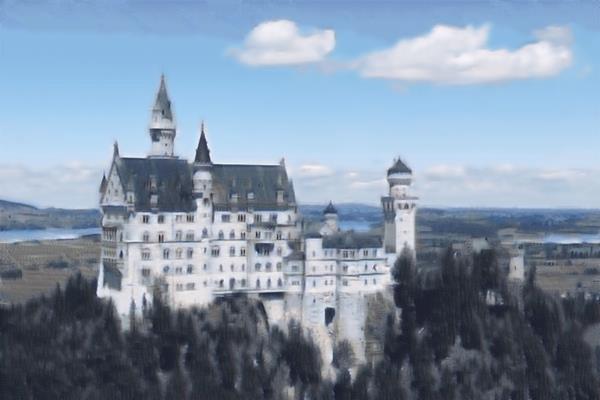} &
  \includegraphics[width=0.118\linewidth]{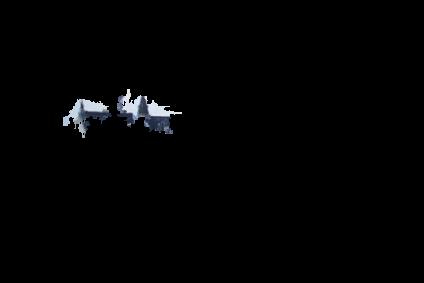} &
       \includegraphics[width=0.118\linewidth]{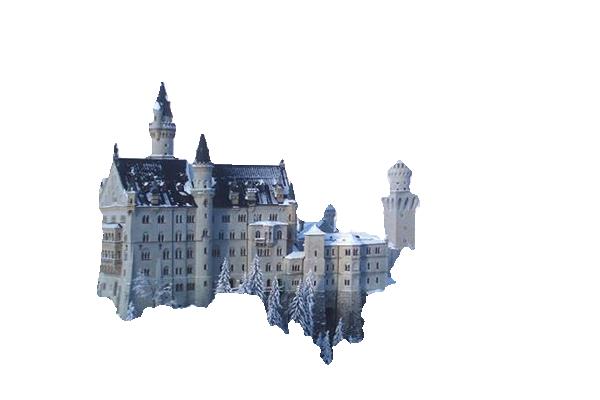} &
              \includegraphics[width=0.118\linewidth]{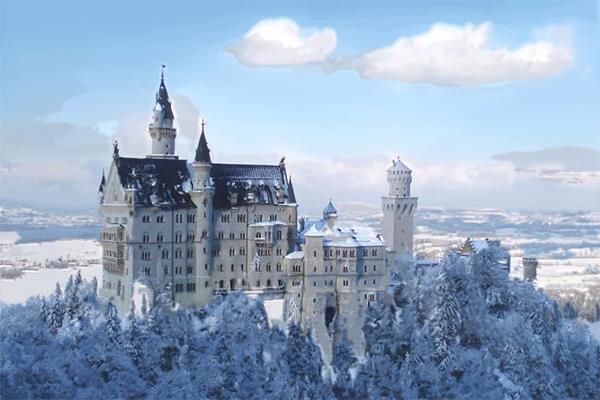} \\
             (a) Input (src) & (b) Input (ref) & (c) SIFT flow & (d) DeepFlow2 & (e) PatchMatch & (f) NRDC & (g) RFM & (h) Ours
  \end{tabular}
  }%

  \caption{Comparison of different dense correspondence methods on input pairs with the same scene but large variations in view, color, and tone. (For the first example, both public result and implementation of RFM are unavailable, so we have to make the result empty here.)}

  \label{fig:S2}
\end{figure*}

\begin{figure*}[!h] \centering

  { \footnotesize
    \setlength{\tabcolsep}{0.003\linewidth}
    \begin{tabular}{cccccccc}
                     \includegraphics[width=0.118\linewidth]{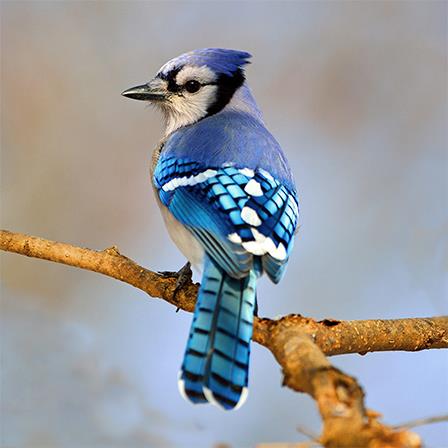} &
            \includegraphics[width=0.118\linewidth]{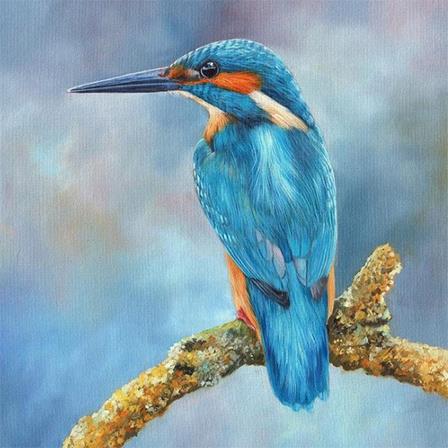} &
   \includegraphics[width=0.118\linewidth]{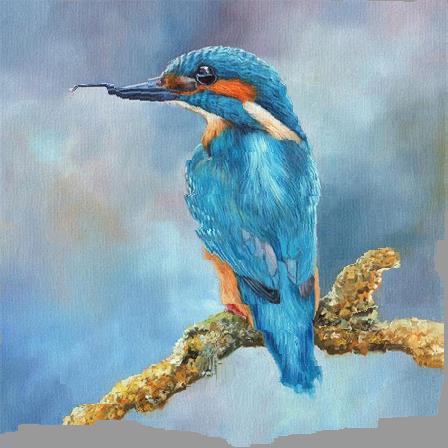} &
\includegraphics[width=0.118\linewidth]{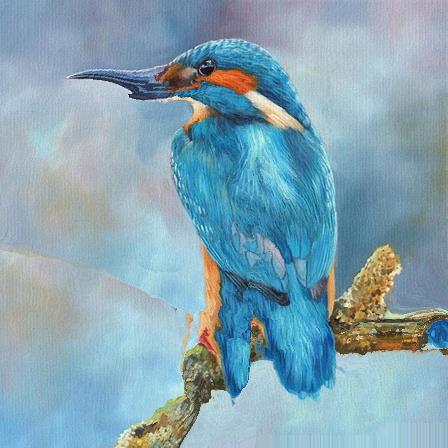} &
 \includegraphics[width=0.118\linewidth]{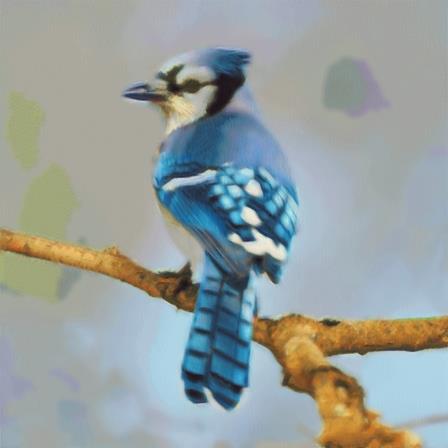} &
  \includegraphics[width=0.118\linewidth]{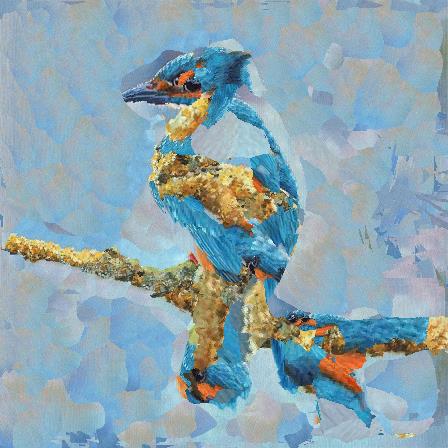} &
       \includegraphics[width=0.118\linewidth]{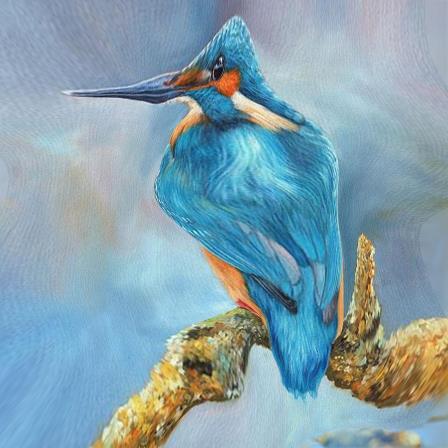} &
              \includegraphics[width=0.118\linewidth]{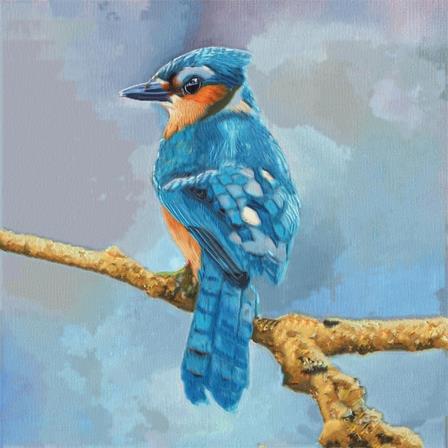} \\

 \includegraphics[width=0.118\linewidth]{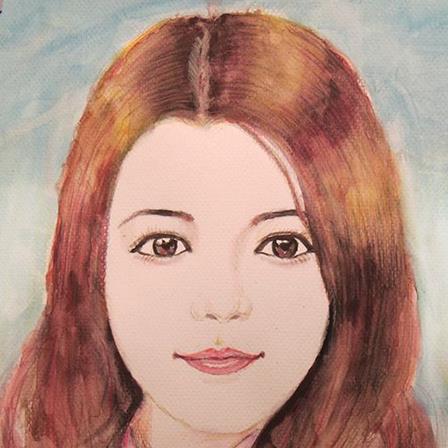} &
            \includegraphics[width=0.118\linewidth]{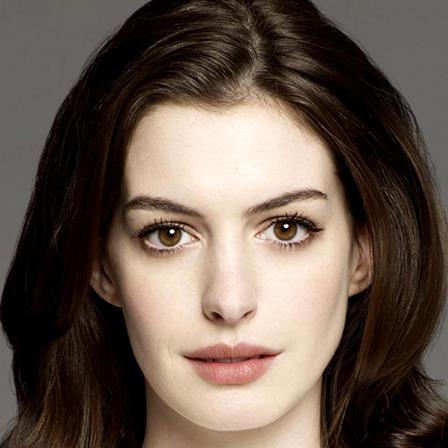} &
   \includegraphics[width=0.118\linewidth]{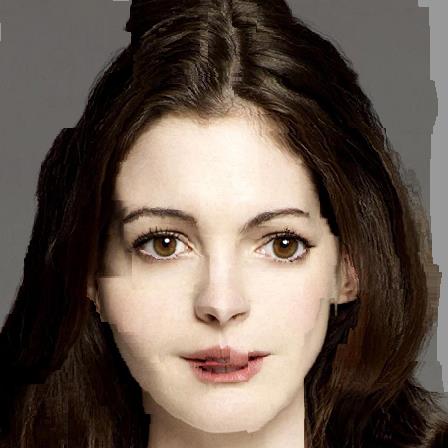} &
\includegraphics[width=0.118\linewidth]{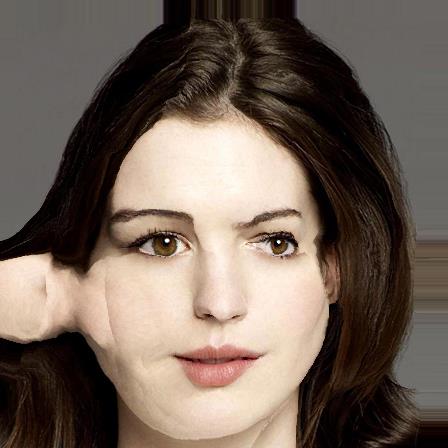} &
 \includegraphics[width=0.118\linewidth]{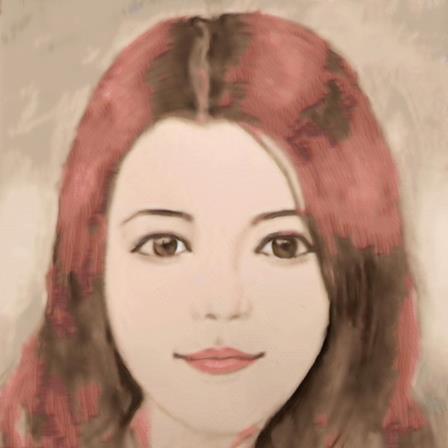} &
  \includegraphics[width=0.118\linewidth]{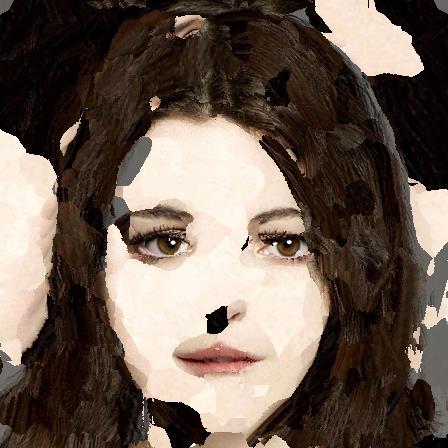} &
         \includegraphics[width=0.118\linewidth]{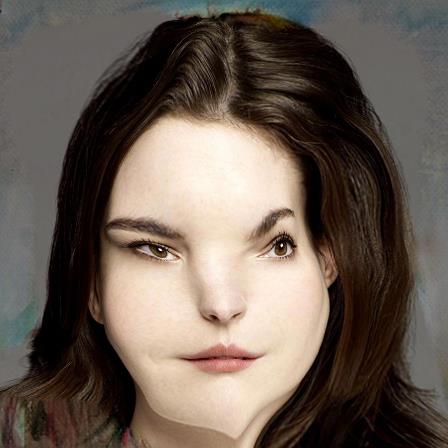}&
              \includegraphics[width=0.118\linewidth]{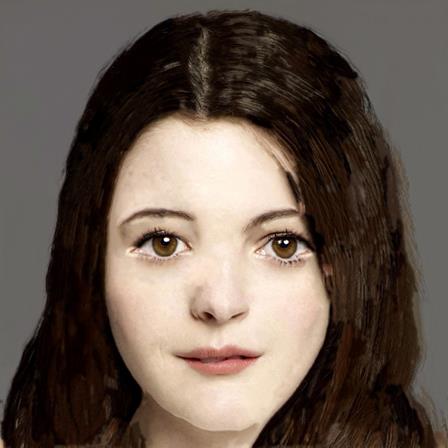} \\

              (a) Input (src) & (b) Input (ref) & (c) SIFT flow & (d) DeepFlow2 & (e) PatchMatch & (f) Daisy flow & (g) Halfway & (h) Ours
    \end{tabular}
  }%

  \caption{Comparison of different dense correspondence methods on input pairs semantically-related but with vastly different styles.}
  \label{fig:S3}
\end{figure*}
}

\section{Evaluations on Matching}

We evaluate the matching quality of our approach and state-of-the-art methods on three different categories of data: (I) the same scene, with very similar appearance and slight camera/object motions (e.g., neighboring frames in video); (II) the same scene, with variations in view, color, and tone (\emph{e.g.}, two photos from different cameras or illumination); (III) semantically-related scene with vastly different styles (e.g., photograph and painting) or appearances. Our default parameter ranges are designed for images with very different appearance like those in category (II) and (III). So for the special case (I) which always requires two views to be perfectly aligned in structures, we specially set $\{\alpha^L\}_{L=4, 3, 2, 1}=\{1.0,1.0,1.0,1.0\}$ to achieve comparable quality to other methods. For category (II) we use default weights plus offset $+0.1$, and for category (III) we use the default weights. 
We keep automatic weights selection based on inputs in future work. All other approaches are based on author-provided implementations with the default settings.

\begin{table}[ht]
\centering
\scalebox{0.95}{
 \begin{tabular}{lcccc}
  \toprule
   & SIFT Flow & DeepFlow2 & PatchMatch & ours\\
  \midrule
 PSNR & 27.97 & 30.04 & 35.18 & 34.16 \\
 Endpoint err. & 0.91 & 0.34 & 22.40 & 3.21\\
  \bottomrule
 \end{tabular}
 }

 \caption{Reconstruction accuracy and flow endpoint error of different dense correspondence estimation methods. }   
 \label{tab:S1}
 
\end{table}

\begin{table*}[t]
\center
\scalebox{0.95}{
 \begin{tabular}{lcccccccccccc|c}
  \toprule
   & aero & bike & boat & bottle & bus & car &chair  & table & mbike & sofa & train & tv &\textbf{ mean}\\
  \midrule
 PatchMatch \cite{barnes2009patchmatch} &6.5&6.3&2.6&2.9&2.3&4.7&3.3&12.5&2.0&0.0&4.2&2.6&4.2\\
  SIFT Flow ~\cite{Liu2011siftflow} &8.1&\textbf{14.3}&5.1&26.1&25&20.9&13.3&6.3&14.3&15.4&4.2&44.7&16.5\\
  Cycle consistency \cite{zhou2016learning}&12.9&6.3&10.3&\textbf{39.1}&27.3&\textbf{23.3}&13.3&\textbf{12.5}&6.1&\textbf{19.2}&\textbf{12.5}&36.8&18.3\\
   Ours&\textbf{19.4}&7.9&\textbf{15.4}&27.5&\textbf{47.7}&11.6&\textbf{20.0}&6.3&\textbf{18.4}&15.4&\textbf{12.5}&\textbf{50.0}&\textbf{21.0}\\
  \bottomrule
 \end{tabular}
 }
 \caption{Correspondence accuracy measured in PCK. The test is conducted on randomly selected 20 pairs of each category of the PASCAL 3D+ dataset. Overall, our method outperforms the three baselines.}
 \label{tab:S3}
\end{table*}

\begin{figure*}[ht] \centering
  { \footnotesize
    \setlength{\tabcolsep}{0.002\linewidth}
    \begin{tabular}{m{0.078\linewidth}cccccccc}
	Source&
	\includegraphics[height=0.078\linewidth]{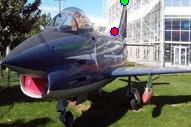}&
	\includegraphics[height=0.078\linewidth]{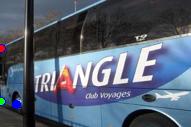}&
	\includegraphics[height=0.078\linewidth]{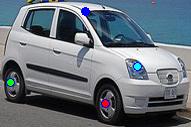}&
	\includegraphics[height=0.078\linewidth]{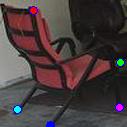}&
	\includegraphics[height=0.078\linewidth]{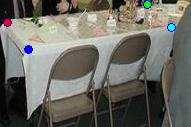}&
	\includegraphics[height=0.078\linewidth]{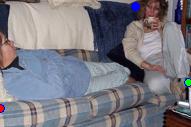}&
	\includegraphics[height=0.078\linewidth]{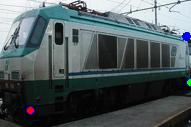}&
	\includegraphics[height=0.078\linewidth]{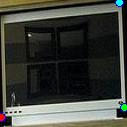}\\
	PatchMatch&
	\includegraphics[height=0.078\linewidth]{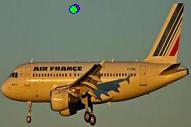}&
	\includegraphics[height=0.078\linewidth]{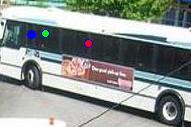}&
	\includegraphics[height=0.078\linewidth]{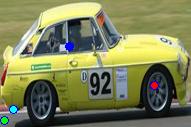}&
	\includegraphics[height=0.078\linewidth]{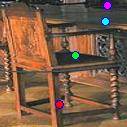}&
	\includegraphics[height=0.078\linewidth]{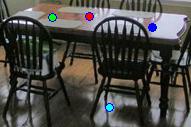}&
	\includegraphics[height=0.078\linewidth]{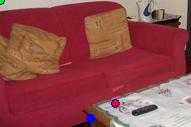}&
	\includegraphics[height=0.078\linewidth]{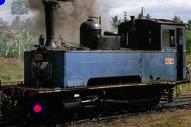}&
	\includegraphics[height=0.078\linewidth]{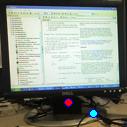}\\
	SIFT Flow&
	\includegraphics[height=0.078\linewidth]{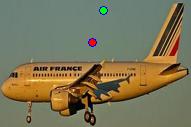}&
	\includegraphics[height=0.078\linewidth]{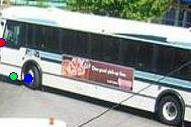}&
	\includegraphics[height=0.078\linewidth]{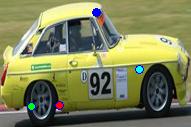}&
	\includegraphics[height=0.078\linewidth]{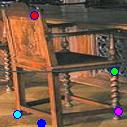}&
	\includegraphics[height=0.078\linewidth]{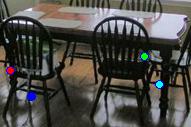}&
	\includegraphics[height=0.078\linewidth]{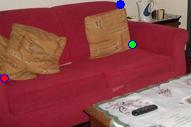}&
	\includegraphics[height=0.078\linewidth]{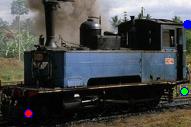}&
	\includegraphics[height=0.078\linewidth]{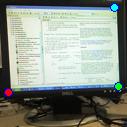}\\
	\vspace{-0.2in}Cycle \quad\quad  consistency &
	\includegraphics[height=0.078\linewidth]{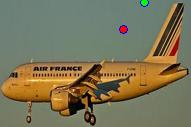}&
	\includegraphics[height=0.078\linewidth]{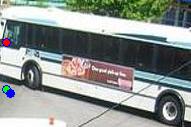}&
	\includegraphics[height=0.078\linewidth]{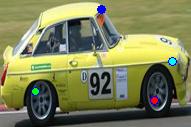}&
	\includegraphics[height=0.078\linewidth]{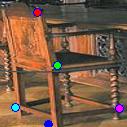}&
	\includegraphics[height=0.078\linewidth]{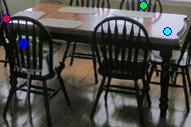}&
	\includegraphics[height=0.078\linewidth]{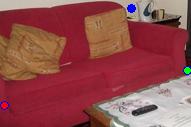}&
	\includegraphics[height=0.078\linewidth]{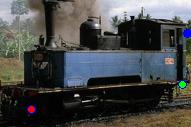}&
	\includegraphics[height=0.078\linewidth]{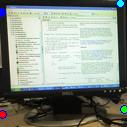}\\
	Ours&
	\includegraphics[height=0.078\linewidth]{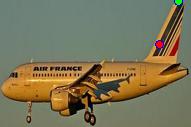}&
	\includegraphics[height=0.078\linewidth]{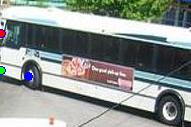}&
	\includegraphics[height=0.078\linewidth]{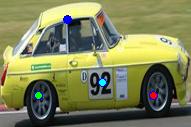}&
	\includegraphics[height=0.078\linewidth]{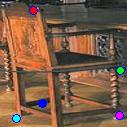}&
	\includegraphics[height=0.078\linewidth]{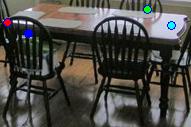}&
	\includegraphics[height=0.078\linewidth]{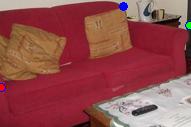}&
	\includegraphics[height=0.078\linewidth]{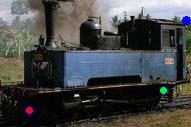}&
	\includegraphics[height=0.078\linewidth]{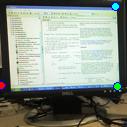}\\
	Reference&
	\includegraphics[height=0.078\linewidth]{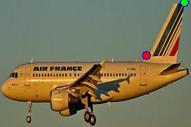}&
	\includegraphics[height=0.078\linewidth]{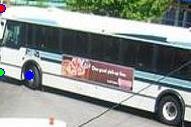}&
	\includegraphics[height=0.078\linewidth]{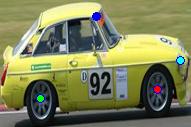}&
	\includegraphics[height=0.078\linewidth]{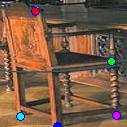}&
	\includegraphics[height=0.078\linewidth]{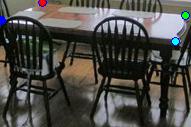}&
	\includegraphics[height=0.078\linewidth]{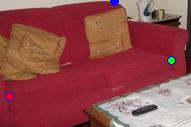}&
	\includegraphics[height=0.078\linewidth]{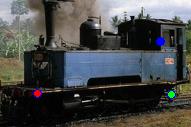}&
	\includegraphics[height=0.078\linewidth]{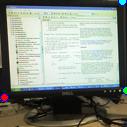}\\
    \end{tabular}
  }%

  \caption{Visual comparison of keypoint transfer performance for different methods on representative examples of the PASCAL 3D+ dataset.}
  \label{fig:S3_CC}
\end{figure*}

Category (I) is tested on the Middlebury optical flow benchmark dataset, total $23$ color image pairs. We compared with PatchMatch~\cite{barnes2009patchmatch}, SIFT Flow~\cite{Liu2011siftflow} and DeepFlow2~\cite{weinzaepfel2013deepmatch}. We evaluate the matching accuracy by both flow endpoint error and reconstruction errors (\emph{i.e.}, PSNR metric). Similar to Patchmatch, our method does not assume any flow continuity, so it performs better than the flow methods (\emph{i.e.}, Sift Flow and DeepFlow2) in reconstruction error but worse in flow endpoint error, as shown on Table \ref{tab:S1}. Since the task of this paper is image reconstruction rather than pure motion estimation, the reconstruction error makes more sense in our applications. Note that our reconstructed results look slightly worse than PatchMatch. This is because our approach does not consider color similarity as well as Patchmatch, which causes inconsistent color matches in our results. Figure~\ref{fig:S1} shows a visual comparison of the three methods and ours.

For category (II), we collect public data from~\cite{yang2007registration,haCohen2011nrdc,Shen2016RFM}; we show some examples in Figure~\ref{fig:S2}. The input pair of images differ significantly in object pose, illumination, and season. Since PatchMatch~\cite{barnes2009patchmatch} is based on color similarity, it may fail to match similar structures with different colors, leading to a result with appearance that is close to the source image (see Figure~\ref{fig:S2}(e)). In contrast, SIFT flow~\cite{Liu2011siftflow} and Deepflow~\cite{weinzaepfel2013deepmatch} are better in matching structures (see Figure~\ref{fig:S2}(c)(d)) because sparse features are matched instead of color. However, there are noticeable mismatches in areas without feature points. Regional foremost matching (RFM)~\cite{Shen2016RFM} and NRDC~\cite{haCohen2011nrdc} work well on these regions where matches are found. However, RFM fails for large non-rigid motion (Figure~\ref{fig:S2}(g))) and NRDC can reliably match only a small fraction of the image in some cases (Figure~\ref{fig:S2}(f)). Our results (Figure~\ref{fig:S2}(h)) look visually comparable with either NRDC or RFM at regions with high confidence, and provide a better guess elsewhere.

\begin{figure*}[t] \centering
  { \footnotesize
    \setlength{\tabcolsep}{0.003\linewidth}
    \begin{tabular}{ccccccc}
    &
        \includegraphics[width=0.156\linewidth]{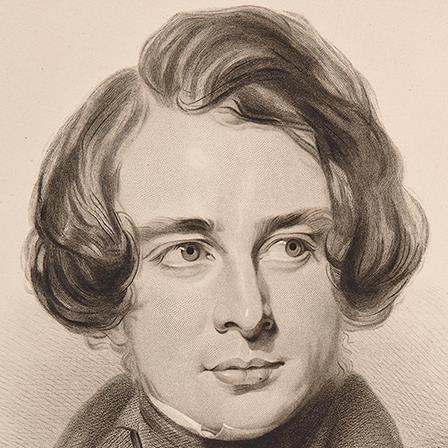} &
      \includegraphics[width=0.156\linewidth]{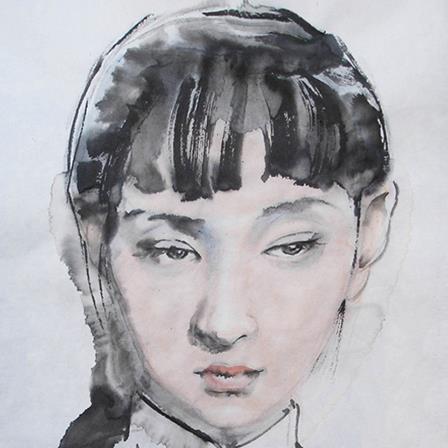} \hspace{0.08\linewidth}&
          &     \includegraphics[width=0.156\linewidth]{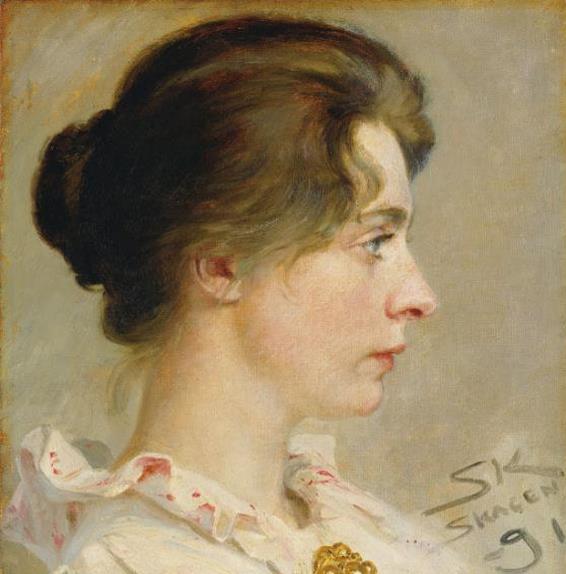} &
          \includegraphics[width=0.156\linewidth]{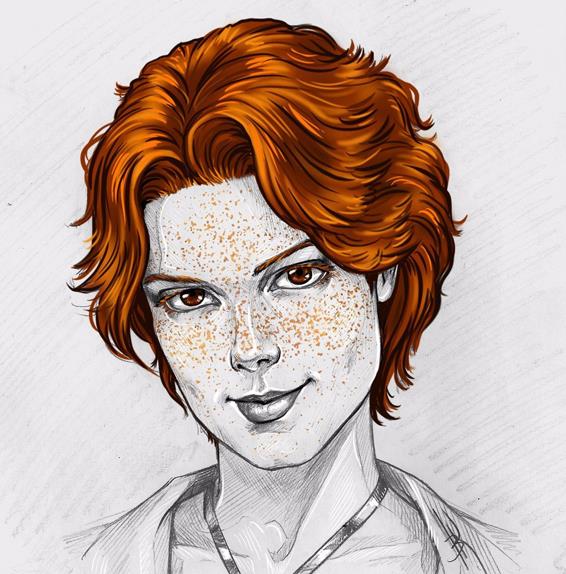}\\
             \includegraphics[width=0.156\linewidth]{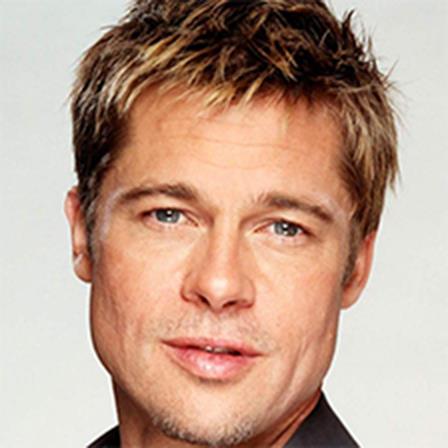} &
      \includegraphics[width=0.156\linewidth]{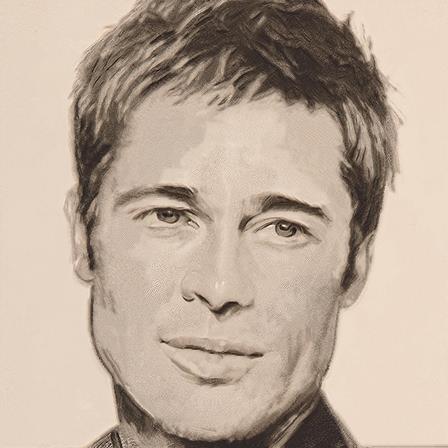} &
      \includegraphics[width=0.156\linewidth]{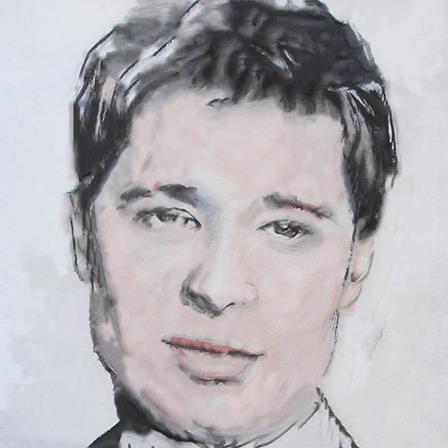} \hspace{0.08\linewidth}&
       \includegraphics[width=0.156\linewidth]{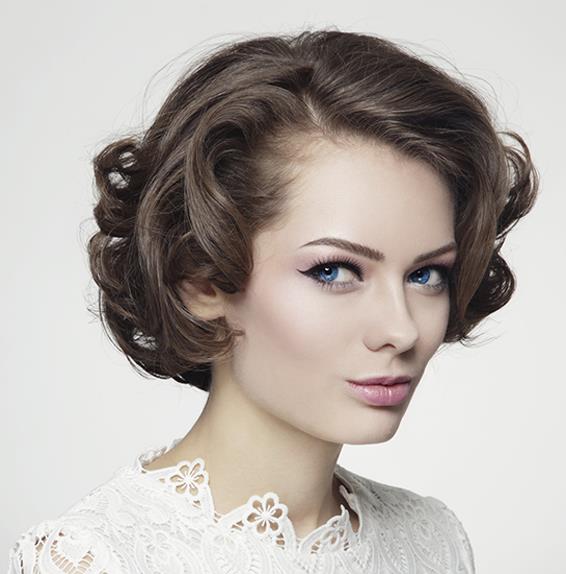} &
       \includegraphics[width=0.156\linewidth]{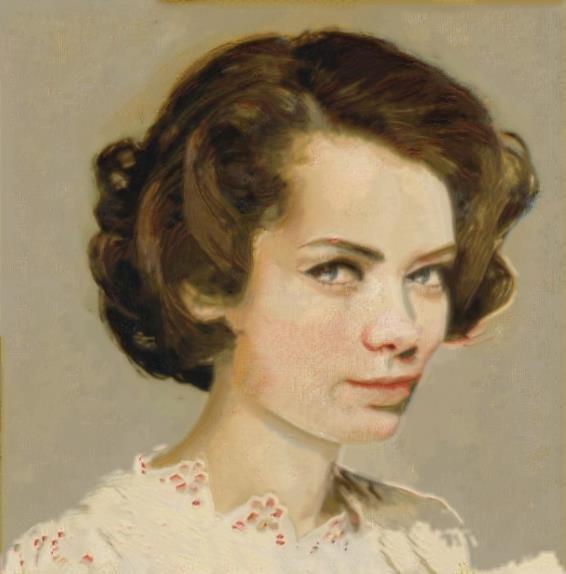} &
      \includegraphics[width=0.156\linewidth]{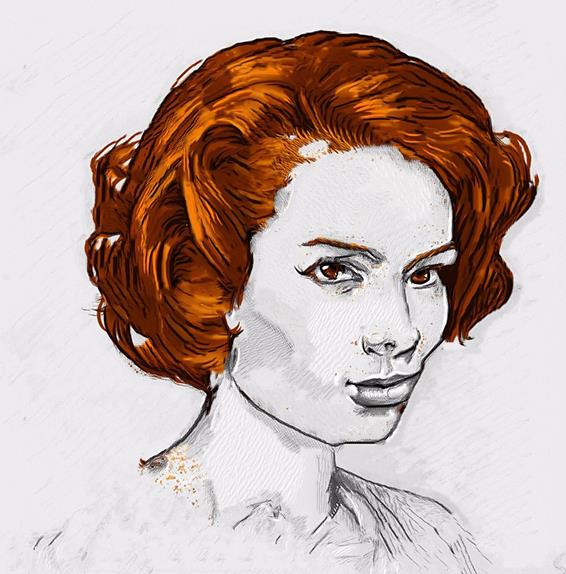} \\

      &
        \includegraphics[width=0.156\linewidth]{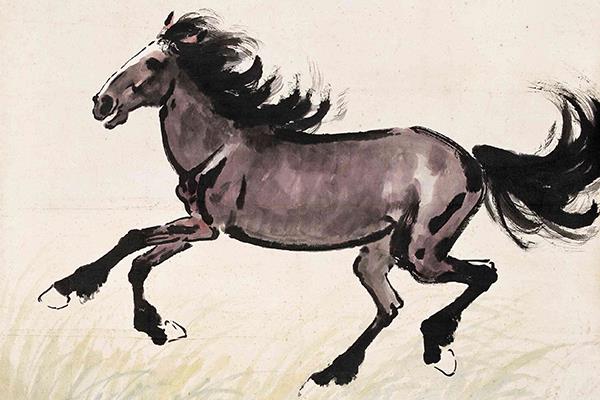} &
      \includegraphics[width=0.156\linewidth]{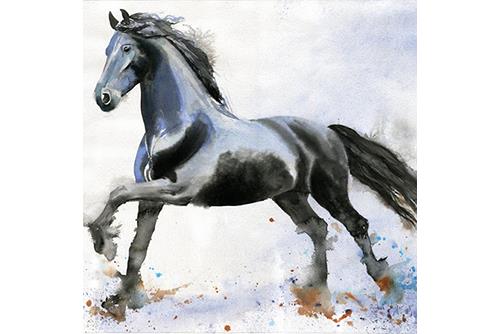} \hspace{0.08\linewidth}&
          &     \includegraphics[width=0.156\linewidth]{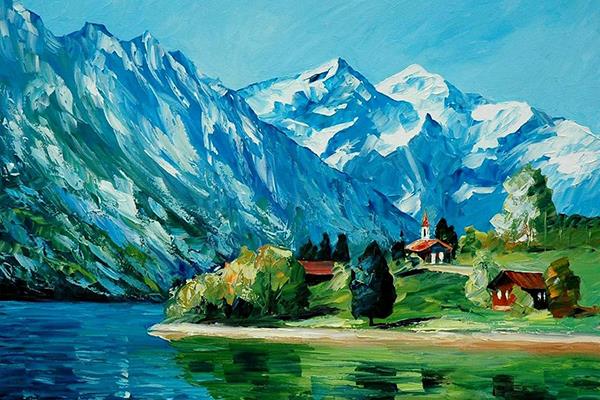} &
           \includegraphics[width=0.156\linewidth]{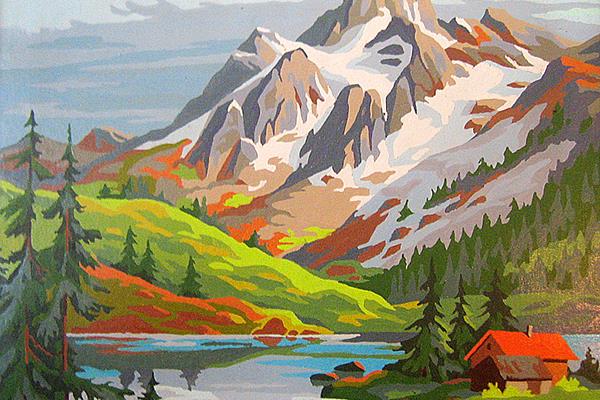}
           \\
           \includegraphics[width=0.156\linewidth]{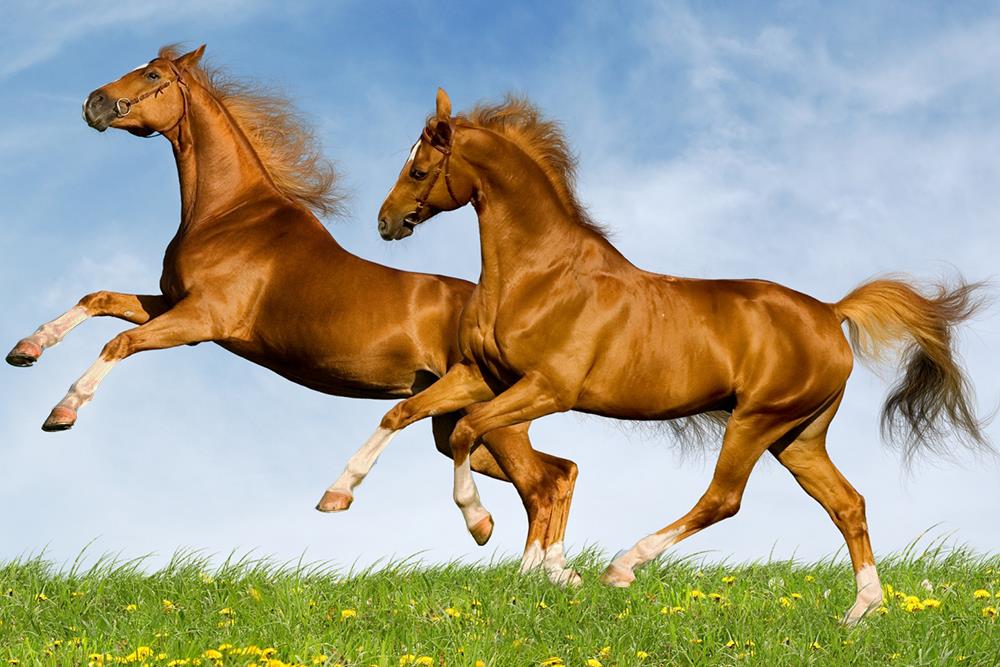} &
      \includegraphics[width=0.156\linewidth]{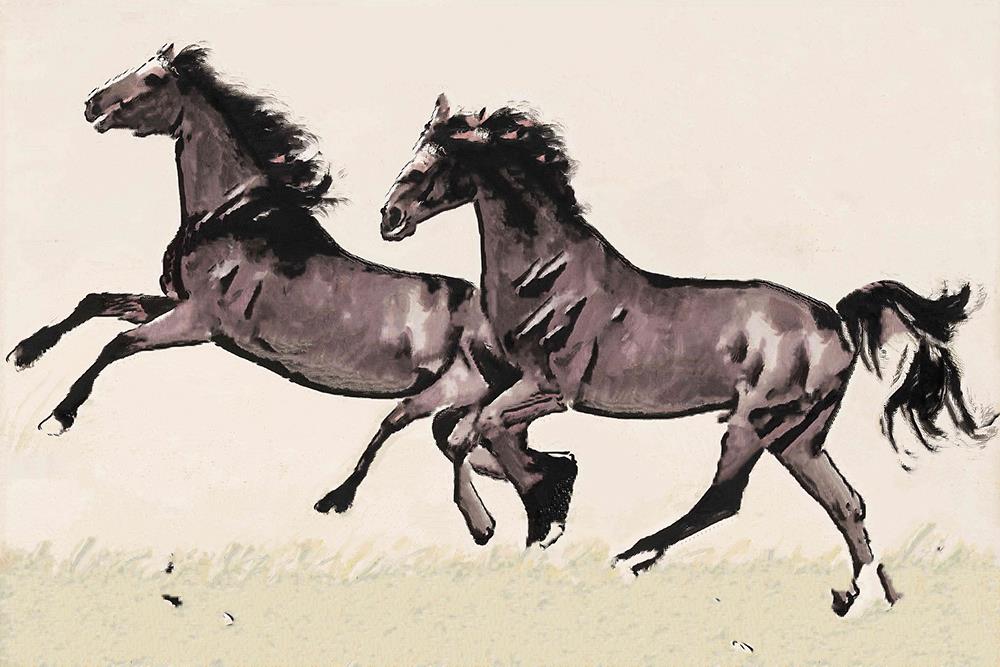} &
      \includegraphics[width=0.156\linewidth]{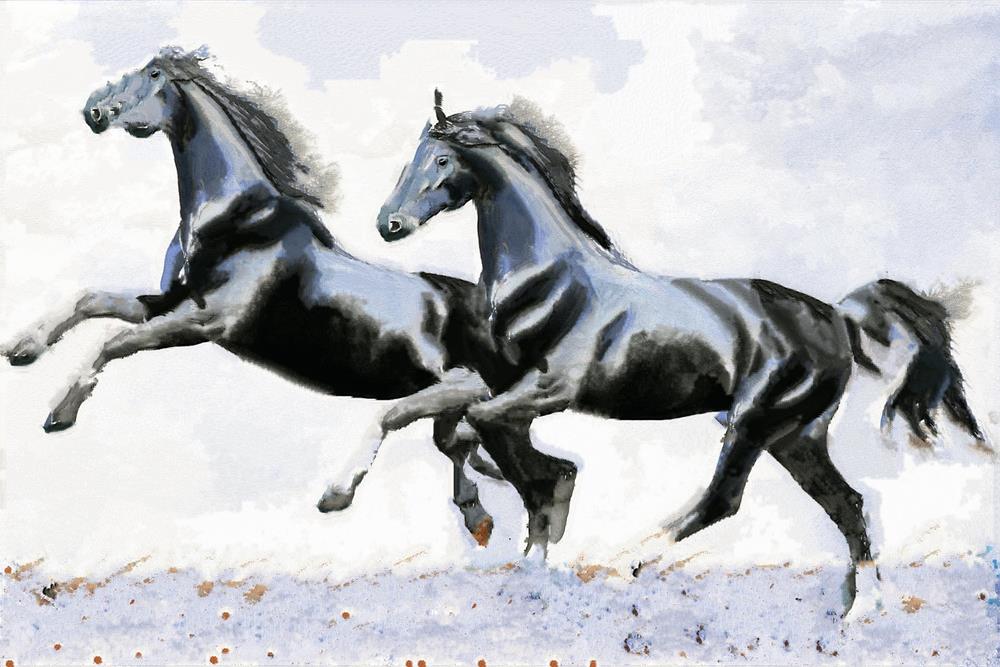}  \hspace{0.08\linewidth}&
       \includegraphics[width=0.156\linewidth]{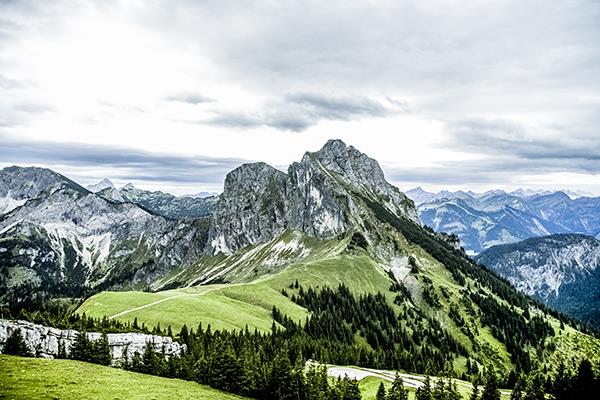} &
       \includegraphics[width=0.156\linewidth]{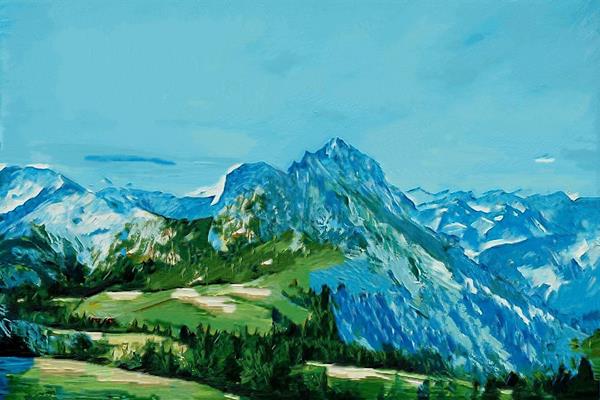} &
      \includegraphics[width=0.156\linewidth]{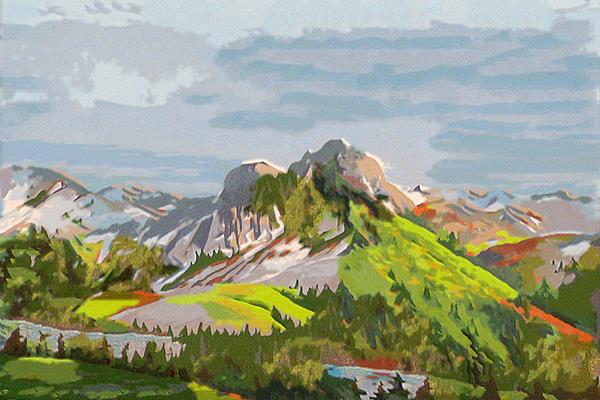} \\
      &
        \includegraphics[width=0.156\linewidth]{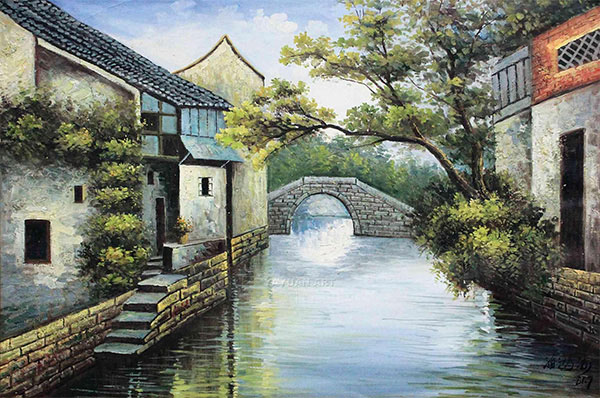} &
      \includegraphics[width=0.135\linewidth]{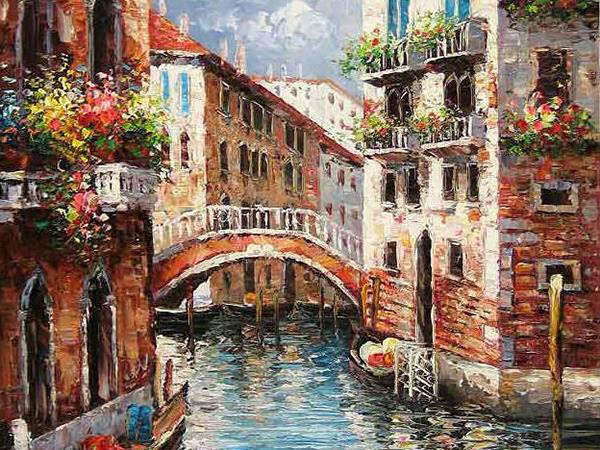} \hspace{0.08\linewidth}&
          &     \includegraphics[width=0.156\linewidth]{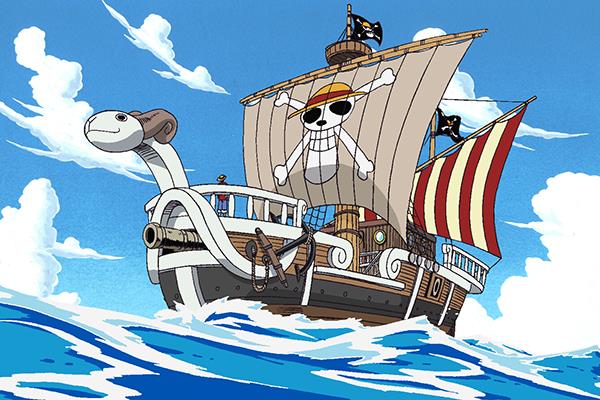} &
           \includegraphics[width=0.156\linewidth]{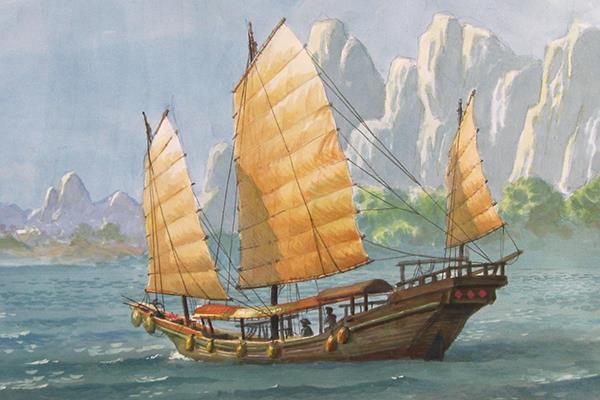} \\
            \includegraphics[width=0.156\linewidth]{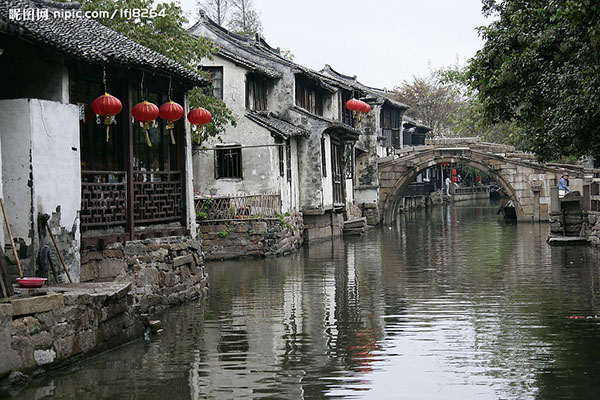} &
      \includegraphics[width=0.156\linewidth]{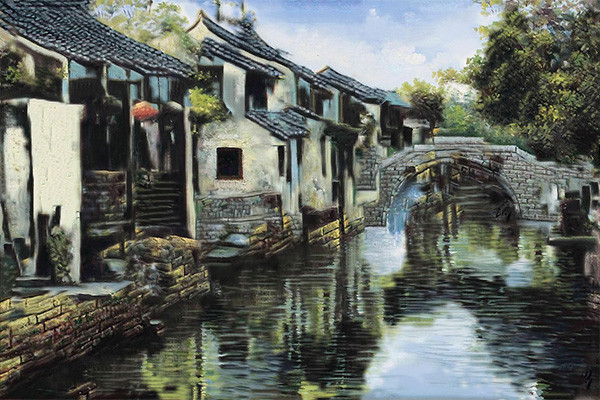} &
      \includegraphics[width=0.156\linewidth]{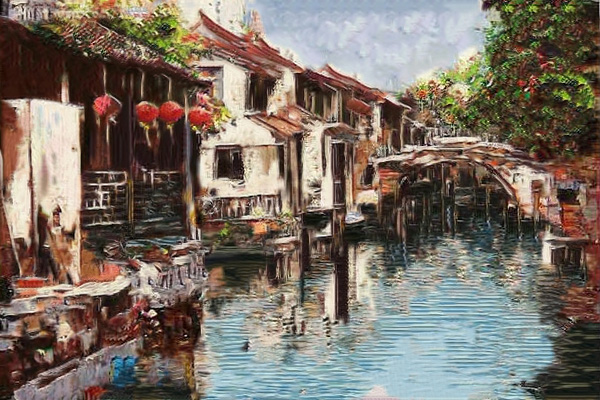}  \hspace{0.08\linewidth}&
       \includegraphics[width=0.156\linewidth]{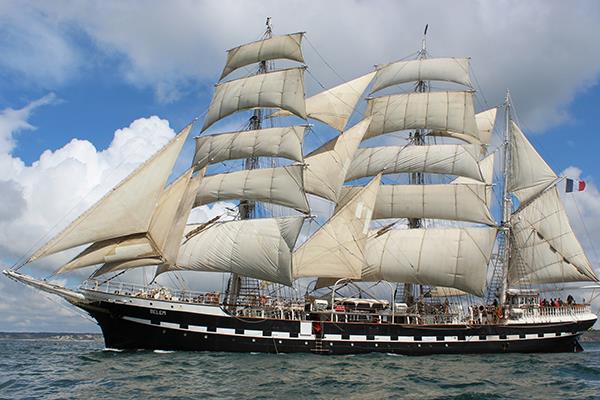} &
       \includegraphics[width=0.156\linewidth]{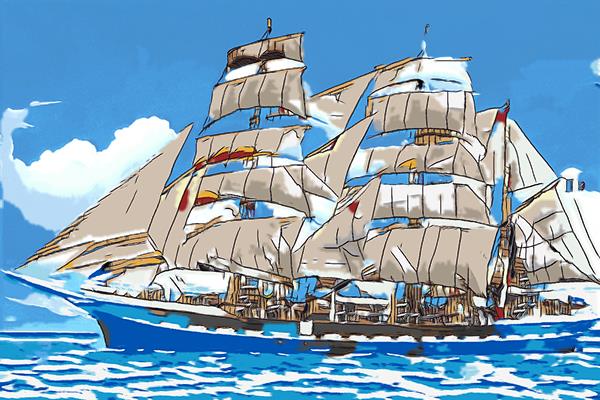} &
      \includegraphics[width=0.156\linewidth]{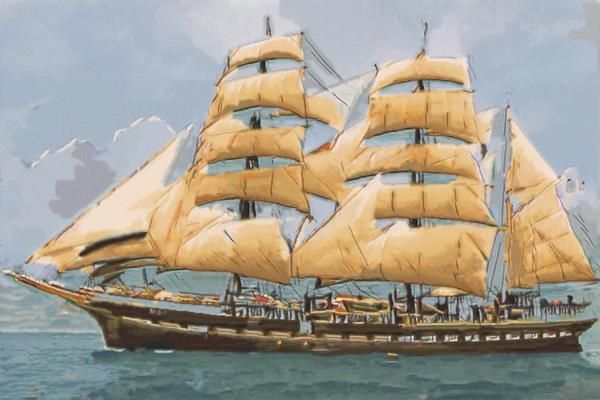}

    \end{tabular}
  }%

  \caption{Our photo-to-style transfer results. For each group, two images in the upper row are reference styles. The leftmost one in the lower row is the input photo, other two images are our results. }
  \label{fig:artfilter}
\end{figure*}

For category (III), we collect pairs of photo and painting online by searching generic words such as ``bird'', ``girl'', and ``portrait''. Two examples are shown in Figure~\ref{fig:S3}. The task would be intractable for existing matching work~\cite{Liu2011siftflow,weinzaepfel2013deepmatch,barnes2009patchmatch,Yang2014daisyflow}, and we can see noticeable artifacts in their results (Figure~\ref{fig:S3}(c)(d)(e)(f)). The Halfway morphing~\cite{Liao2014morphing} addresses this kind of problem. However, their method relies on a modified SSIM term to find low-level feature correspondences, and needs user interaction to provide high-level matching. Without user input, they may fail as well on these cases (Figure~\ref{fig:S3}(g)). By contrast, our approach is automatic and can produce visually acceptable results (Figure~\ref{fig:S3}(h)).

We further evaluate on the Pascal 3D+ dataset~\cite{xiang2014beyond}, which provides annotated sparse correspondences for semantically-related objects with remarkably different appearances. For each category in the Pascal 3D+ dataset, we randomly sample 20 image pairs from the training and validation datasets. Cycle consistency \cite{zhou2016learning} is a representative work considering high-level semantic information for dense matching. We conduct the same evaluation as that of Zhou et al.~\cite{zhou2016learning} for all competing methods. The quality of correspondences is measured by the percentage of correct keypoint transfer (PCK) over all pairs suggested by Zhou et al.~\cite{zhou2016learning}. The quantitative comparisons between different methods are shown in Table \ref{tab:S3} and the visual comparisons on representative pairs are shown in Figure~\ref{fig:S3_CC}. Both our method and cycle consistency obtain better performance than methods based on low-level feature, e.g. SIFT Flow and PatchMatch. Overall, our method performs better than cycle consistency, even though the features we used are not trained on the Pascal 3D+ dataset.

\section{Applications}

In this section, we show how our approach can be effectively used for four different tasks of visual attribute transfer: photo to style, style to style, style to photo, and photo to photo. For comparison with other methods, we again ran author-released implementations with default settings, or submit our images to their services or apps. More comparisons and results can be found in our supplemental material. All results are better viewed in electronic form.

\begin{figure*}[h] \centering
  { \footnotesize
    \setlength{\tabcolsep}{0.003\linewidth}
    \begin{tabular}{ccccccc}

         \includegraphics[width=0.135\linewidth]{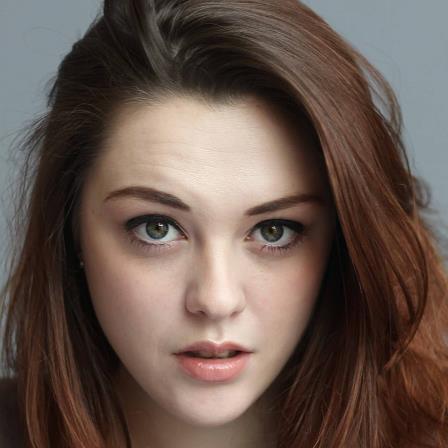} &
      \includegraphics[width=0.135\linewidth]{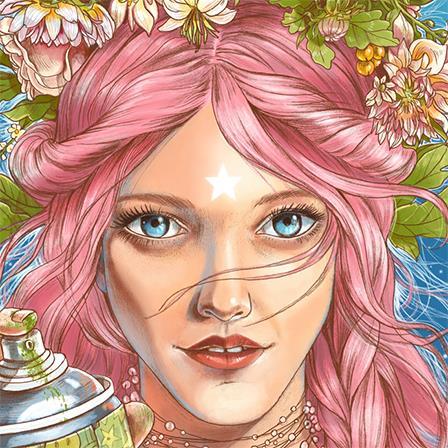} &
      \includegraphics[width=0.135\linewidth]{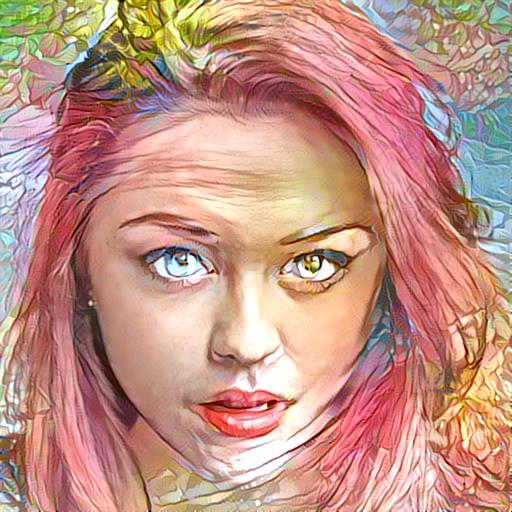} &
      \includegraphics[width=0.135\linewidth]{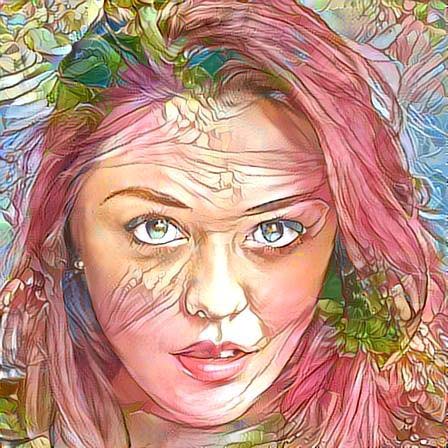} &
      \includegraphics[width=0.135\linewidth]{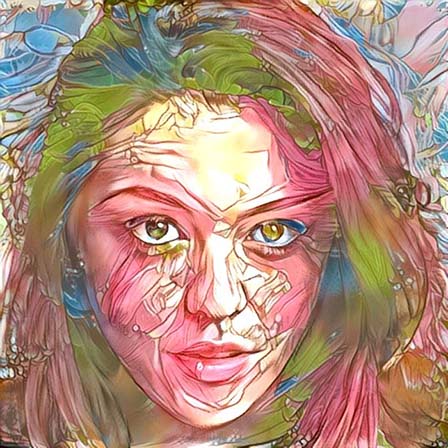} &
        \includegraphics[width=0.135\linewidth]{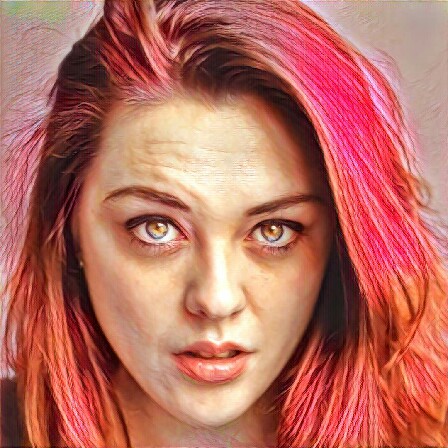}&
          \includegraphics[width=0.135\linewidth]{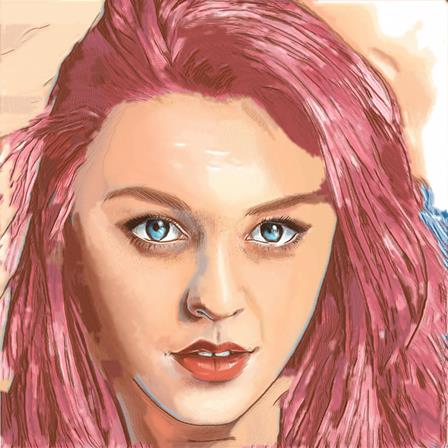} \\
                 Input (src)& Input (ref)  & Neural style & Deep style & Ostagram & Prisma & Ours\\
 \includegraphics[width=0.135\linewidth]{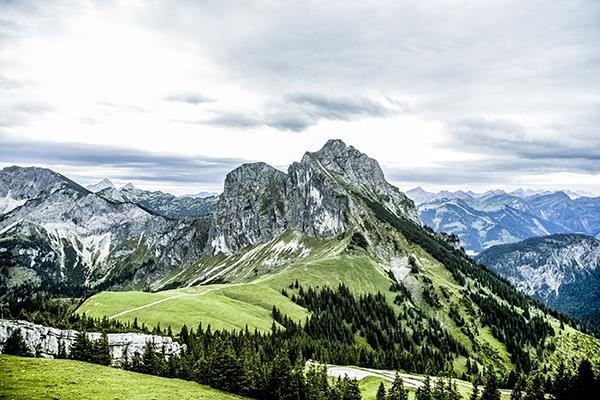} &
      \includegraphics[width=0.135\linewidth]{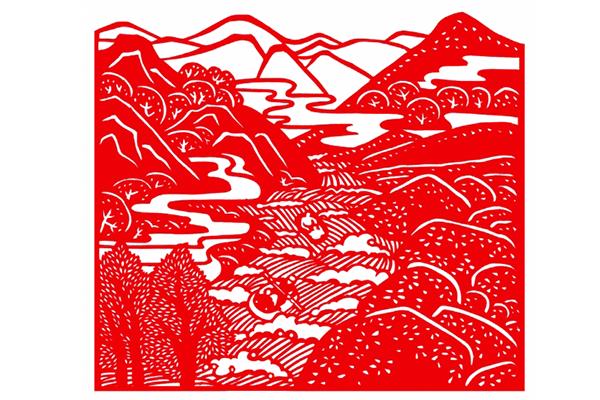} &
      \includegraphics[width=0.135\linewidth]{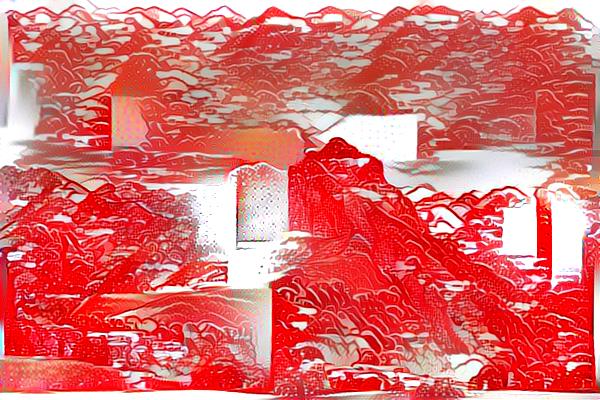} &
      \includegraphics[width=0.135\linewidth]{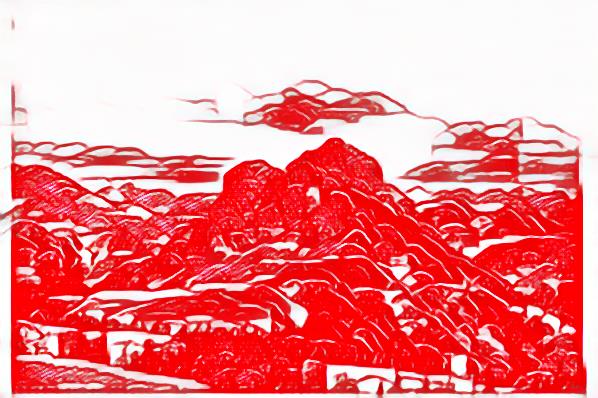} &
      \includegraphics[width=0.135\linewidth]{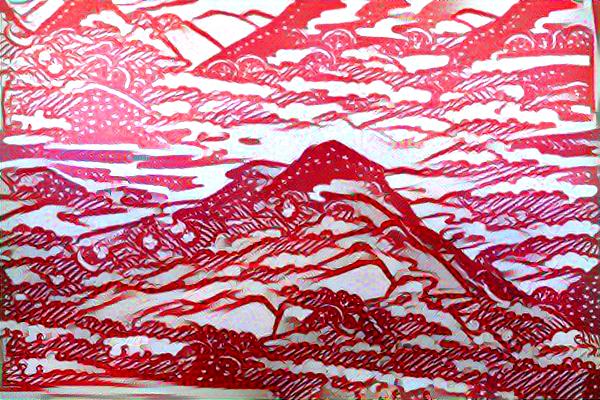} &
        \includegraphics[width=0.135\linewidth]{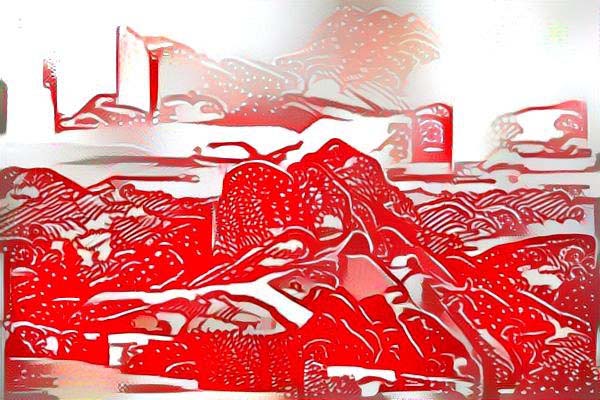}&
          \includegraphics[width=0.135\linewidth]{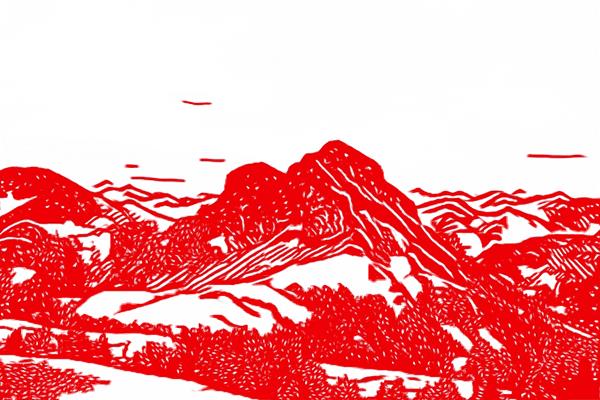} \\
                 Input (src)& Input (ref)  & Neural style &  Perceptual loss & CNNMRF & Deep style & Ours\\
                  \includegraphics[width=0.135\linewidth]{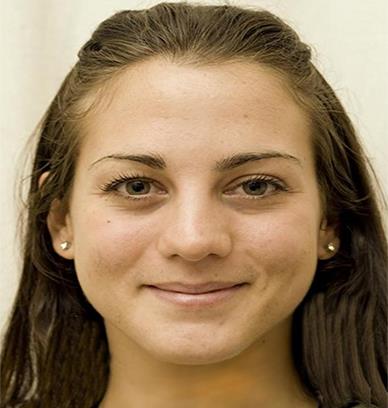} &
      \includegraphics[width=0.135\linewidth]{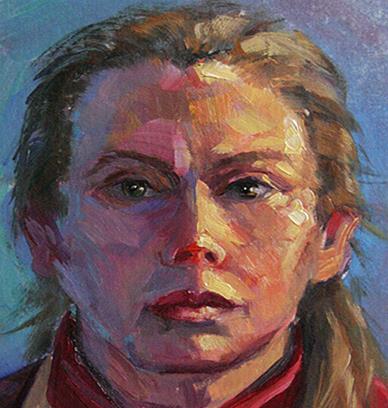} &
      \includegraphics[width=0.135\linewidth]{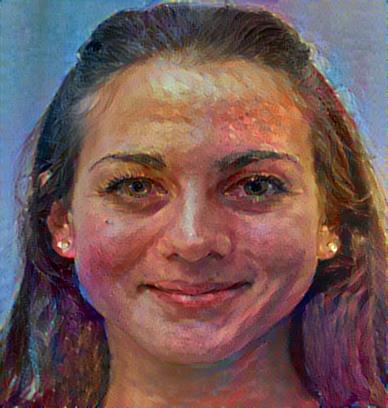} &
      \includegraphics[width=0.135\linewidth]{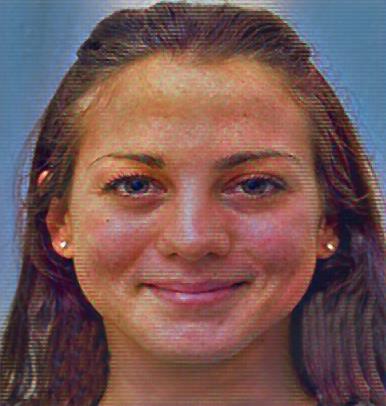} &
      \includegraphics[width=0.135\linewidth]{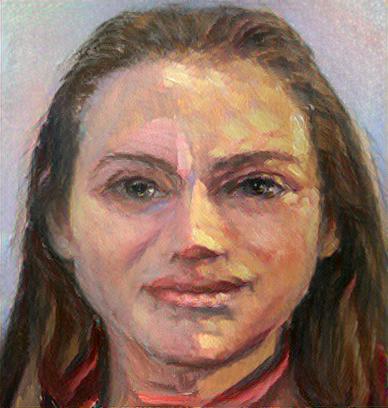} &
        \includegraphics[width=0.135\linewidth]{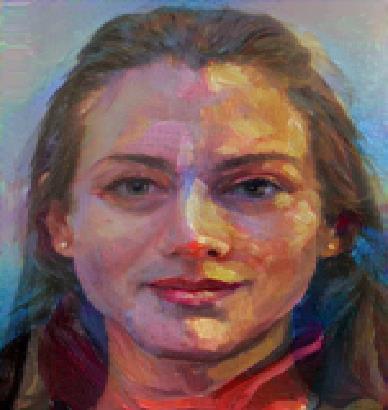}&
          \includegraphics[width=0.135\linewidth]{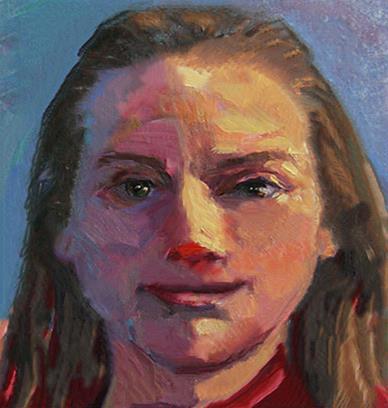}
     \\
       Input (src)& Input (ref) & Neural style & Perceptual loss & CNNMRF & Portrait style & Ours

    \end{tabular}
  }%

  \caption{Comparison with other style-transfer methods and apps based on neural network.}

  \label{fig:stylecompare}
\end{figure*}

\begin{figure}[!h] \centering
  { \footnotesize
    \setlength{\tabcolsep}{0.003\linewidth}
    \begin{tabular}{ccc}

        \includegraphics[width=0.191\linewidth]{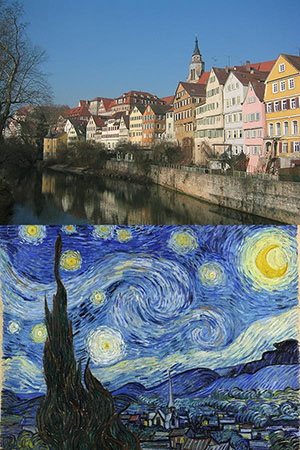} &
      \includegraphics[width=0.382\linewidth]{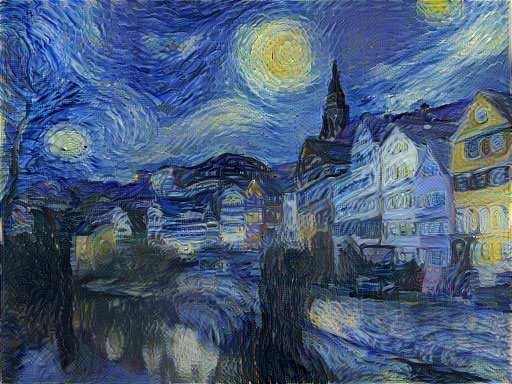} &
      \includegraphics[width=0.382\linewidth]{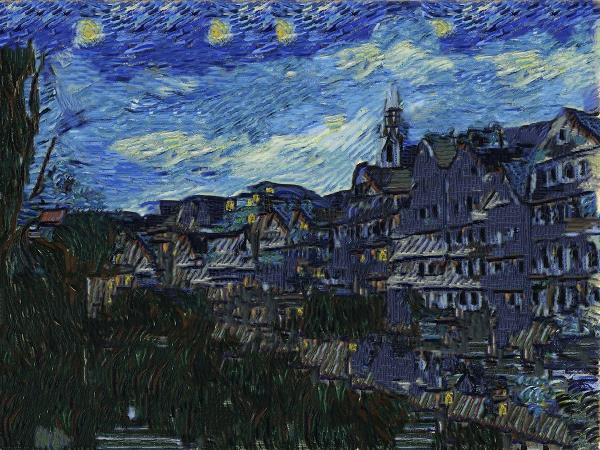} \\
                 Inputs &  Neural style & Ours\\
    \end{tabular}
	     \begin{tabular}{ccc}

        \includegraphics[width=0.321\linewidth]{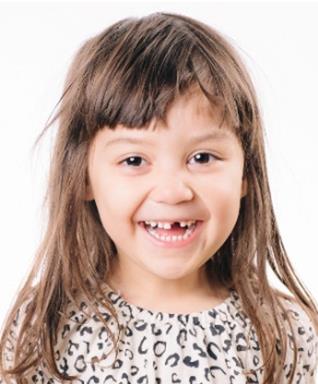} &
      \includegraphics[width=0.321\linewidth]{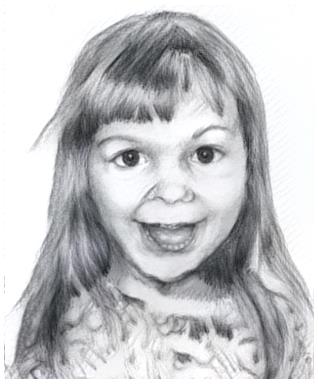} &
      \includegraphics[width=0.321\linewidth]{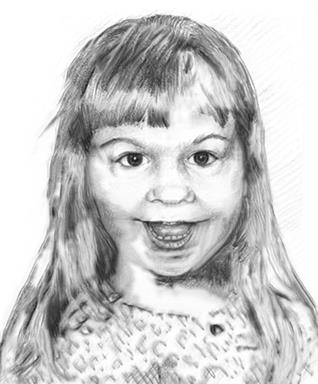} \\
                 Input ($A$) &  CNNMRF ($A'$)  & Ours ($A'$)\\
         \includegraphics[width=0.321\linewidth]{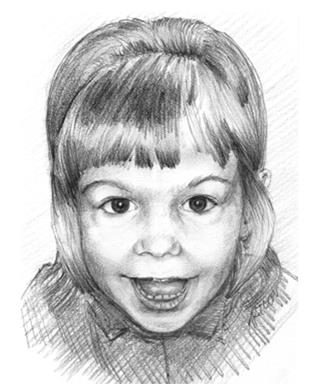} &
      \includegraphics[width=0.321\linewidth]{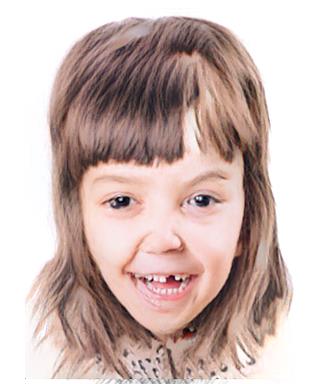} &
      \includegraphics[width=0.31\linewidth]{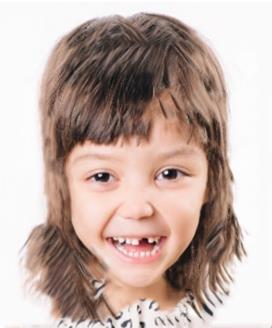} \\
                 Input ($B'$) &  CNNMRF ($B$)  & Ours ($B$)

    \end{tabular}

  }%

  \caption{Comparison with Neural style and CNNMRF on their examples.}
  \label{fig:stylecompare1}
\end{figure}

\begin{figure*}[t]
\footnotesize
\centering
   \includegraphics[width=0.95\linewidth]{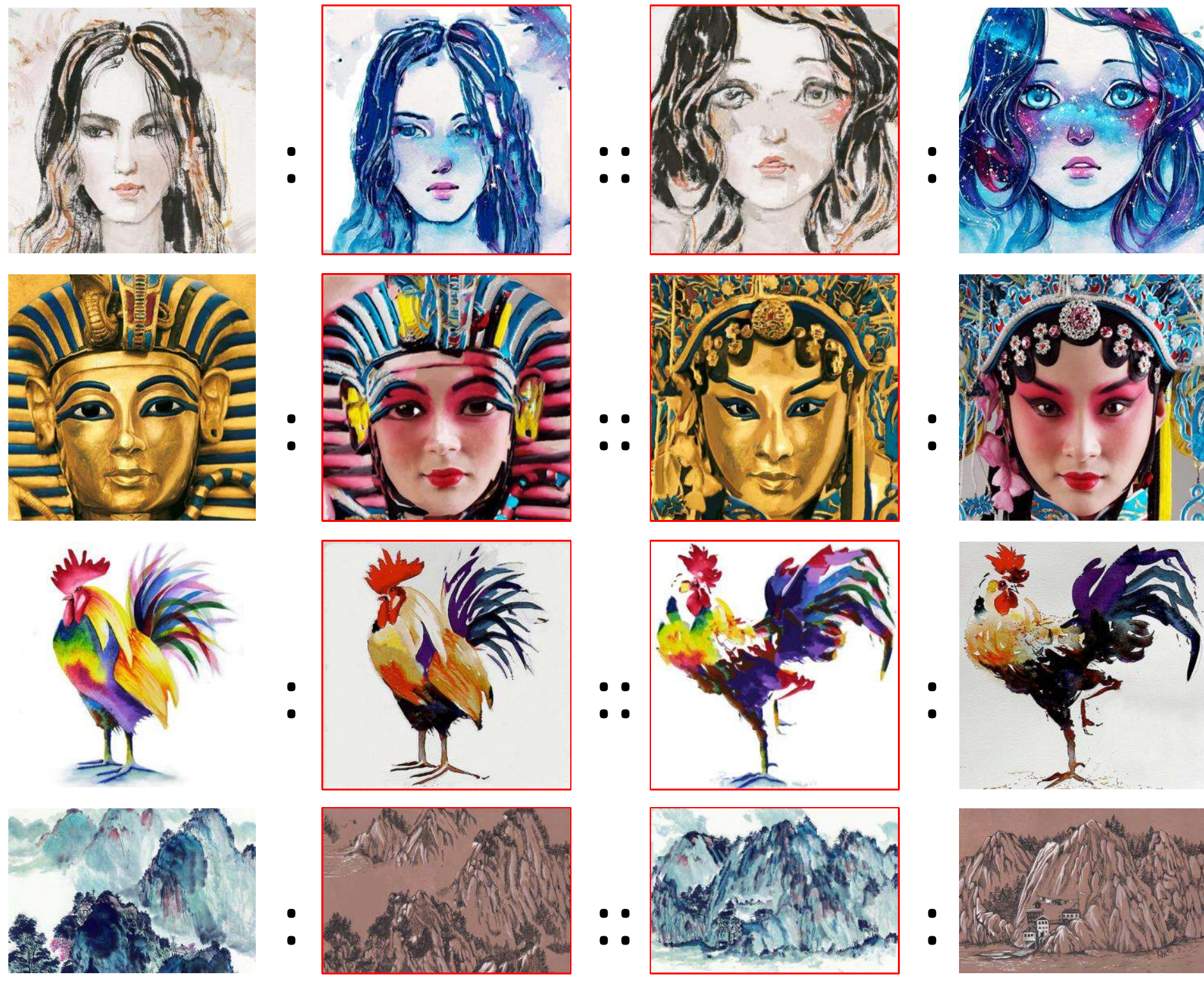}
   \centering
{
\begin{tabular}{p{.22\linewidth}p{.22\linewidth}p{.22\linewidth}p{.22\linewidth}}
\centering{$A \quad \!$(input)$\!$} & \centering{$A' \quad  \!$(output)$\!$} & \centering{$B \quad  \!$(output)$\!$} & \centering{$B'\quad  \!$(input)$\!$}
\end{tabular}
}

   \caption{Style-swap results. }
  
   \label{fig:swap}
   \end{figure*}

\subsection{Photo to Style}   

One key application of our approach is to transfer a photo to a reference artistic style. Users can easily borrow specific artwork stylization (possibly by famous or professional artists) to render their own photos for sharing and entertainment. A few examples are shown in Figure~\ref{fig:artfilter}. The recent technique of neural style transfer~\cite{alexander2015deepdream,gatys2016neural,johnson2016perceptual,chuanli2016mrf,selim2016} generated very impressive stylization results, and some apps or services (\emph{e.g.}, Prisma\footnote{http://prisma-ai.com/}, Google Deep Style\footnote{http://www.deepstylegenerator.com/}, Ostagram\footnote{https://ostagram.ru/}) based on CNN are also very popular.

In comparison to these approaches, ours is capable of higher quality content-specific stylization that better preserves structures. Neural style~\cite{gatys2016neural} and perceptual loss methods~\cite{johnson2016perceptual} rely on global statistics matching, and as a result, do not guarantee local structure-to-structure transfer. For example, the face in their results (top row of Figure~\ref{fig:stylecompare}) contains flower textures. If two inputs are totally unrelated as shown on top row of Figure~\ref{fig:stylecompare1}, both ours and theirs are visually different but acceptable. Google Deep Style and Ostagram are two online consumer services. They also do not appear to transfer structures effectively in Figure \ref{fig:stylecompare}. The portrait style approach~\cite{selim2016} enforces constraints of face alignment to maintain the face structure. Ghosting artifacts on non-face regions are evident, as shown in bottom row of Figure~\ref{fig:stylecompare}. CNNMRF~\cite{chuanli2016mrf} uses a non-parametric model, but different from ours, it is independent at each layer. In both methods, artifacts occur at misaligned regions. This can be seen at the boundary between neck and shirt on ours and the left eye on theirs (\Fref{fig:stylecompare1}(\emph{bottom row})).

The other methods optimize for the pixel color using a loss function that combines content and style information, while ours directly samples reference patches using our estimated NNF. This is the primary difference between ours and other methods. Although our results less faithfully capture the content at times, they are able to better reproduce the style. As an example, while the Neural Style method has captured the appearance of the woman's face more faithfully than ours in \Fref{fig:stylecompare}(\emph{bottom row}), our version is a better reproduction of the style.

\begin{figure*}[t] \centering
  { \footnotesize
    \setlength{\tabcolsep}{0.003\linewidth}
    \begin{tabular}{cccccc}
       \includegraphics[width=0.156\linewidth]{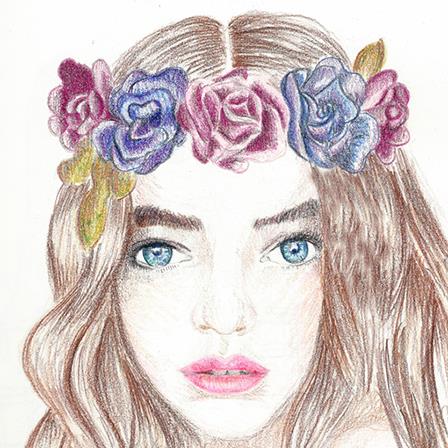} &
       \includegraphics[width=0.156\linewidth]{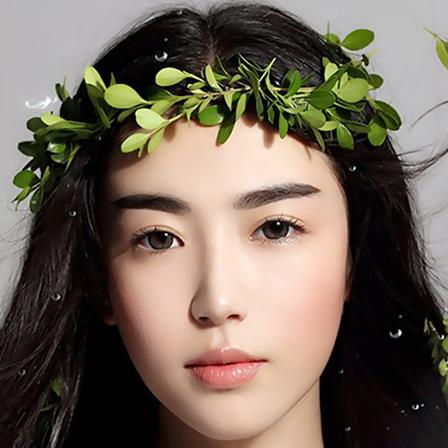} &
      \includegraphics[width=0.156\linewidth]{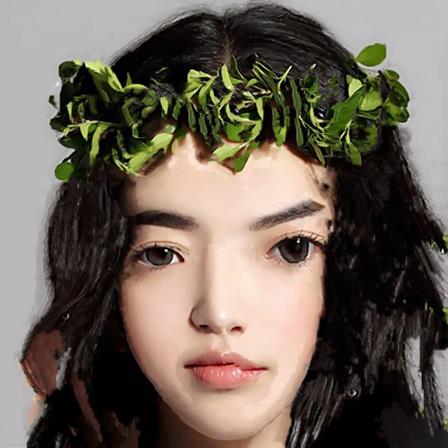} &
      \hspace{0.01\linewidth}
      \includegraphics[width=0.156\linewidth]{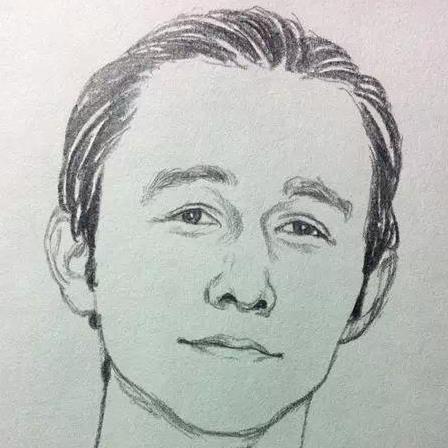} &
      \includegraphics[width=0.156\linewidth]{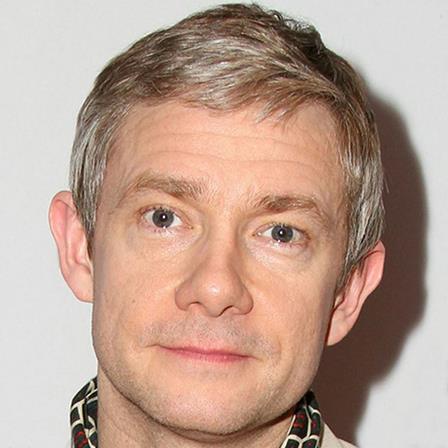} &
      \includegraphics[width=0.156\linewidth]{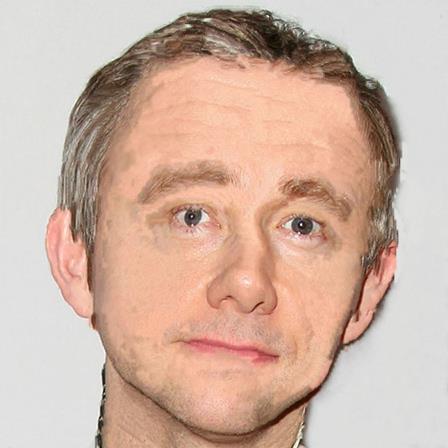} \\
  \includegraphics[width=0.156\linewidth]{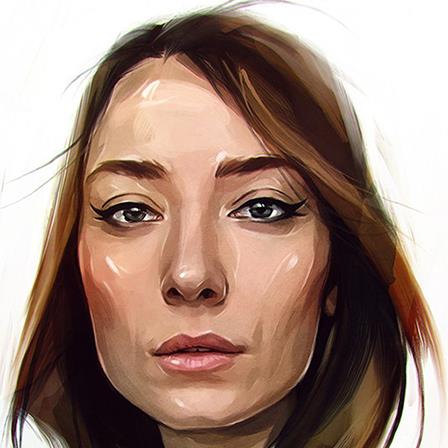} &
      \includegraphics[width=0.156\linewidth]{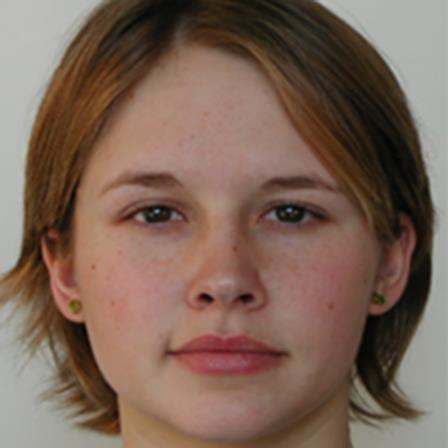} &
      \includegraphics[width=0.156\linewidth]{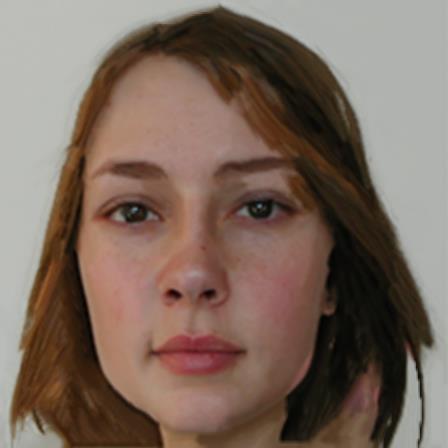} &
       \hspace{0.01\linewidth}
         \includegraphics[width=0.156\linewidth]{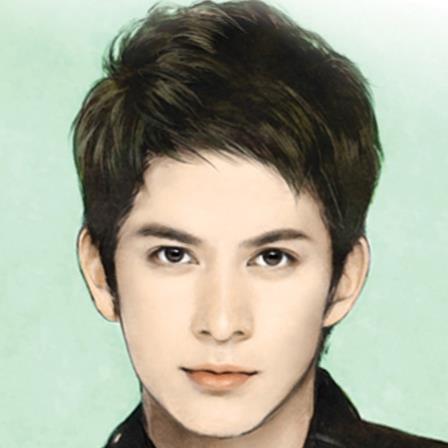} &
      \includegraphics[width=0.156\linewidth]{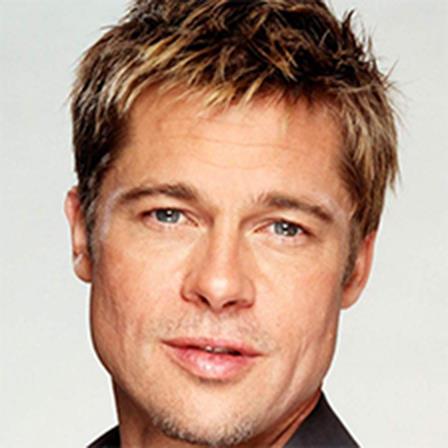} &
      \includegraphics[width=0.156\linewidth]{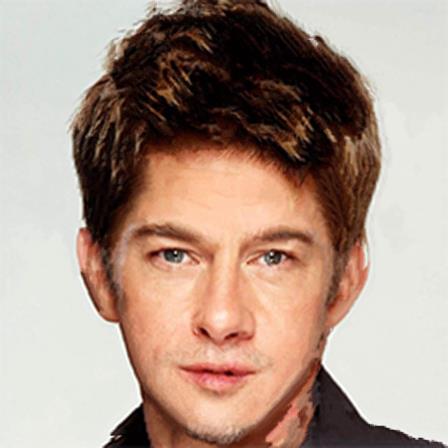} \\
 	Input (src) & Input (ref) & Output&  \hspace{0.01\linewidth} Input(src) & Input (ref) & Output
    \end{tabular}

     \caption{Results of converting a sketch or a painting to a photo.} 
   \label{fig:photo}
   }
   \end{figure*}

\subsection{Style to Style}
When input pairs of images are two content-related artworks but with vastly different styles, our method is able to swap the styles. To our knowledge, this is a new and fun effect, which has not been demonstrated before. For example, we are able to re-factor the ``Pharaoh status" to the style of ``Peking Opera facial makeup", as shown in Figure~\ref{fig:swap}(\emph{second row}). Our approach surprisingly works well for this application. Some results are shown in Figure~\ref{fig:swap}.

\subsection{Style to Photo}
   
\begin{figure*}[t] \centering
  { \footnotesize
    \setlength{\tabcolsep}{0.003\linewidth}
    \begin{tabular}{cccc}         		
      \includegraphics[width=0.238\linewidth]{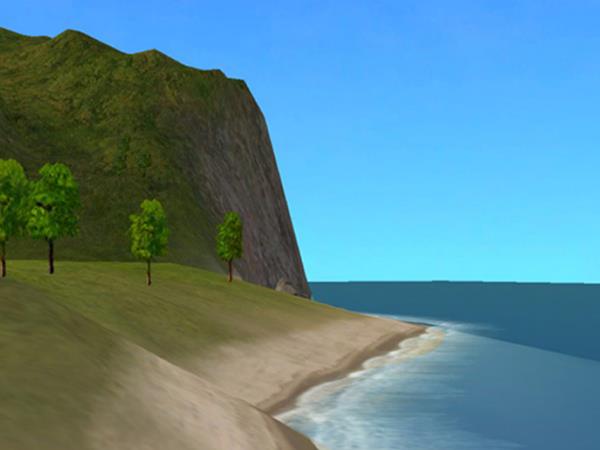} &
       \includegraphics[width=0.238\linewidth]{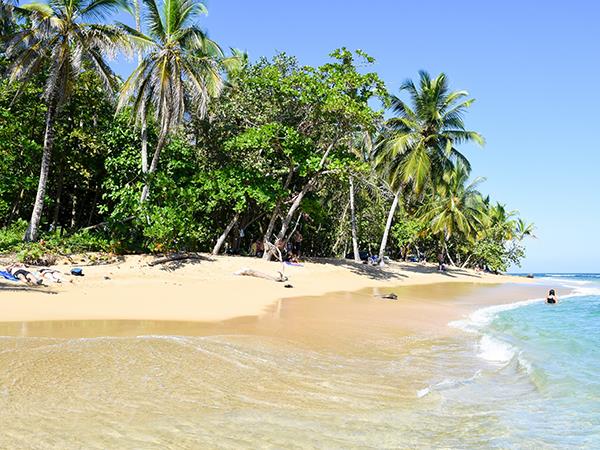} &
      \includegraphics[width=0.238\linewidth]{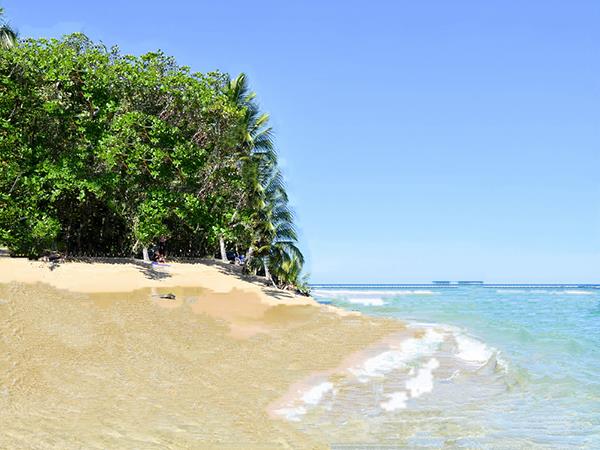} &
      \includegraphics[width=0.238\linewidth]{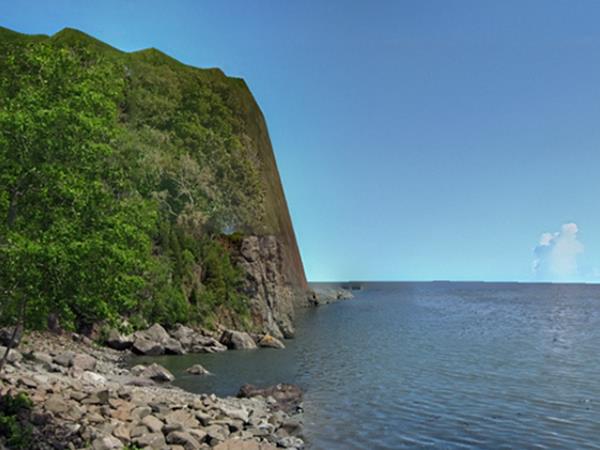} \\
 \includegraphics[width=0.238\linewidth]{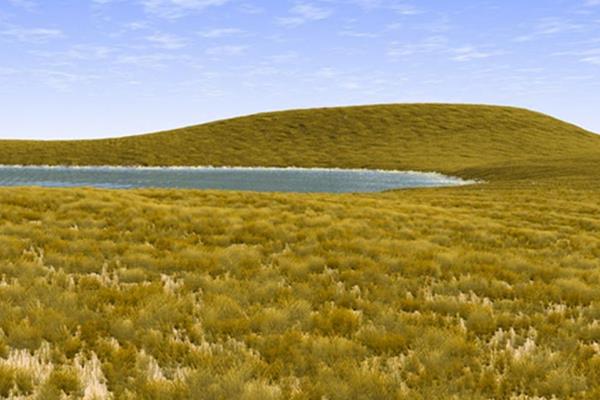} &
       \includegraphics[width=0.238\linewidth]{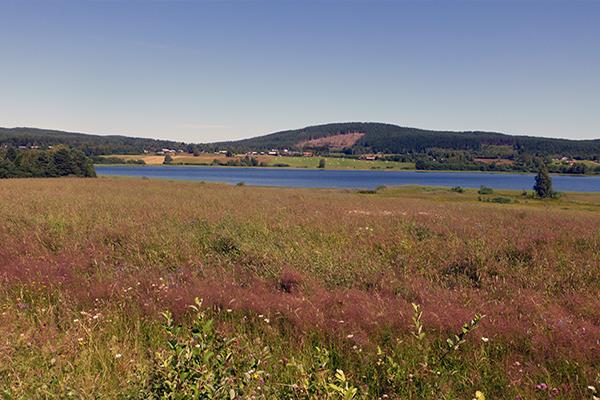} &
      \includegraphics[width=0.238\linewidth]{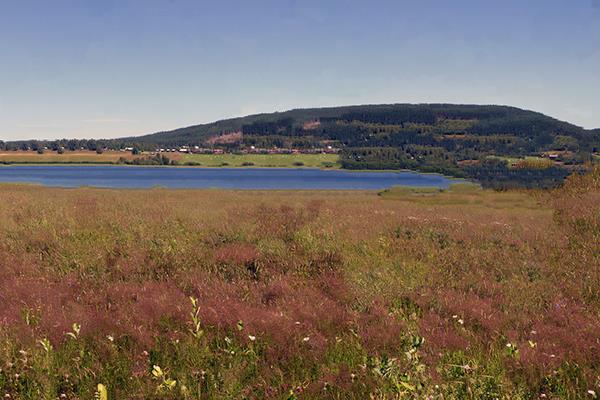} &
     \includegraphics[width=0.238\linewidth]{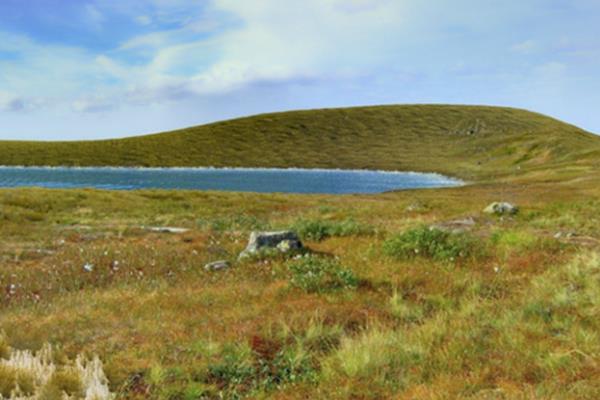} \\
   	Input (src) & Input (ref) & Ours & CG2Real

    \end{tabular}

  \caption{Comparisons of our CG-to-photo results with CG2Real method on their examples. For the inputs in each group, the source CG is used by both ours and theirs, but the reference photo is only for ours. Their multiple reference photos are not given in their paper. }
  \label{fig:p2pc}
  }
\end{figure*}

\begin{figure}[t] \centering
  { \footnotesize
    \setlength{\tabcolsep}{0.003\linewidth}
    \begin{tabular}{ccc}         		
      \includegraphics[width=0.192\linewidth]{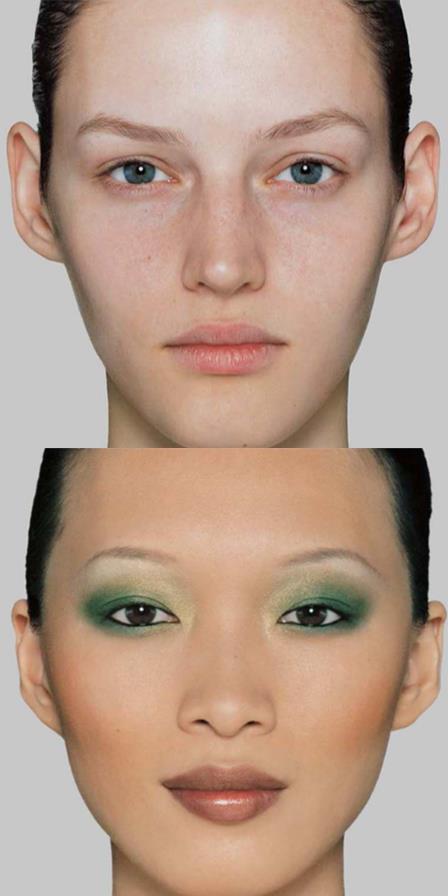} &

      \includegraphics[width=0.384\linewidth]{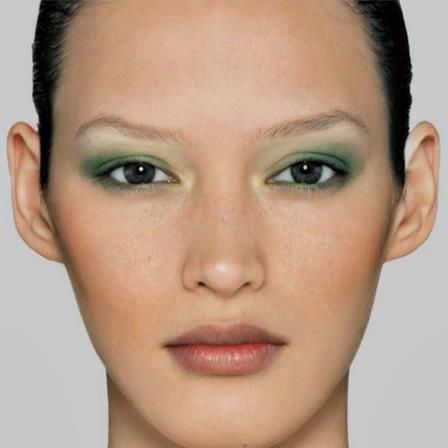}  &
        \includegraphics[width=0.384\linewidth]{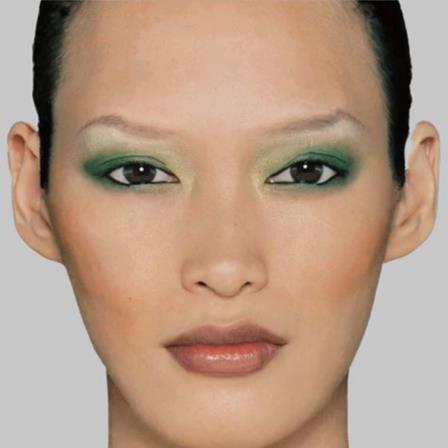} \\
      Inputs & $\hat{A'}$ (with) & $A'$ (without)
    \end{tabular}

  \caption{Comparisons of transfer color with (middle column) and without (right column) WLS refinement. }
  \label{fig:p2p}
  }
\end{figure}

This can be seen as the inverse problem of photo to style. Unfortunately, this task is much more difficult than style transfer. One reason is that  artworks tend to lack detail in favor of creativity. Another reason is that humans are very sensitive to slight structure misalignments; the quality bar for photorealism is known to be very high. Generally, our approach is less successful on this task compared with the photo-to-style application. However, our approach is still able to transfer sketches or paintings to real photographs, with more plausible results when both images are very related. Figure~\ref{fig:photo} show some example results.

We further compare our methods with the CG2Real~\cite{johnson2011cg2real} on their provided examples, shown in Figure~\ref{fig:p2pc}. They retrieve a small number of photos with similar global structure of the input CG, identify the corresponding regions by co-segmentation and transfer the regions to CG. In contrast to their region-to-region transfer, our approach builds a pixel-to-pixel mapping, which can better maintain the structures of the input CG image, shown in Figure~\ref{fig:p2pc}(\emph{first row}). Moreover, our results are visually comparable to theirs, even though we use only one reference photo while CG2Real uses multiple photos.

\subsection{Photo to Photo}
Photo to photo mapping can be in the form of color or tone transfer. Our approach enables local transfer in corresponding regions. Generally, only color or tone are borrowed from the reference, not allowing details. For this goal, we slightly modify our algorithm by applying a refinement on the reconstructed result. Specifically, we use weighted least squares filter (WLS)~\cite{farbman2008edge} to smooth both input image $A$ and reconstructed result $A'$ with the guidance $A$. The final result $\widetilde{A}'$ is obtained by:
\begin{equation}
\widetilde{A}'=\text{WLS} (A',A)+A-\text{WLS} (A,A),
\end{equation}
which implies that we only need to maintain colors of $A'$ while discarding details of $A'$, and filling in with details of $A$. It is similar for $B$.
Without the refinement, the details from the reference will introduce distortion to the fine-grained structures of objects in the source image, for example, the eyes and nose in the comparison with and without the refinement (Figure~\ref{fig:p2p}).

\begin{figure*}[!h] \centering
  { \footnotesize
    \setlength{\tabcolsep}{0.003\linewidth}
    \begin{tabular}{ccccc}    
		 &     
      \includegraphics[width=0.188\linewidth]{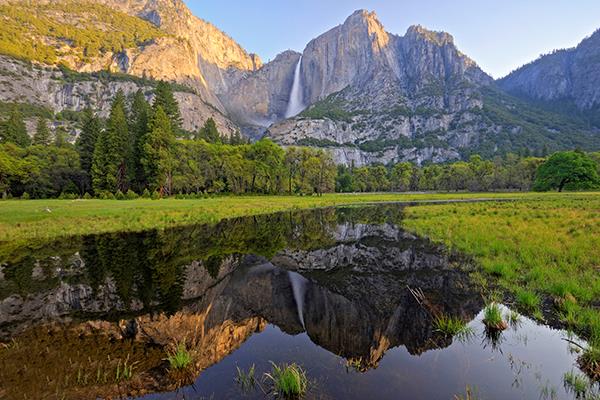} &     
      \includegraphics[width=0.188\linewidth]{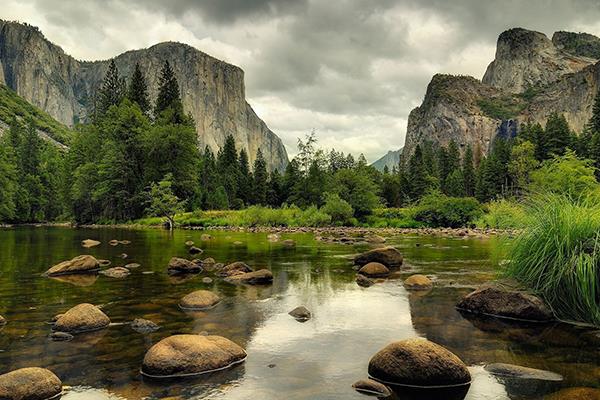} &
      \includegraphics[width=0.188\linewidth]{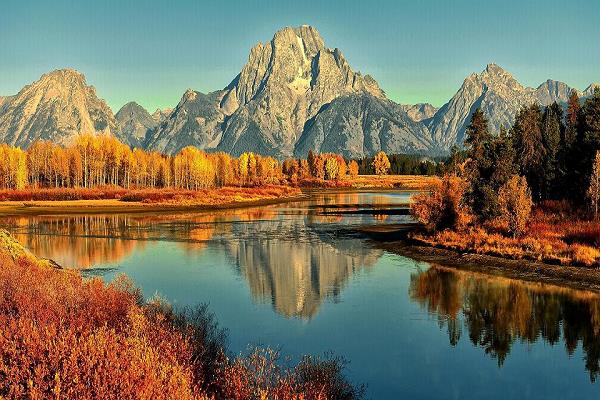} &
      \includegraphics[width=0.188\linewidth]{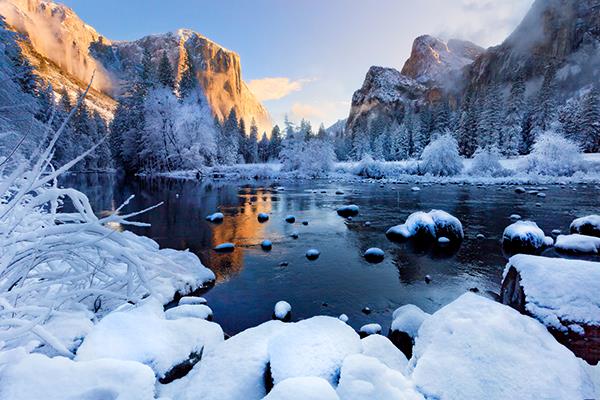} \\
&Input (ref 1) & Input (ref 2) & Input (ref 3) &Input (ref 4) \\
          \includegraphics[width=0.188\linewidth]{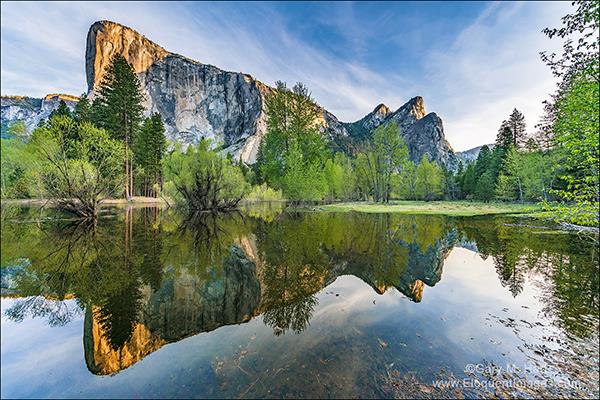}   &     
      \includegraphics[width=0.188\linewidth]{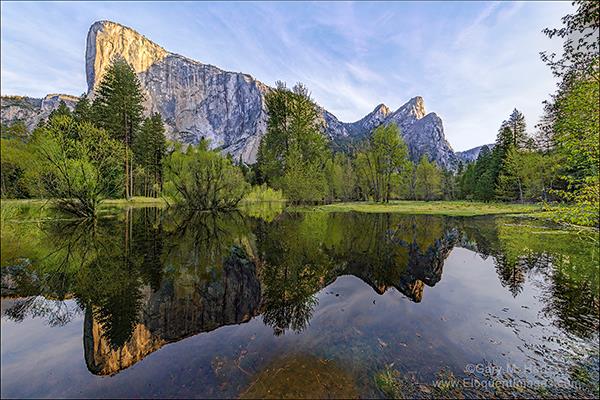} &     
      \includegraphics[width=0.188\linewidth]{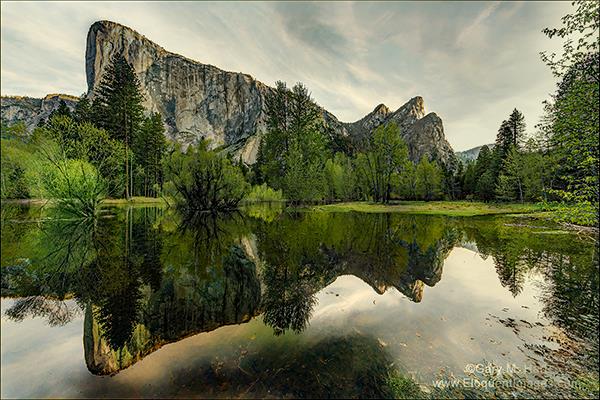} &
      \includegraphics[width=0.188\linewidth]{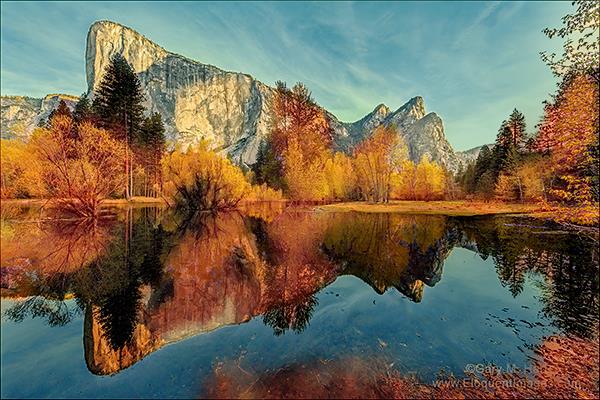} &
      \includegraphics[width=0.188\linewidth]{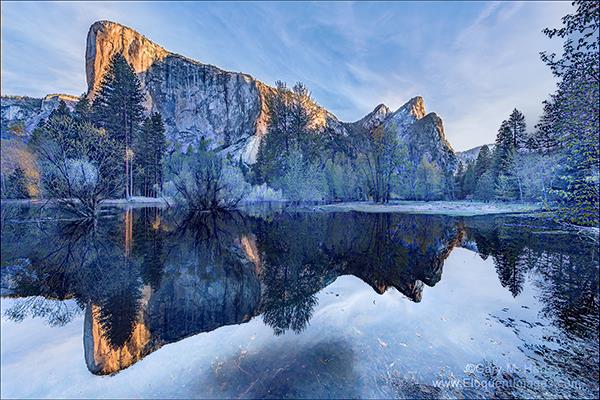}     
            \\  
 Input (src)&Output 1 &Output 2 & Output 3 &Output 4
    \end{tabular}

  \caption{Results of generating time-lapse sequences with references of another semantic-related scene. }
  \label{fig:timelapse}
  }
\end{figure*}

\begin{figure*}[!h] \centering
  { \footnotesize
  
    \includegraphics[width=0.97\linewidth]{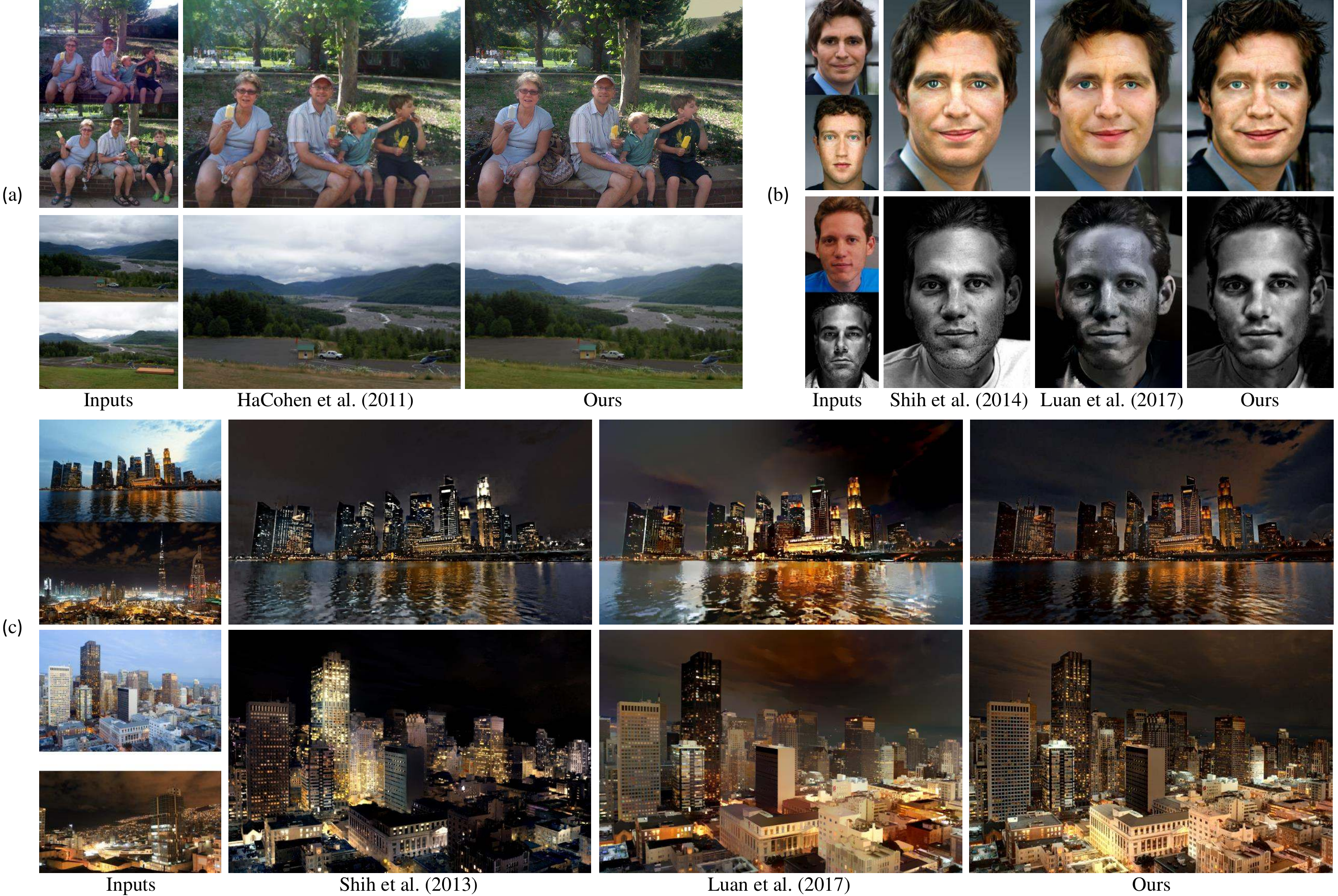}

  \caption{Comparison of our photo-to-photo results with other state-of-art methods on their examples. For the inputs in each group, the upper one is the source photo and the lower one is the reference photo.}
  \label{fig:colortrans} 
  }
\end{figure*}

We further show two examples compared with NRDC~\cite{haCohen2011nrdc} in Figure~\ref{fig:colortrans}(a). NRDC uses a global color mapping curve based on a small potion of matches. By contrast, our transfer is local, which may produce better region-to-region transfer than NRDC, as shown in the grass region of Figure~\ref{fig:colortrans}(a) (\emph{second row}). Such local transfer is sometimes sensitive to the quality of matching. For instance, we observe some inaccurate color mappings, like saturated sky region in Figure~\ref{fig:colortrans}(a) (\emph{first row}), caused by mismatched regions. Beyond the capability of NRDC, our approach can also transfer color across images with different scenes, such as the time-lapse example shown in Figure~\ref{fig:timelapse}, due to the identified semantic correspondences, e.g., tree to tree, mountain to mountain.

Compared with other local color transfer methods, e.g. shih et al.\cite{shih2013data} and Luan et al.\cite{luan2017deep} in Figure~\ref{fig:colortrans}(c), our method can produce visually comparable results to theirs. However, our algorithm takes a single image as reference, while their methods require either a full time-lapse video \cite{shih2013data} or optional precomputed segmentation masks \cite{luan2017deep}. When compared with Luan et al.\cite{luan2017deep} carefully, we can find some undesired posterization effects (e.g., forehead in Figure~\ref{fig:colortrans}(b) and buildings in Figure~\ref{fig:colortrans}(c)), which do not occur in ours. Compared with Shih et al. \cite{shih2014style}, both ours and Luan et al. \cite{luan2017deep} fail to transfer some details (e.g., eye highlights) as shown in Figure~\ref{fig:colortrans}(b).

\begin{figure}[t] \centering
  { \footnotesize    
   
\includegraphics[width=0.95\linewidth]{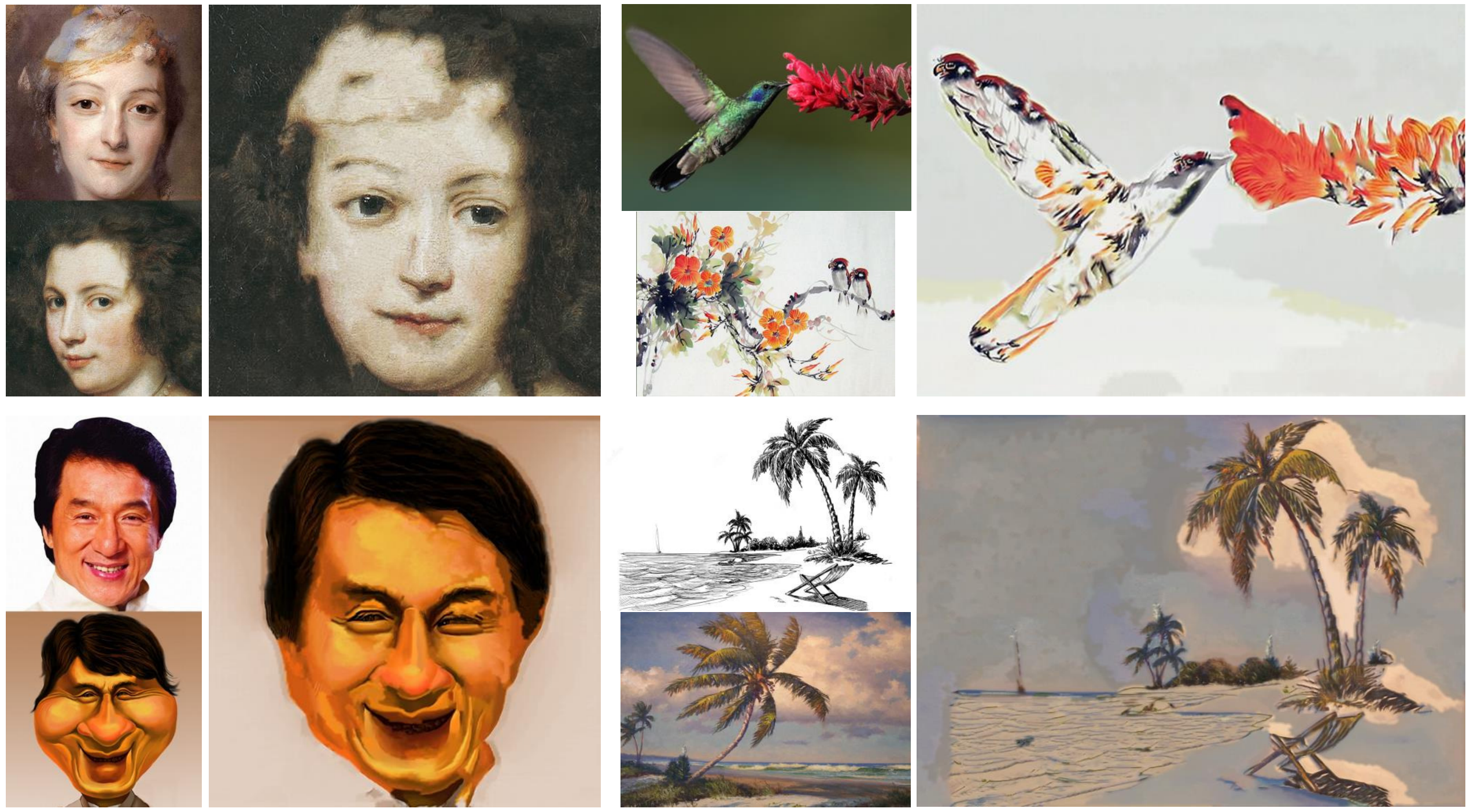} 

     \caption{Some examples of failure cases.}
   \label{fig:limit}
   }
   \end{figure}
   
\section{Discussion}
\label{sec:limit}

Semantic-level matching is not fool-proof, since our deep match approach relies on a pre-trained VGG network that has limited capability in representing all objects or fine-grained structures. A possible improvement would be to train our network on a domain-specific dataset for better image representation. If an object is found in one image but not the other, it is not clear how semantic matching can be done. We show a failure case in Figure~\ref{fig:limit}(\emph{top-left}), which shows the mismatched hat region. 

For the scenes which are semantically related but vary a lot in scales and view points, our method still fails to build correct correspondences as shown in Figure~\ref{fig:limit}(\emph{top-right}). Addressing these cases would require either pre-scaling images or allowing patches to be rotated and scaled in our NNF search like Generalized PatchMatch. However adding rotation or scale is nontrivial since geometric transformations in the image domain are not well preserved with those in feature domain because of non-linear modules (\emph{e.g.}, ReLU).  

Moreover, our method may fail to find correspondences in textureless regions that have very low neural activation, like the background in Figure~\ref{fig:limit}(\emph{bottom-right}). This problem may be addressed by either analyzing pixel intensity or by explicitly enforcing smoothness in our energy function.

Our photo-to-style transfer application is unable to produce geometry style transfer, like the case shown in Figure~\ref{fig:limit}(\emph{bottom-left}). The assumption in our work is to maximally preserve content structure. We may relax the assumption in future work.

\section{Concluding Remarks}

We have demonstrated a new technique for transferring visual attributes across semantically-related images. We adapted the notion of image analogy to a deep feature space for finding semantically-meaningful dense correspondences. We show that our method outperforms previous methods where image pairs exhibit significant variance in their appearance including lighting, color, texture, and style. We have shown that our approach is widely applicable for visual attribute transfer in real-world images, as well as additional transfer challenges such as content-specific stylization, style to photo, and style to style. We believe this method may also be proven useful for a variety of computer graphics and vision applications that rely on semantic-level correspondences. 

\section*{Acknowledgements}

We thank the anonymous reviewers for helping us to improve this paper, David Wipf for his help in proofreading. We also acknowledge to the authors of our image and style examples but we do not own the copyrights of them.

\bibliographystyle{acmsiggraph}
\nocite{*}
\bibliography{analogy}
\end{document}